\documentclass[lettersize,journal]{IEEEtran}
\usepackage{amsmath,amssymb,amsfonts}
\usepackage{algorithmic}
\usepackage{floatrow}
\usepackage{algorithm}
\usepackage{array}
\usepackage[label font=bf,labelformat=simple]{subfig}
\usepackage{booktabs}
\usepackage{enumitem}
\usepackage{multirow}
\usepackage{textcomp}
\usepackage{stfloats}
\usepackage{url}
\usepackage{verbatim}
\usepackage{graphicx}
\usepackage{xcolor}
\usepackage{cite}
\usepackage{soul}
\DeclareMathSymbol{\mhyphen}{\mathord}{AMSa}{"39}
\pdfoutput=1
\begin{document}

\title{On Noise Stability and Robustness of Adversarially Trained Networks on NVM Crossbars
\thanks{This work was supported in part by the National Science
Foundation, Vannevar Bush Faculty Fellowship, the
Center for Brain-Inspired Computing (C-BRIC), one of six centers in JUMP, funded by Semiconductor Research Corporation (SRC) and DARPA, and by the U.S. Department of Energy, Office of Science, for support of  microelectronics research, under contract number DE-AC02-06CH11357.}
}

\author{\IEEEauthorblockN{Chun Tao\IEEEauthorrefmark{1},
Deboleena Roy\IEEEauthorrefmark{1},
Indranil Chakraborty\IEEEauthorrefmark{1},
Kaushik Roy\IEEEauthorrefmark{1}}\\
\IEEEauthorblockA{\IEEEauthorrefmark{1}School of Electrical and Computer Engineering, Purdue University, West Lafayette, Indiana 47907\\
tao88@purdue.edu, roy77@purdue.edu, ichakra@purdue.edu, kaushik@purdue.edu}
}

\maketitle

\begin{abstract}
Applications based on Deep Neural Networks (DNNs) have grown exponentially in the past decade. To match their increasing computational needs, several Non-Volatile Memory (NVM) crossbar based accelerators have been proposed. Recently, researchers have shown that apart from improved energy efficiency and performance, such approximate hardware also possess intrinsic robustness for defense against adversarial attacks. Prior works have focused on quantifying this intrinsic robustness for vanilla networks, that is DNNs trained on unperturbed inputs. However, adversarial training of DNNs, i.e. training with adversarially perturbed images, is the benchmark technique for robustness, and sole reliance on intrinsic robustness of the hardware may not be sufficient. 

In this work, we explore the design of robust DNNs through the amalgamation of adversarial training and the intrinsic robustness offered by NVM crossbar based analog hardware. First, we study the noise stability of such networks on unperturbed inputs and observe that internal activations of adversarially trained networks have lower Signal-to-Noise Ratio (SNR), and are sensitive to noise compared to vanilla networks. As a result, they suffer significantly higher performance degradation due to the approximate computations on analog hardware; on an average $2\times$ accuracy drop. Noise stability analyses clearly show the instability of adversarially trained DNNs. On the other hand, for adversarial images generated using Square Black Box attacks, ResNet-10/20 adversarially trained on CIFAR-10/100 display a robustness improvement of $20-30\%$ under high $\epsilon_{attack}$ (degree of input perturbation). For adversarial images generated using Projected-Gradient-Descent (PGD) White-Box attacks, the adversarially trained DNNs present a $5-10\%$ gain in robust accuracy due to the underlying NVM crossbar when $\epsilon_{attack}$ is greater than the {\em epsilon} of the adversarial training ($\epsilon_{train}$). Our results indicate that implementing adversarially trained networks on analog hardware requires careful calibration between hardware non-idealities and $\epsilon_{train}$ to achieve optimum robustness and performance.
\end{abstract}


\section{Introduction}
Today, Deep Neural Networks (DNNs) are the backbone of most Artificial Intelligence (AI) applications and have become the benchmark for a gamut of optimization tasks. These networks, trained using Deep Learning (DL) \cite{goodfellow2016deep}, have demonstrated superlative performance against traditional methods in the fields of Computer Vision \cite{voulodimos2018deep}, Natural Language Processing \cite{young2018recent}, Recommender Systems \cite{cheng2016wide}, etc. Their widespread usage has spurred rapid innovations in the design of dedicated hardware for such Machine Learning (ML) workloads. 

A significant portion of the computations inside a DNN are Matrix Vector Multiplications (MVM). Latest special-purpose accelerators such as GoogleTPU \cite{jouppi2017datacenterold}, Microsoft BrainWave  \cite{chung2018serving}, and NVIDIA V100 \cite{nvidiav100} improve the MVM operations by co-locating memory and processing elements, and thereby achieve high performance and energy efficiency. However, these accelerators are built on digital CMOS which faces the arduous challenge of scaling saturation \cite{xu2018scaling}. To counter the eventual slowdown of Moore's Law \cite{schaller1997moore} and to alleviate the "memory wall", several Non-Volatile Memory (NVM) technologies such as RRAM \cite{wong2012metal}, PCRAM \cite{wong2010phase} and spin devices \cite{fong2015spin} have been developed as promising alternatives. Memory is programmed as resistance levels within a device, which are then arranged in a crossbar-like manner, leading to efficient MVM operations. Note, MVM operations are executed in the analog domain, and can significantly reduce power and latency when compared to digital von-Neumann CMOS architectures \cite{chakraborty2020resistive}. In recent years, several NVM crossbar based accelerators have been proposed \cite{shafiee2016isaac,ankit2019puma} that leverage this compute efficiency. Note, however, such in-memory crossbar based computing can also be done using standard CMOS based memories, albeit with much higher area. 

While design of specialized ML hardware can address the growing computational needs of DNN-based applications, there still remains several challenges in the ubiquitous adoption of Deep Learning. One of the primary concerns is the susceptibility of DNNs to malicious actors. First noted by \cite{szegedy2013intriguing}, now it is widely known that the performance of a DNN can be artificially degraded with carefully designed ``adversarial" inputs. By exploiting gradient-based learning mechanism, an adversary can generate new data that is visually indistinguishable from the original data, but it forces the model to classify incorrectly. Such attacks on DNNs are called ``adversarial attacks" and their strength depends on how much information the attacker has of the model and its training data \cite{goodfellow2014explaining}. Securing the model weights and the training data, and obscuring the gradients of the model itself have been partially effective against such adversaries \cite{dhillon2018stochastic, buckman2018thermometer, athalye2018obfuscated}. However, the most effective defense against such attacks is ``adversarial training" \cite{madry2018towards, shafahi2019adversarial, wong2019fast}, where the model is trained on the adversarial inputs. 

When implementing DNNs in any real-world application, both energy efficiency and security play significant roles. Up until recently, research in these two domains have progressed almost independently of each other. In the recent past, there have been several works that explore Algorithm-Hardware co-design opportunities for robust and efficient implementation \cite{guesmi2021defensive, cappelli2021adversarial, roy2020robustness}. In particular, it has been shown that approximate/analog hardware can provide additional robustness to DNNs by obscuring the true gradients and parameters. While prior works have focused on the robustness gain for models with no other defenses, in this work, we analyze the performance of adversarially trained models on approximate hardware, specifically, NVM crossbars. 

\begin{figure*}[t]
	\centering
	\includegraphics[width=0.8\textwidth, keepaspectratio]{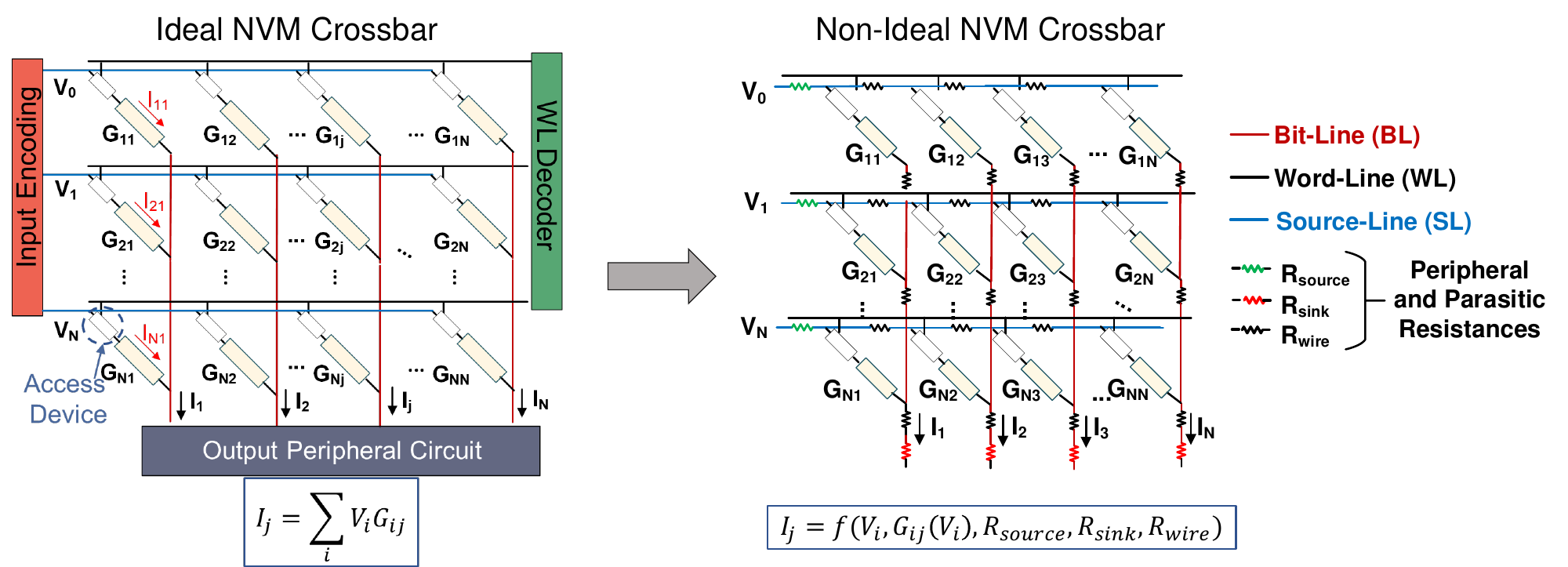}
	\caption{(Left) Illustration of NVM crossbar which produces output current $I_j$, as a dot-product of voltage vector, $V_i$ and NVM device conductance, $G_{ij}$. (Right) Various peripheral and parasitic resistances modify the dot-product computations into an interdependent function of the analog variables (voltage, conductance and resistances) in a non-ideal NVM crossbar. \vspace{-0.5cm}}
	\label{fig:crossbar}
\end{figure*}

When an MVM operation is executed in an NVM crossbar, the output is sensed as a summation of currents through the resistive NVM devices. The non-ideal behaviour of the crossbar and its peripheral circuits results in error in the final calculation, and overall performance degradation of the DNN \cite{chakraborty2018technology, chakraborty2020geniex}. As these deviations depend on multiple analog variables (voltages, currents, resistances), they are nearly impossible to estimate without a complete model of the architecture and the device parameters, which are difficult for an attacker to obtain. In \cite{roy2020robustness}, the authors demonstrated that this non-ideal behaviour can be utilized to obscure the true gradients of the DNN and provide robustness against various types of adversarial attacks. Their analyses of the benefit of intrinsic robustness was limited to vanilla DNNs, i.e. networks trained on the unperturbed, or ``clean", images. However, for a robust DNN implementation, adversarial training is essential. Hence, in this work, we study the performance and robustness of adversarially trained networks when implemented on such NVM crossbars. We observe that adversarial training reduces the noise stability of DNNs, i.e. its ability to withstand perturbations within the network during inference. To understand this behavior of adversarially trained DNNs, we computed noise stability parameters formulated in \cite{arora2018stronger}. We also note that the robustness gain from the crossbar is applicable only for certain degrees of attack perturbations. Finally, we compare NVM crossbar non-idealities with Gaussian noise injection. To summarize, our main contributions are:
\begin{itemize}
    \item We explored the challenges of designing adversarially robust DNNs through the amalgamation of adversarial training and the intrinsic robustness offered by NVM crossbar based analog computing hardware.
    \item We analyzed the performance of adversarially trained DNNs on NVM crossbars for unperturbed images.  Compared against vanilla DNNs (networks trained on unperturbed data), we observed that adversarially trained networks are less noise stable, and hence, suffer greater performance degradation on NVM crossbar based analog hardware.
    \item We performed a noise stability analysis and computed noise stability parameters for vanilla and adversarially trained DNNs, demonstrating that adversarially trained DNNs are less noise stable. \cite{arora2018stronger}.
    \item For Non-Adaptive Square Black Box attacks \cite{ACFH2020square}, we demonstrated that the non-idealities incur a loss in robustness when $\epsilon_{attack}$, the degree of input perturbations, is small. However, when $\epsilon_{attack}$ is large enough, the NVM crossbar non-idealities provide a robustness gain to the DNNs.
    \item For Non-Adaptive Projected-Gradient-Descent White-Box Attacks \cite{madry2018towards}, we demonstrated that the non-idealities provide a gain in robustness when the epsilon of the adversarial attack ($\epsilon_{attack}$) is greater than the epsilon of the adversarial training ($\epsilon_{train}$). In fact, with careful co-design, one can implement a DNN trained with lower $\epsilon_{train}$ that will have the same or even higher robustness than a DNN trained with a higher $\epsilon_{train}$ while maintaining a higher natural test accuracy.
\end{itemize}

The rest of the paper is organized as follows. In section {\ref{sec:back:hw}}, we provide a background on NVM crossbar based analog hardware, and in section {\ref{sec:back:sw}} the prior works on robustness from analog computing is discussed, followed by section {\ref{sec:back:stability}} where we introduce works on noise stability. Next, in section {\ref{sec:eval}}, we describe in detail the evaluation framework used in our experiments. This is followed by section {\ref{sec:results}}, where we report our findings and discuss the implications. Finally, in section {\ref{sec:conclusion}}, we provide the conclusions of our findings.

\section{Background and Related Work}
\label{sec:back}

\subsection{In-Memory Analog Computing Hardware}
\label{sec:back:hw}

In-memory analog computing with NVM technologies are being extensively studied for machine learning (ML) workloads \cite{burr2015experimental, ambrogio2018equivalent, cai2019fully} because of their inherent ability to perform efficient matrix-vector multiplications, the key computational kernel in DNNs. The NVM analog compute primitive comprises of a two-dimensional cross-point array with peripheral circuits at the input to provide analog voltages; and peripherals at the output to convert analog currents to digital values, as shown in Fig. \ref{fig:crossbar}. At the intersection of horizontally (source-line) and vertically (bit-line) running metal lines, there are NVM devices that can be programmed to discrete number of conductance states \cite{hu2016dac}. To perform MVM operation using NVM crossbars, analog voltages, $V_i$, are applied simultaneously at the source-lines. The multiplication operations are performed between vector $V_i$ and conductance matrix, $G_{ij}$, encoded in each NVM device using the principle of Ohm's law. Finally, the product, $V_iG_{ij}$, representing the resulting current, $I_{ij}$, through each NVM device, is summed up using Kirchoff's current law to produce dot-product outputs, $I_j$ at each column:
\begin{equation}
    I_j =\sum_i I_{ij} =\sum_i V_iG_{ij}
\end{equation}
As all the columns operate in parallel, NVM crossbars can reduce the $O(N^2)$ MVM operation between input vector $V$ and conductance matrix $G$, onto $O(1)$ to produce output vector $I=VG$. 

Despite the inherent parallelism in NVM crossbars, while executing the MVM operation, the computation itself can suffer from errors due the the analog nature of computing. These errors are introduced due to several non-idealities arising from the I-V characteristics of the NVM devices, as well as peripheral and parasitic resistances such as $R_{source}$, $R_{sink}$, $R_{wire}$, as shown in Fig. \ref{fig:crossbar}. The design parameters for NVM crossbars, such as crossbar size, ON resistance etc. can have significant impact on the severity of the functional errors introduced by these non-idealities \cite{chakraborty2020geniex}. That is due to the the dependence of the effective resistance of a crossbar column on these design parameters. 

For example, larger crossbar sizes reduce the effective resistance, thus resulting in more errors due to parasitic and peripheral resistances. On the other hand, a higher ON resistance of the device increases the effective resistance, leading to less errors. Overall, these design parameters govern the relationship between the output current vector under the influence of non-idealities, $I_{ni}$, and the input arguments, $V$, and $G$. Effectively, in presence of non-idealities in NVM crossbar, the non-ideal output current vector, $I_{ni}$, can be expressed as:
\begin{equation}
    I_{ni} = f(V, G (V), R_{source}, R_{sink}, R_{wire})
    \label{eq:non-ideal}
\end{equation} 
where the conductance matrix $G(V)$ is a function of $V$.

NVM crossbar non-idealities can be either linear (source, sink, and wire resistances, etc.) or non-linear (access device or selector non-linearity, shot and thermal noise, etc.). The non-linear behaviors can also be data-dependent. We use GENIEx \cite{chakraborty2020geniex}, a neural network based crossbar simulation tool, to model the non-idealities of NVM crossbars, with the non-ideal relationship shown in Eq. \ref{eq:non-ideal}. NVM crossbar non-idealities cause deviations in activations of every layer from their expected value. This deviation accumulates as it propagates through the network, and results in degradation in DNN accuracy during inference. Incidentally, such difference in activation and expected value have been also demonstrated to impart adversarial robustness under several kinds of attacks \cite{roy2020robustness}.

\subsection{Adversarial Robustness with Analog Computing}
\label{sec:back:sw}
Prior works have shown that various facets of analog computing can be incorporated for adversarial robustness. In \cite{cappelli2021adversarial}, the authors use an Optical Processor Unit (OPU) as an analog defense layer. It performs a fixed random transformation, and obfuscates the gradients and the parameters, to achieve adversarial robustness. While \cite{cappelli2021adversarial} used analog processing as an extra layer in their computation, \cite{roy2020robustness} implemented neural networks on an NVM crossbar based analog hardware, and analyzed the impact on accuracy and robustness. They demonstrated two types of adversarial attacks, one where the attacker is unaware of the hardware (Non-Adaptive Attacks), and the second being Hardware-in-Loop attacks, where the attacker has access to the NVM crossbar. They showed that during Non-Adaptive Attacks on a vanilla DNN, i.e. a network trained on unperturbed images with no other defenses, the gradient obfuscation by the analog hardware provides substantial robustness against both Black Box and White Box Attacks. However, once the attacker gains complete access to the hardware, it can generate Hardware-in-Loop attacks and this significantly diminishes the robustness gain observed earlier. In this work, we further expand the analysis of Non-Adaptive Attacks on adversarially trained DNNs implemented on NVM crossbars.
\vspace{-2mm}

\subsection{Noise Stability Analysis}
\label{sec:back:stability}
To describe the resilience of pretrained DNNs to noise perturbations, multiple works have investigated the noise stability of DNNs. A notion of noise stability can be posed as the stability of output with respect to random noise injected at the unit nodes \cite{morcos2018noise}. In \cite{langford2002noise}, the authors performed noise sensitivity analysis on learned models, and associated "flatness" of a network to the stability of its output against noise added to its trained parameters.
Similarly, \cite{roy2021noise} defined noise stability of a DNN to be its ability to withstand perturbations within the network during inference. In this work, we compute stability parameters \cite{arora2018stronger} for both vanilla and adversarially trained DNNs, and use the analyses to show the vulnerability of adversarially trained DNNs to analog hardware evaluation, described below:
\begin{itemize}[leftmargin=1em]
\item \textbf{Layer Cushion}
The layer cushion of convolutional layer $i$ is defined to be the largest number $\mu_i$ such that for any input $x \in S$ (sample set),
\begin{equation}
\vspace{-5pt}
    \mu_i \lVert A^i \rVert _{F} \lVert \phi(x^{i-1}) \rVert \leq \lVert A^i\phi(x^{i-1})\rVert 
\label{eq:lc}
\end{equation}
In the above equation, $A^i$ is the convolutional matrix of layer $i$, $x^{i-1}$ is the vector right before ReLU after convolutional layer $i$ and batch normalization, and $\phi$ operator indicates the RELU activation function.

$1/{\mu_i}^2$ is equal to the noise sensitivity of $A^i$ at input $\phi(x^i)$ with respect to Gaussian noise $\eta \sim \mathcal{N}(0,I)$ \cite{arora2018stronger}. Layer cushion thus relates to the degree of deviation at each convolution layer when the DNN is evaluated on analog hardware.

\item \textbf{Interlayer Cushion}
The interlayer cushion between convolutional layer i and j, where $i<j$, is defined to be the largest number $\mu_{i,j}$ such that for any $x\in S$,
\begin{equation}
\vspace{-5pt}
    \mu_{i,j} \cdot \frac{1}{\sqrt{n_1^i n_2^i}} \lVert J^{i,j}_{x^i} \rVert _F \lVert x^i\rVert \leq \lVert J^{i,j}_{x^i}x^i \rVert
\vspace{-1pt}
\label{eq:interlc}
\end{equation}
For two layer $i \leq j$, $M^{i.j}$ denotes the operator for composition of these layers \cite{arora2018stronger}. $J^{i,j}_{x^i}$ is the Jacobian of this operator at input $x^i$. The input to convolutional layer $i$ has dimension $n_1^i \times n_2^i$. The normalization term $\frac{1}{\sqrt{n_1^i n_2^i}}$ is from Definition 8 of \cite{arora2018stronger}. During the forward propagation, the degree of output noise at the $j^{th}$ layer scales with the number of parameters of the weight filter. For the convolutional layer,
the upper bound of the error is $\frac{1}{\sqrt{h^{i}n_{1}^{i}n_{2}^{i}}}$ compared to $\frac{1}{\sqrt{h^i}}$ in the fully connected layer \cite{arora2018stronger}. $1/{\mu_{i,j}}^2$ is the upper bound of noise sensitivity of $J^{i,j}_x$ at $x^i$, with respect to Gaussian noise $\eta \sim \mathcal{N}(0,I)$ \cite{arora2018stronger}. If different pixels in $\eta$ are independent Gaussian variables, we can expect $\lVert J^{i,j}_{x^i}\eta \rVert \approx \frac{1}{\sqrt{h^{i}n_{1}^{i}n_{2}^{i}}} \lVert J^{i,j}_{x^i} \rVert _F \lVert \eta \rVert$, explaining the extra factor $\frac{1}{\sqrt{n_1^i n_2^i}}$. Similar to layer cushion, $\mu_{i,j}$ explains the deviations accumulated over layer-wise convolutions due to analog computation.

\item \textbf{Activation Contraction}
Activation contraction of layer i is defined to be the smallest number $c$ such that for any $x\in S$,
\begin{equation}
\vspace{-5pt}
    \lVert \phi(x^i) \rVert \geq \lVert x^i \rVert /c
\vspace{-1pt}
\label{eq:ac}
\end{equation}

$c$ indicates the ability of a DNN to filter out noise generated during convolution through its activation layers.

\item \textbf{Interlayer Smoothness}
A Gaussian noise $\eta$ is injected at $x^i$ for layer $i$. In our experiments, the norm of $\eta$ is $10\%$ that of $x^i$. Then the interlayer smoothness $\rho$ is defined to be the largest number such that over noise $\eta$, for two layers $i<j$, and any $x\in S$,
\begin{equation}
\vspace{-5pt}
    \lVert M^{i,j}(x^i+\eta)-J^{i,j}_{x^i}(x^i+\eta) \rVert \leq \frac{ \lVert \eta \rVert \lVert x^j \rVert}{\rho \lVert x^i \rVert}
\vspace{-1pt}
\label{eq:intersmoothness}
\end{equation}
\end{itemize}

$\rho$ measures the alignment of input and weight at layer $i$ versus that of noise and weight. A DNN is more prone to propagate noise in convolution if the noise vector is more aligned to weight matrix.

\section{Evaluating Adversarially Trained Networks on NVM Crossbars}
\label{sec:eval}
\begin{table*}[ht]
\centering
\caption{ResNet Architectures used for CIFAR-10/100}
\label{table:resnet}
\begin{tabular}{cccccc}
\toprule
& & \multicolumn{2}{c}{CIFAR-10} & \multicolumn{2}{c}{CIFAR-100}\\
\cmidrule{3-6}
Group Name & Output Size & ResNet10w1  & ResNet10w4 & ResNet20w1 & ResNet20w4 \\
\cmidrule(r){1-6}
conv0 & $32\times32$ & $3 \times 3, 16 $ &  $3 \times 3, 16 $ &  $3 \times 3, 16 $ &  $3 \times 3, 16 $\\
conv1 & $32\times32$ 
& $\left[\begin{array}{l}
    3 \times 3, 16 \\
    3 \times 3, 16
  \end{array}\right]\times 1$
& $\left[\begin{array}{l}
    3 \times 3, 64 \\
    3 \times 3, 64
  \end{array}\right]\times 1$ 
& $ \left[\begin{array}{l}
    3 \times 3, 16 \\
    3 \times 3, 16
  \end{array}\right] \times 3$
& $ \left[\begin{array}{l}
    3 \times 3, 64 \\
    3 \times 3, 64
  \end{array}\right] \times 3$ \\
\addlinespace
conv2 & $16\times16$ 
& $\left[\begin{array}{l}
    3 \times 3, 32 \\
    3 \times 3, 32
  \end{array}\right]\times 1$
& $\left[\begin{array}{l}
    3 \times 3, 128 \\
    3 \times 3, 128
  \end{array}\right]\times 1$
& $ \left[\begin{array}{l}
    3 \times 3, 32 \\
    3 \times 3, 32
  \end{array}\right] \times 3$
& $ \left[\begin{array}{l}
    3 \times 3, 128 \\
    3 \times 3, 128
  \end{array}\right] \times 3$  \\ 
\addlinespace  
conv3 & $8\times8$ 
& $\left[\begin{array}{l}
    3 \times 3, 64 \\
    3 \times 3, 64
  \end{array}\right]\times 2$
& $\left[\begin{array}{l}
    3 \times 3, 256 \\
    3 \times 3, 256
  \end{array}\right]\times 2$
& $ \left[\begin{array}{l}
    3 \times 3, 64 \\
    3 \times 3, 64
  \end{array}\right] \times 3$
& $ \left[\begin{array}{l}
    3 \times 3, 256 \\
    3 \times 3, 256
  \end{array}\right] \times 3$ \\ 
 & & Avg Pool & Avg Pool & Avg Pool & Avg Pool\\ 
Linear & $1 \times 1$ & $64 \times 10$ &  $256 \times 10$ &  $64 \times 100$ &  $256 \times 100$  \\ 
\bottomrule
\end{tabular}
\vspace{-0.3cm}
\label{tab:models}
\end{table*}

In this work, we implement adversarially trained DNNs on NVM crossbars based analog hardware and analyze their performance. We first study the effect on natural accuracy, i.e. their performance in the absence of any attack. Analog computations are non-ideal in nature, and the degree of performance degradation depends on the resilience of the DNN to internal perturbations at inference. The degree of performance degradation helps us estimate the noise stability of adversarially trained networks, and understand how they differ from regular networks. Next, we evaluate the performance of these DNNs under adversarial attack and identify the benefit offered by NVM crossbars. Our experimental setup is described in detail in this section. 

\subsection{Datasets and Network Models}

We perform our training and evaluation on two image recognition tasks: 
\begin{itemize}
    \item \textbf{CIFAR-10} \cite{krizhevsky2009learning}: It is a dataset containing $50,000$ training images, and $10,000$ test images, each of dimension $32\times32$ across $3$ RGB channels. There are total $10$ classes. We use two different network architectures for CIFAR-10, a $10$-layer ResNet \cite{he2016deep}, ResNet10w1 and a $4\times$ inflated version of it, Resnet10w4 (Table \ref{tab:models}).
    \item \textbf{CIFAR-100} \cite{krizhevsky2009learning}: This dataset also contains $50,000$ training images, and $10,000$ test images, each of dimension $32\times32$ across $3$ RGB channels, and there are total $100$ classes. We use two different network architectures for CIFAR-100, a $20$-layer ResNet \cite{he2016deep}, ResNet20w1 and a $4\times$ inflated version of it, ResNet20w4 (Table \ref{tab:models}).
\end{itemize}

\subsection{Adversarial Attacks}

We evaluate the DNNs against Non-Adaptive Attacks, where the attacker has no knowledge of the analog hardware, and assumes the DNN is implemented on accurate digital hardware. There exists a variety of attacks depending on how much information about the DNN is available to the attacker. The strongest category of attacks is White-Box Attacks, where the attacker has full knowledge of the DNN weights, inputs and outputs, and we chose this scenario to test our adversarially trained DNNs. We use Projected Gradient Descent (PGD) \cite{madry2018towards} to generate $l_{\infty}$ norm bound iterative perturbations,  per the following equation:
\begin{equation}
\label{eq:PGD}
    x^{t+1} = \Pi_{x+S}(x^{t} + \alpha sgn(\nabla_{x}L(\theta, x^{t}, y))
\end{equation}

The adversarial example generated at $(t+1)^{th}$ iteration is represented as $x^{t+1}$.  $L(\theta, x, y)$ is the DNN's cost function. It depends on the DNN parameters $\theta$, input $x$, and labels $y$. $S$ is the set of allowed perturbations, and is defined using the the attack epsilon ($\epsilon_{attack}$) as $S = \Big(\delta | \big(x + \delta \geq max(x + \epsilon_{attack}, 0)\big) \land \big(x + \delta \leq min(x + \epsilon_{attack}, 255)\big)\Big) $, where $x \in [0, 255]$. The input is an 8-bit image, and every pixel is a value between $0$ to $255$. For all our experiments, we set the number of iterations to 50.  The $\epsilon_{attack}$ is set within the range $[2,16]$ and signifies the maximum distortion that can be added to any pixel.

Another non-adaptive attack that does not rely on gradient information is the Square Black Box attack \cite{ACFH2020square}. The attacker queries the model on digital hardware for activation values of the last layer (logits), and then it performs a random search \cite{moonICML19} to select a localized square of pixels. Random perturbation bounded by $l_{\infty}$ norm is sampled. If the perturbation succeeds in increasing the loss, the image is updated. We set the sample size for the attack to 10000 and query limit to 1000 for Resnet10w1 and Resnet20w1, and sample size to 1000 and query limit to 500 for Resnet10w4 and Resnet20w4. The $\epsilon_{attack}$ is set within the range $[2,16]$.

\subsection{Adversarial Training}
First, we train our DNNs on the unperturbed dataset, i.e. "clean" images. We train two DNNs, ResNet10w1 and ResNet10w4 on CIFAR-10, and two DNNs, ResNet20w1 and ResNet20w4 on CIFAR-100. These $4$ DNNs are referred to as vanilla DNNs, and form the baseline for our later experiments. Next, we generate adversarially trained DNNs, by using iterative PGD training \cite{madry2018towards}.  At every training step, the batch of clean images is iterated over the network several times to generate a batch of adversarial images, which is then used to update the weights of the network. For a single training procedure, the epsilon of the adversarial image ($\epsilon_{train}$) is fixed, and the iterations is set to 50. For each network architecture in Table \ref{tab:models}, we generate 4 DNNs, each trained with a different $\epsilon_{train}$ (=[2,4,6,8]) for 200 epochs.

\subsection{Emulation of the Analog Hardware}
\label{subsec:emulation}
\begin{table}[h]
\centering
\caption{NVM Crossbar Model Description \cite{chakraborty2020geniex}}
\begin{tabular}{cccc}
\toprule
& \multicolumn{3}{c}{Crossbar parameters} \\ 
\cmidrule(r){2-4}
NVM Crossbar Model & Size   & $R_{ON}$ ($\Omega$)  & $NF$ \\
\cmidrule(r){1-4}
32$\times$32\_100k  & 32$\times$32    & 100k    &     0.14  \\ 
64$\times$64\_100k  & 64$\times$64    & 100k    &     0.26  \\
\bottomrule
\end{tabular}
\label{tab:params}
\vspace{-0.3cm}
\end{table}

\begin{table}[h!]
\centering
\caption{Functional Simulator precision parameters}
\label{tab:puma-param}
\begin{tabular}{ccc}
\toprule
Simulation & & \\ 
Parameters & CIFAR-10 & CIFAR-100 \\ 
\cmidrule(r){1-3}
$I_w$       & 4 &    4  \\
$W_w$       & 4  &   4  \\
$I_{bit}$   & 16 &  16  \\
$W_{bit}$   & 16 &  16  \\
$I_{i-bit}$ & 13 &  12  \\
$W_{i-bit}$ & 13 &  12  \\
$O_{bit}$   & 32 &  32  \\ 
$ADC_{bit}$ & 14 & 14 \\
$Acm_{bit}$ & 32 & 32 \\
$Acm_{f-bit}$ & 24 & 24 \\
\bottomrule
\end{tabular}
\vspace{-0.3cm}
\end{table}

In order to evaluate the performance of our adversarially trained DNNs on NVM crossbars in the presence of non-idealities, we need to a) model Equation \ref{eq:non-ideal} to express the transfer characteristics of NVM crossbars and b) map the convolutional and fully-connected layers of the workload on a typical spatial NVM crossbar architecture such as PUMA \cite{ankit2019puma}. The modelling of Eq. \ref{eq:non-ideal} is performed by considering the GENIEx \cite{chakraborty2020geniex} crossbar modeling technique which uses a 2-layer perceptron network where the inputs to the network are concatenated $V$ and $G$ vectors and the outputs are the non-ideal current, $I_{ni}$. The perceptron network is trained using data obtained from HSPICE simulations of NVM crossbars considering the aforementioned non-idealities with different combinations of $V$ and $G$ vectors and matrices, respectively. 

We generate two crossbar models: 32x32\_100k and 64x64\_100k, the corresponding parameters given in Table \ref{tab:params}. 
The NVM device used in the crossbar was an RRAM device model from \cite{guan2012spice}. Each crossbar model has a Non-ideality Factor, defined as, 
\begin{equation}
   NF = Average\Big(\frac{Output_{ideal}- Output_{non-ideal}}{Output_{ideal}} \Big)
\end{equation}

The GENIEx network estimates the NF factor by forward propagating $V$ and $G$ inputs through the trained 2-layer perceptron: its prediction of $f_{R}(V,G)$ is used to calculate the average degree of deviation of $I_{non-ideal}$ to $I_{ideal}$. The performance is benchmarked against HSPICE simulation results for the same V and G input combinations.

In our experiments, we consider two crossbar models of different sizes. Higher crossbar size results in higher deviations in computations.

Next, we use a simulation framework, proposed in \cite{chakraborty2020geniex}, to map DNN layers on NVM crossbars. 
Afterwards, we integrate these NVM crossbar models with our PyTorch framework, using the PUMA functional simulator \cite{ankit2019puma, chakraborty2020geniex}, and map DNN layers (all convolutional and linear layers) on NVM crossbars. Such a mapping onto a spatial NVM crossbar architecture like PUMA \cite{ankit2019puma} consists of three segments: i) Lowering, ii) Tiling and iii) Bit Slicing. Lowering refers to dividing the convolutional or linear layer operation into individual MVM operation. Tiling involves distributing bigger MVM operations into smaller crossbar-sized MVM sub-operations. Finally, since each NVM device can only hold upto a limited number of discrete levels, weight-slicing is performed with multiple crossbars implementing the weight slices; the output of the crossbars are accumulated to obtain the MVM output. Note, to reduce noise, the input to the crossbars are also bit-streamed using one-bit DACs. We set all precision parameters, given in Table \ref{tab:puma-param}, for 16 bit fixed point operations, and ensure that for an ideal crossbar behaviour, there is negligible accuracy loss due to reduction in precision. Thus, in our experiments with NVM crossbars any change in performance is solely due to non-ideal behaviour. The bit slicing and fixed-point precision parameters are defined as:
\begin{itemize}
    \item Input Stream Width ($I_w$) - Bit width of input fragments after slicing.
    \item Weight Slice Width ($W_w$) - Bit width of weight fragments after slicing.
    \item Input precision ($I_{bit}$) - Fixed-point precision of the inputs. This is divided into integer bits ($I_{i-bit}$) and fractional bits ($I_{f-bit}$).
    \item Weight precision ($W_{bit}$) - This refers to fixed-point precision of the weights. This is divided into integer bits ($W_{i-bit}$) and fractional bits ($W_{f-bit}$).
    \item Output precision ($O_{bit}$) - Fixed-point precision of the outputs of MVM computations.
    \item ADC precision ($ADC_{bit}$) - The bit precision of the ADCs at the crossbar outputs.
    \item Accumulator Width precision ($Acm_{bit}$) - Fixed-point precision of the ADC accumulator. 
\end{itemize}
\vspace{-0.3cm}

\subsection{Gaussian Noise Injection}
First, we randomly sample 1000 images from both CIFAR-10 and CIFAR-100 datasets, and evaluate the pretrained DNNs on digital hardware of 32x32\_100k, and 64x64\_100k NVM crossbars. We observe the outputs post convolution at every layer of the DNN, and post linear layer. We take the difference between digital and analog outputs, i.e. $Z_{analog}-Z_{digital}$. Here, $Z$ is the output of a layer before the activation function. The mean and standard deviation (sample deviation) of the distribution of $Z_{analog}-Z_{digital}$ are computed. The non-zero mean and standard deviation of the difference between analog and digital implementation is due to the non-idealities of the NVM crossbars. We modify Resnet-10 and Resnet-20 architectures by injecting Gaussian noise $\mathcal{N} \sim \eta (\mu, \sigma)$ layers at the outputs of each convolutional layers (including residual convolutions) and the linear layer. We perform 2 sets of experiments by matching the mean and standard deviation of the Guassian noise with that of crossbar variations $Z_{analog}-Z_{digital}$:
\begin{itemize}[leftmargin=1em]
    \item Matching $\sigma$: a white Gaussian noise layer with zero-mean is inserted at post convolution/linear locations. At each layer, the parameter standard deviation is passed with the standard deviation of $Z_{analog}-Z_{digital}$. The Gaussian noise injected becomes $\mathcal{N} \sim \eta (0, \sigma(Z_{analog}-Z_{digital}))$.
    \item Matching $\mu$ and $\sigma$: a Gaussian noise layer is inserted at post convolution/linear locations. At each layer, mean and standard deviation of $Z_{analog}-Z_{digital}$ are both passed as parameters for the Gaussian noise. The Gaussian noise injected becomes $\mathcal{N} \sim \eta (\mu(Z_{analog}-Z_{digital}), \sigma(Z_{analog}-Z_{digital}))$.
\end{itemize}
After Gaussian noise injection layers are added to the architecture and activated in the forward functions of the pretrained DNNs, we evaluate their performance on digital hardware, while performing White-Box PGD attack ($\epsilon_{attack} \in [0,16]$) assuming noise-free conditions. The adversarial accuracy of the noise-injected DNNs and the noise-free DNNs are compared, and the change in adversarial robustness is noted. We cross-compare this robustness with adversarial robustness of adversarailly trained networks on NVM crossbars, to verify that crossbar implementation endows robustness to DNNs that cannot be reproduced by noise injection.

\section{Results}
\label{sec:results}

\begin{figure*}[h]
\centering
\sidesubfloat[]{\label{fig:dig_vs_an_c10_r10w1}\includegraphics[width=0.21\linewidth]{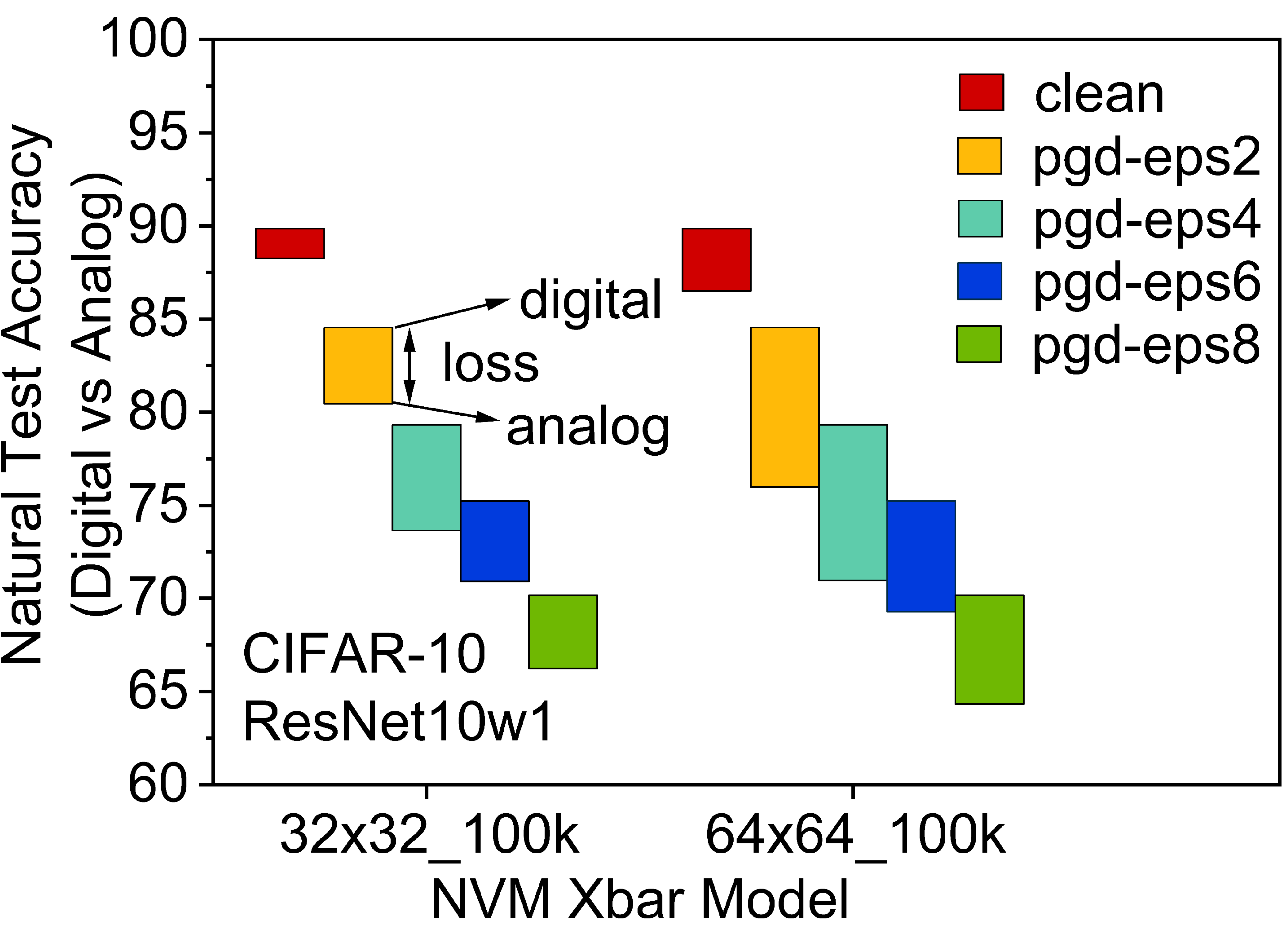}} 
\hfil
\sidesubfloat[]{\label{fig:dig_vs_an_c10_r10w4}\includegraphics[width=0.21\linewidth]{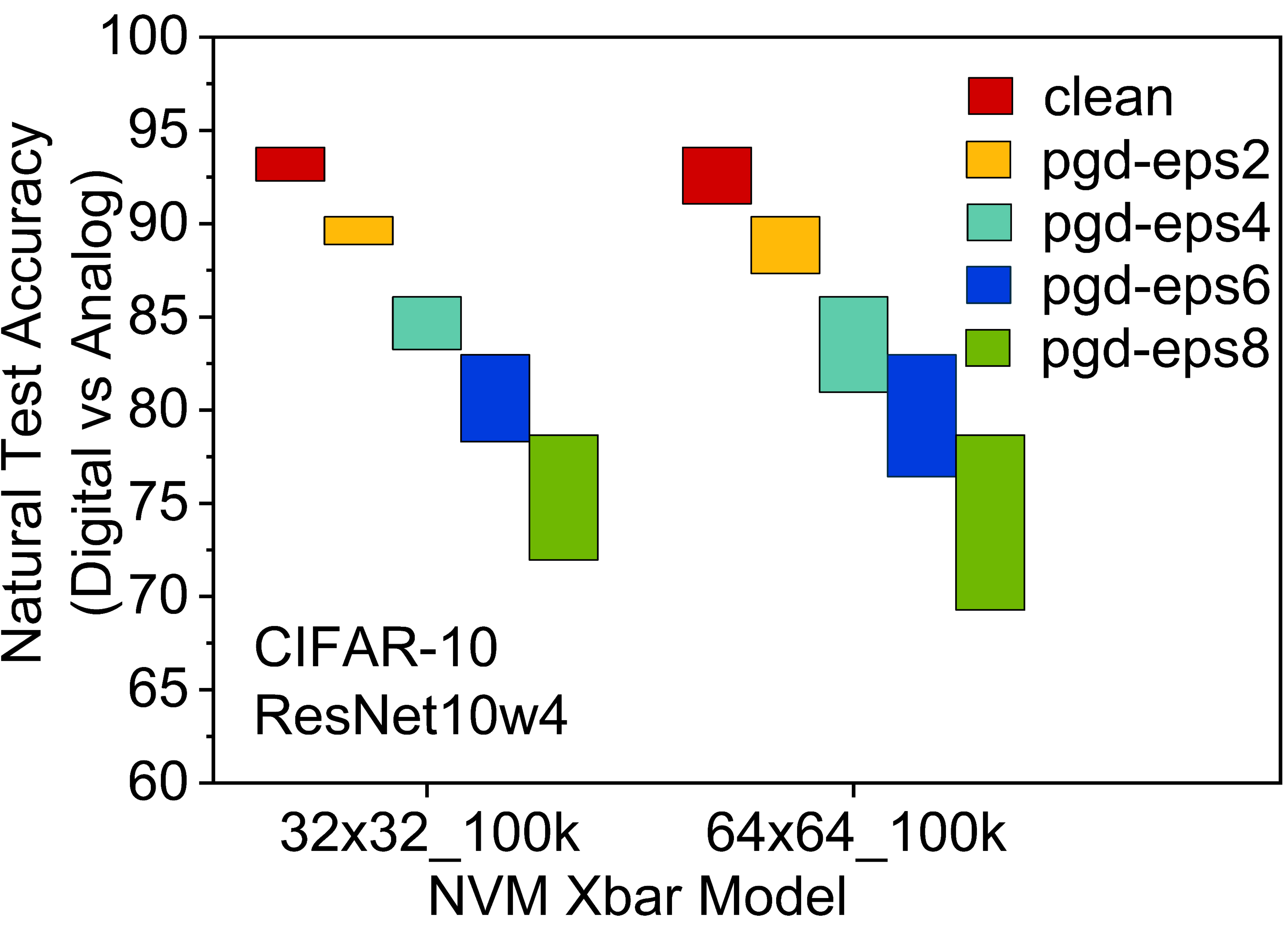}}
\hfil
\sidesubfloat[]{\label{fig:dig_vs_an_c100_r20w1}\includegraphics[width=0.21\linewidth]{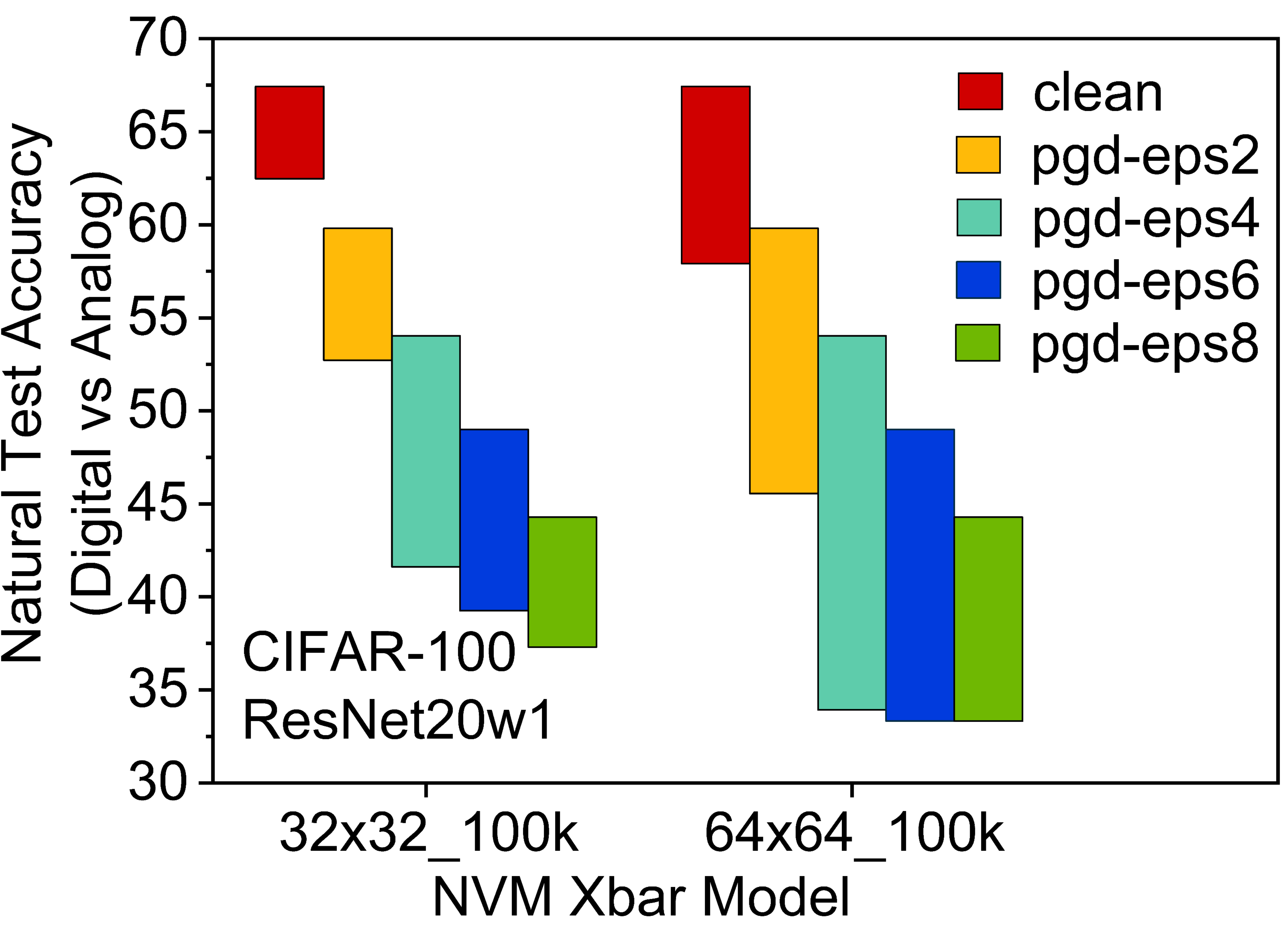}} 
\hfil
\sidesubfloat[]{\label{fig:dig_vs_an_c100_r20w4}\includegraphics[width=0.21\linewidth]{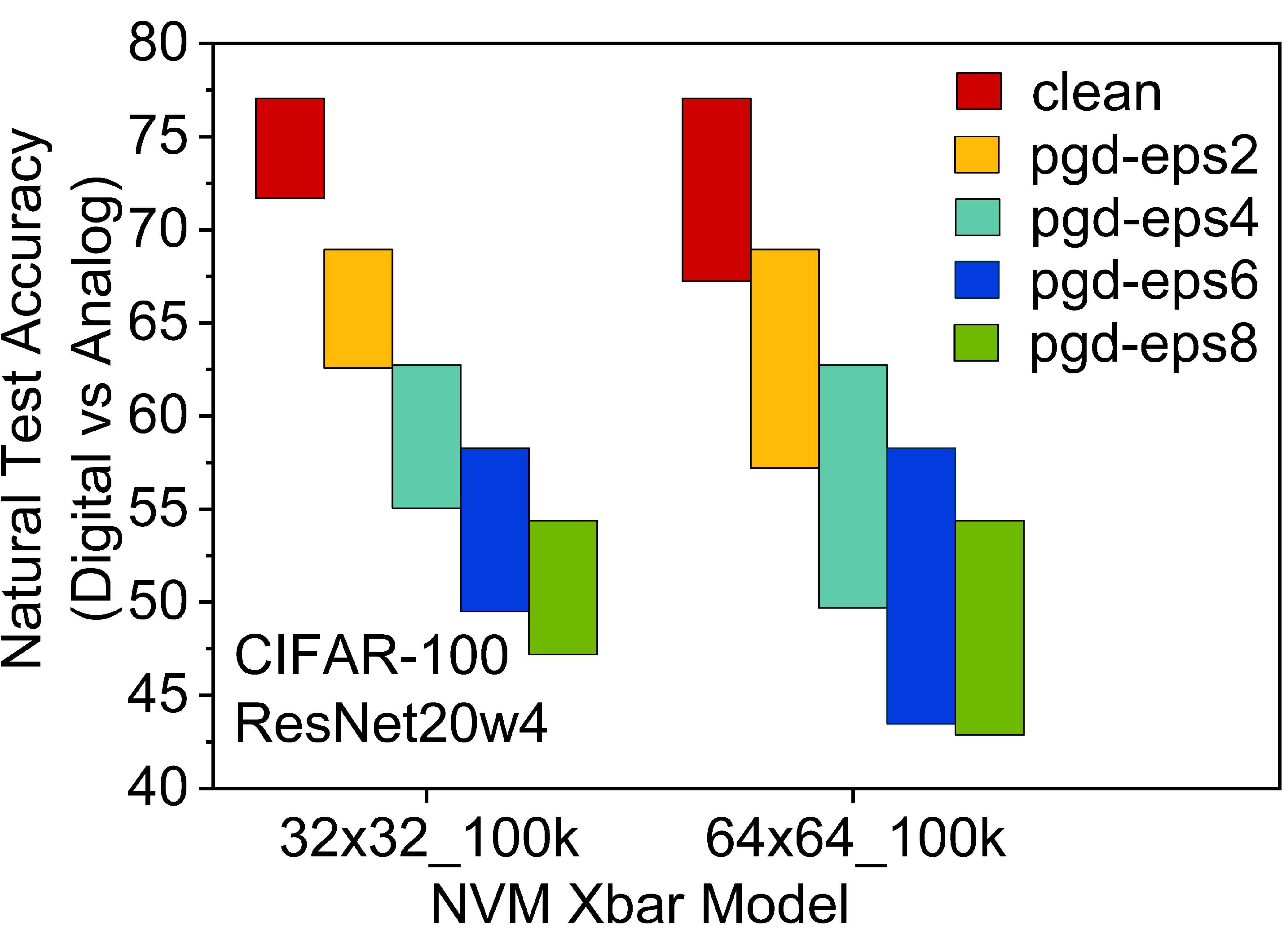}}
\vspace{-0.3cm}
\caption{Digital vs Analog Natural Test Accuracy for vanilla and adversarially trained DNNs. \textit{clean}: vanilla training with unperturbed images. \textit{pgd-epsN}: PGD adversarial training with $\epsilon_{train}= N = [2,4,6,8]$ and iter $=50$}
\label{fig:dig_vs_an}
\end{figure*}

\subsection{Noise Stability of Adversarially Trained Networks}
\begin{figure*}[h]
\centering
\sidesubfloat[]{\label{fig:snr64x64_100k_c10r10w1}\includegraphics[width=0.25\linewidth]{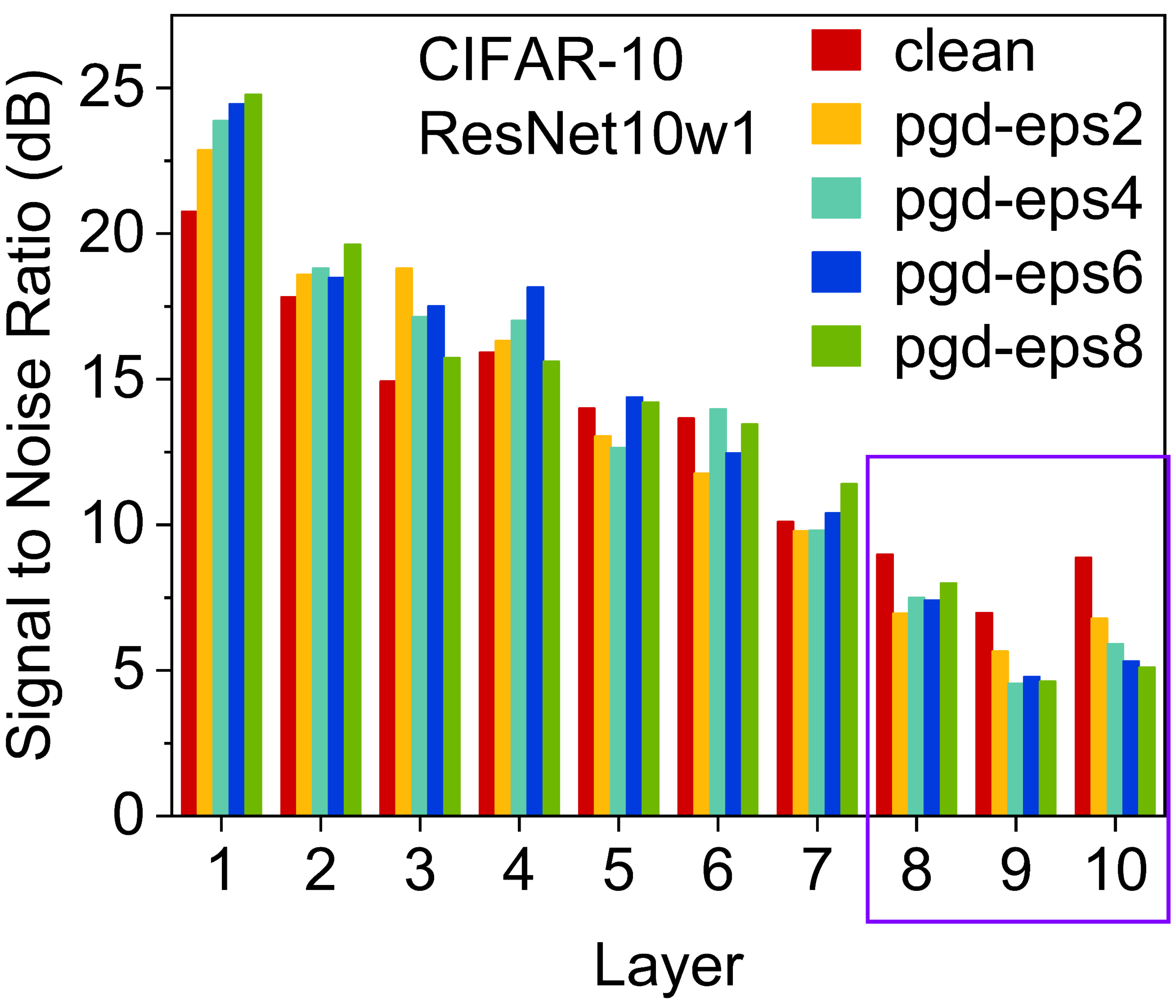}} 
\hfil
\sidesubfloat[]{\label{fig:snr64x64_100k_c10r10w4}\includegraphics[width=0.25\linewidth]{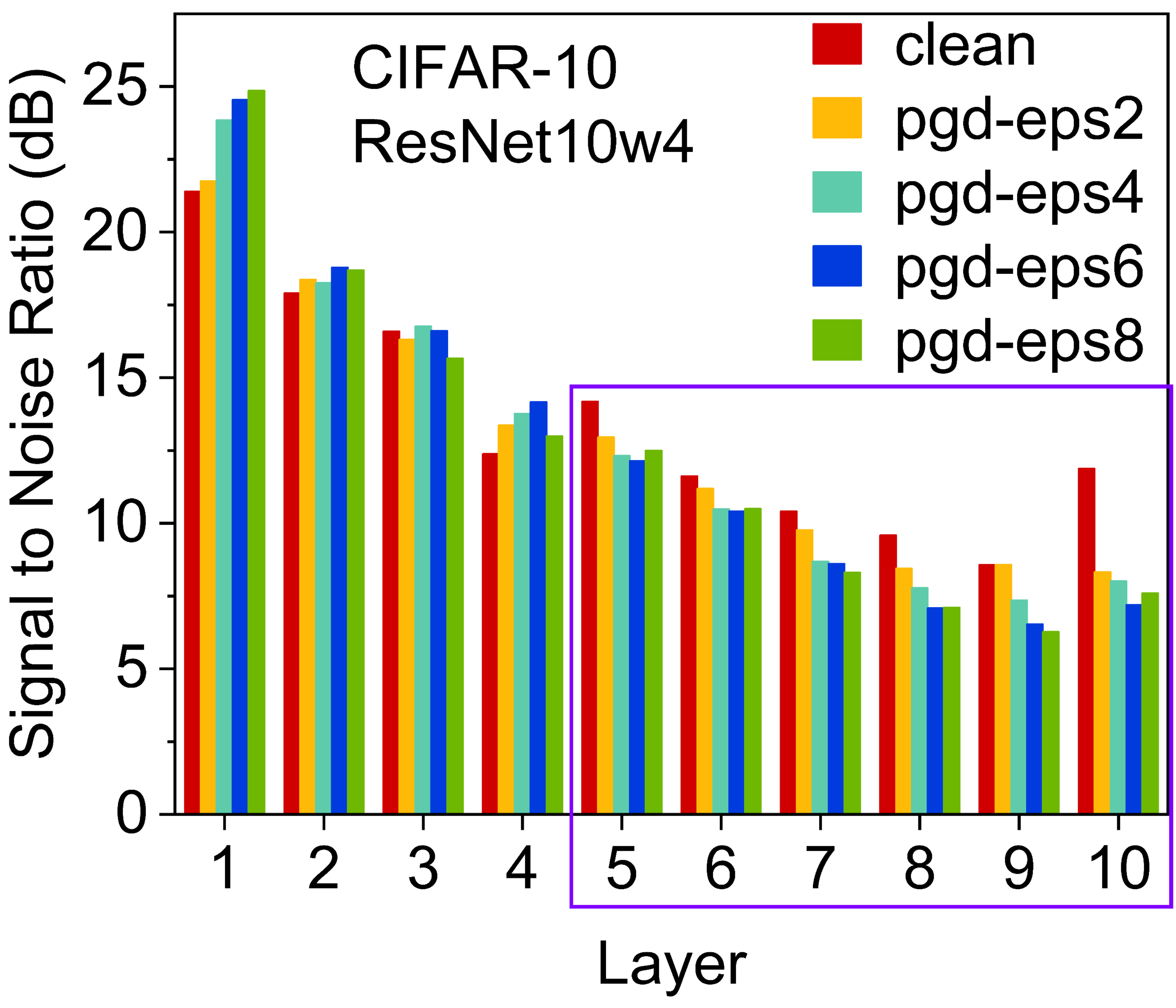}} \\
\hfil
\sidesubfloat[]{\label{fig:snr64x64_100k_c100r20w1}\includegraphics[width=0.43\linewidth]{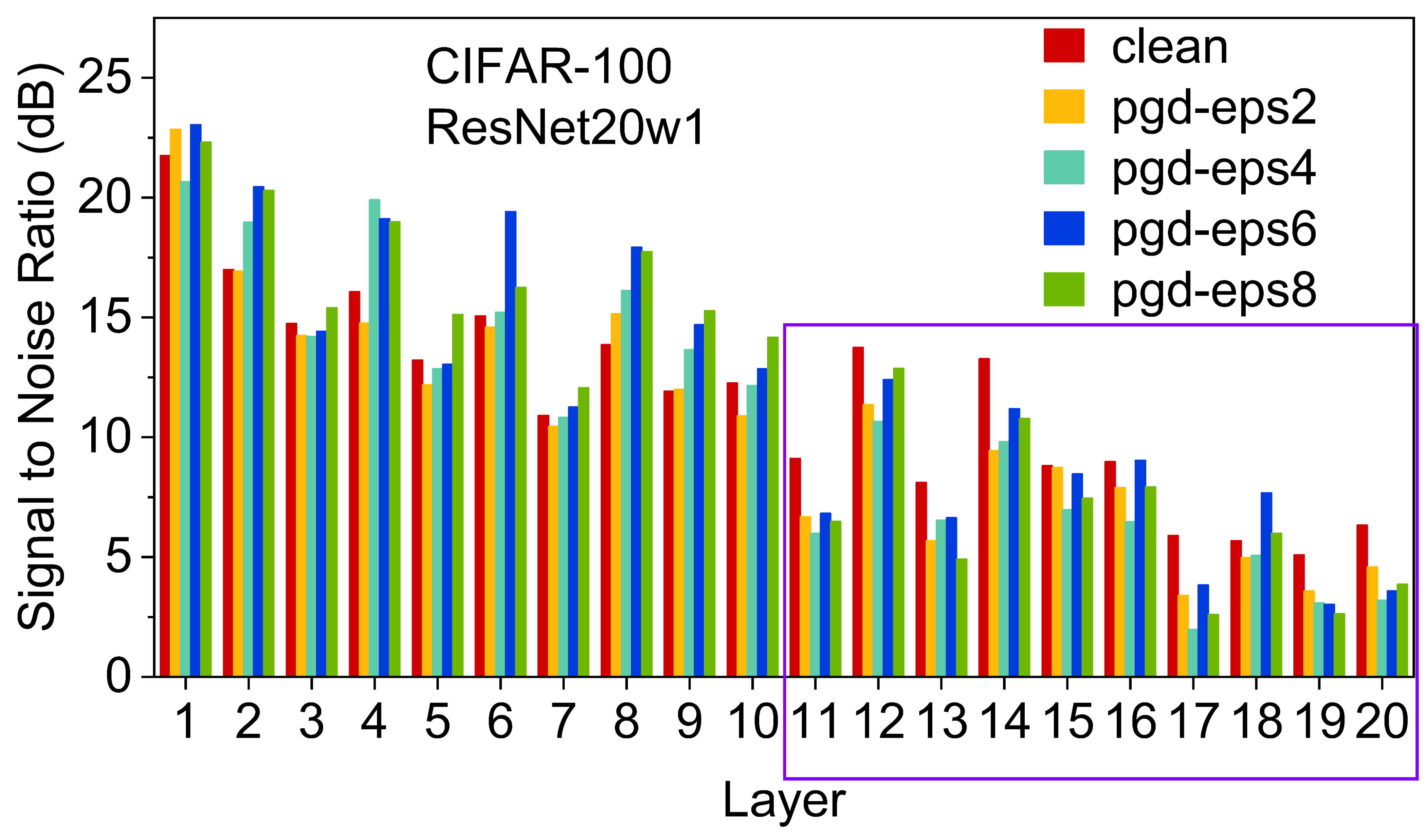}} 
\hfil
\sidesubfloat[]{\label{fig:snr64x64_100k_c100r20w4}\includegraphics[width=0.43\linewidth]{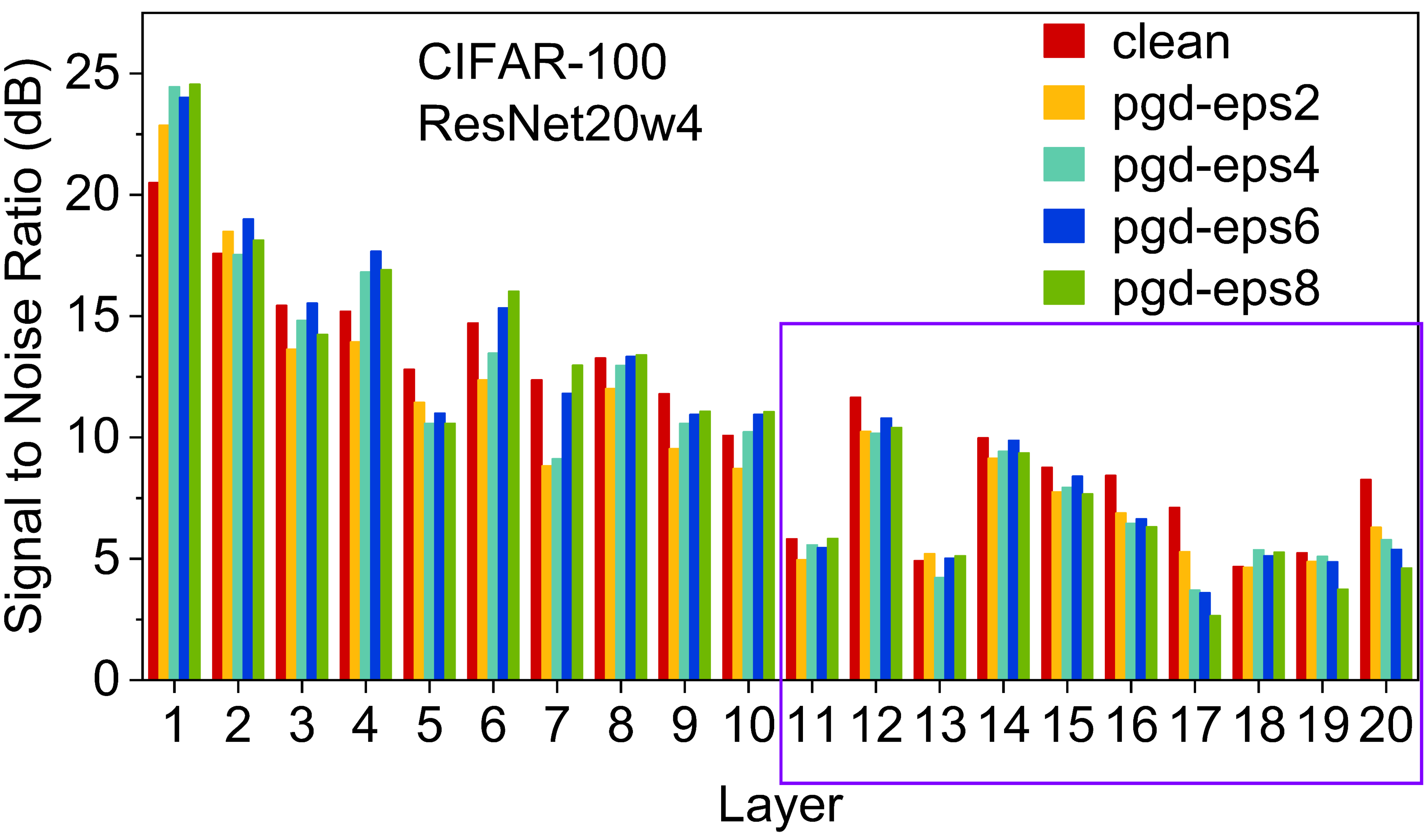}}
\vspace{-0.3cm}
\caption{Signal to Noise ($SNR$) at the output of every layer. for vanilla and adversarially trained DNNs. \textit{clean}: vanilla training with unperturbed images. \textit{pgd-epsN}: PGD adversarial training with $\epsilon_{train}= N = [2,4,6,8]$ and iter $=50$. NVM crossbar model: 64x64\_100k ($NF = 0.26$).}
\label{fig:snr}
\end{figure*}

\begin{figure}[h]
\centering
\sidesubfloat[]{\label{fig:ns64x64_100k_c10r10w1}\includegraphics[width=0.43\linewidth]{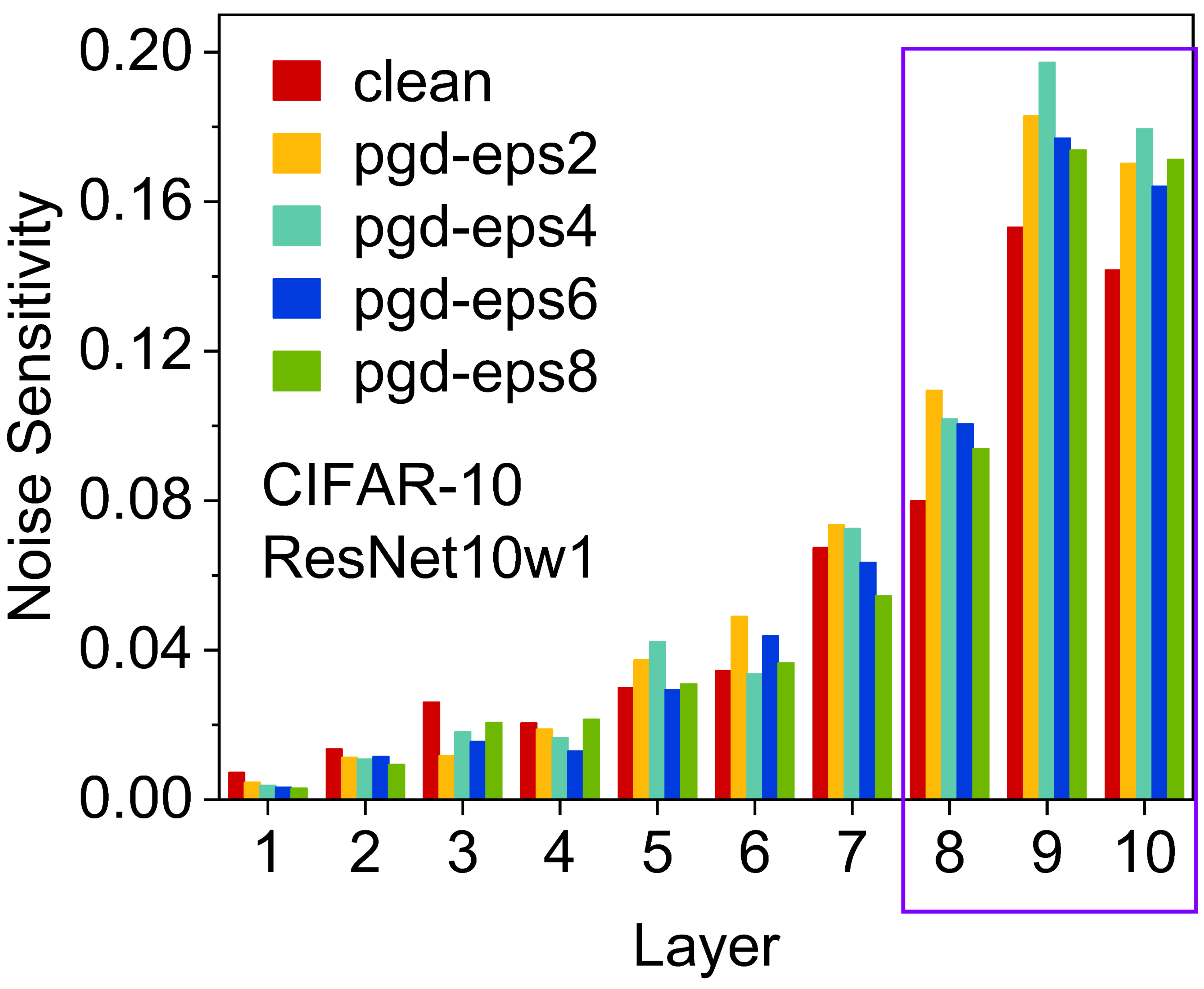}} 
\hfil
\sidesubfloat[]{\label{fig:ns64x64_100k_c10r10w4}\includegraphics[width=0.43\linewidth]{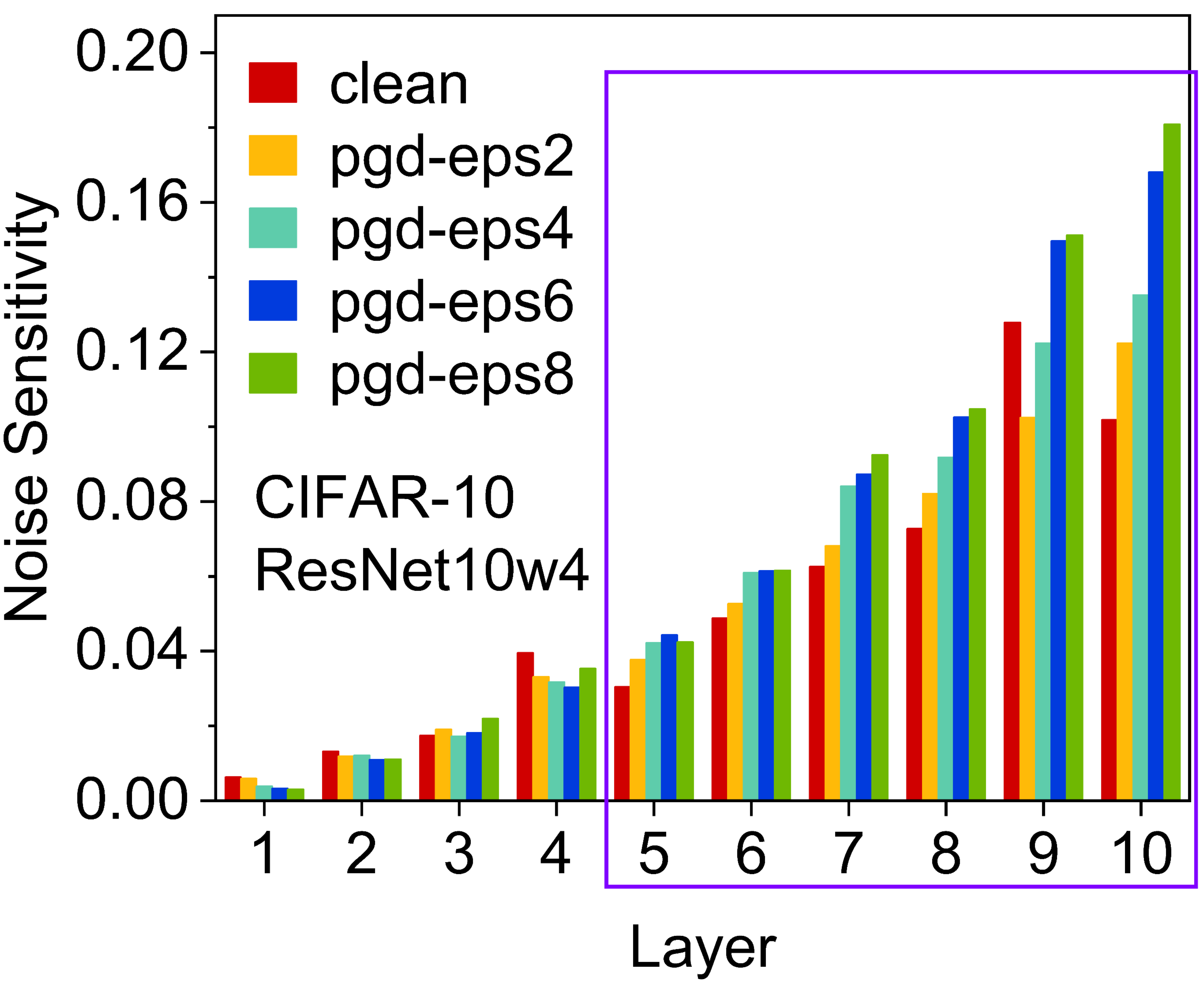}}
\caption{Noise Sensitivity at the output of every layer. for vanilla and adversarially trained DNNs. \textit{clean}: vanilla training with unperturbed images. \textit{pgd-epsN}: PGD adversarial training with $\epsilon_{train}= N = [2,4,6,8]$ and iter $=50$. NVM crossbar model: 64x64\_100k ($NF = 0.26$).\vspace{-0.3cm}}
\label{fig:ns}
\end{figure}

At first, we analyze the performance of the vanilla and adversarially trained DNNs on NVM crossbars in the absence of any adversarial attack. We define this accuracy as the Natural Test Accuracy of the DNNs. In Fig. \ref{fig:dig_vs_an}, we present a floating column chart where the top of a column represents the accuracy on accurate digital hardware, and where the bottom of the column represents the accuracy on the NVM hardware. The length of the column signifies the drop in accuracy due to non-idealities. Adversarial training, in itself, reduces the natural test accuracy of a DNN -- higher the training epsilon ($\epsilon_{train}$), lower is the test accuracy. When implemented on 2 different NVM crossbar models, 32x32\_100k and 64x64\_100k, the adversarially trained DNNs for all 4 network architectures suffer far greater accuracy degradation than their vanilla counterparts (i.e. DNNs trained on ``clean" images).  Also, Table \ref{tab:params} shows  32x32\_100k has lower non-ideality factor (NF) compared to 64x64\_100k, i.e. the smaller crossbar has lesser deviations than the larger crossbar. This translates into smaller accuracy drop for 32x32\_100k compared to 64x64\_100k.

On the NVM crossbar model, 32x32\_100k, the average accuracy drop for vanilla DNNs is 3.4\% with the maximum of 5.38\% for Resnet20w4 on CIFAR-100 dataset (Fig. \ref{fig:dig_vs_an_c100_r20w4}). Whereas for adversarially trained DNNs, the average is 6.2\%, and the maximum drop is 12.44\% for ResNet20w1($\epsilon_{train}=4$) on CIFAR-100 dataset (Fig. \ref{fig:dig_vs_an_c100_r20w1}).

Similarly, for NVM crossbar model, 64x64\_100k, the average accuracy drop for vanilla DNNs is 6.4\% with the maximum of 9.83\% for Resnet20w4 on CIFAR-100 dataset (Fig. \ref{fig:dig_vs_an_c100_r20w4}). Whereas for adversarially trained DNNs, the average is 10.3\%, and the maximum drop is 20.12\% for ResNet20w1($\epsilon_{train}=4$) on CIFAR-100 dataset (Fig. \ref{fig:dig_vs_an_c100_r20w1}).

Thus, on an average, adversarially trained DNNs suffer $2\times$ as much performance degradation on NVM crossbar, when compared to vanilla DNNs. To further investigate the cause of this accuracy drop, we look at the output post convolution at every layer of the DNN. To quantify the deviation from ideal, we define two new metrics. The first is Signal to Noise Ratio ($SNR$), and it is defined as, 
\begin{equation}
\label{eq:snr}
    SNR = log_{10}\Big(\sum_{N}\frac{|Z_{analog}|^{2}}{|Z_{digital} - Z_{analog}|^{2}}\Big)
\end{equation}
Here, $Z$ is the output of a layer before the activation function, i.e. the output right after the MVM operation. The numerator signifies the total signal strength, while the denominator accounts for the noise in the signal, which is the difference between the digital and NVM crossbar implementation.  

The second metric is Noise Sensitivity, $NS$, and uses the same definition as in \cite{arora2018stronger},
\begin{equation}
\label{eq:ns}
    NS= \sum_{N}\frac{|Z_{digital} - Z_{analog}|^{2}}{|Z_{digital}|^{2}}
\end{equation}

To computer $SNR$ and $NS$ we randomly sampled 1000 ($N$) images -- $10\%$ of the total test set for both CIFAR-10 and CIFAR-100, and evaluated the DNNs on 64x64\_100k NVM crossbar model as it has the higher non-ideality factor. In Fig. \ref{fig:snr}, we observe that for later layers, $SNR$ of adversarially trained DNNs is lower than that of vanilla networks. In case of CIFAR-10, layers 8-10 of ResNet10w1, and layers 5-10 of ResNet10w4 had lower $SNR$ for the adversarially trained versions compared to the vanilla version. Similarly, in case of CIFAR-100, layers 11-20 (except layer 18) of of both ResNet20w1 and ResNet20w4 have lower $SNR$ for the adversarially trained versions. Within a DNN, the later layers have greater impact on the classification scores, as they are the closest to the final linear classifier. The lower $SNR$ indicates that the noise introduced by the NVM non-idealities distorts the MVM outputs for adversarially trained DNNs by a larger margin as compared to the vanilla DNNs. Similarly, in Fig. \ref{fig:ns}, we observe that the noise sensitivity, $NS$, of the corresponding later layers is higher for adversarially trained networks, correlating with the trends of $SNR$ and accuracy.

Our observations are in line with theoretical and empirical studies of adversarially trained DNNs. The process of adversarial training creates weight transformations that are less noise stable, i.e. given some perturbations at the layer input, it generates greater deviations in the layer output. The optimization landscape of adversarial loss has increased curvature and scattered gradients \cite{duesterwald2019exploring, liu2020loss}. The solutions, i.e DNN weights, achieved by the adversarial training often doesn't lie within a stable minima. If one assumes the non-ideal deviations in the NVM crossbar as changes in the weights, then for vanilla DNNs, with stable, flat minimas,  these changes have small effect on the loss, and the accuracy drop is low. However, for an adversarially trained DNN, the same degree of changes results in a greater change in loss, and hence greater performance degradation. Thus, this work provides a hardware-backed validation of the complex loss landscape of adversarial training.

To further investigate the cause of lower $SNR$ and higher $NS$ of the adversarially trained models, and to come up with theoretical validation to support the performance degradation on analog hardware during inference, we compute the four noise stability parameters that the authors formalized in \cite{arora2018stronger}.

\begin{figure}[!htb]
\centering
\sidesubfloat[]{\label{fig:lc_r10w1}\includegraphics[width=0.43\linewidth]{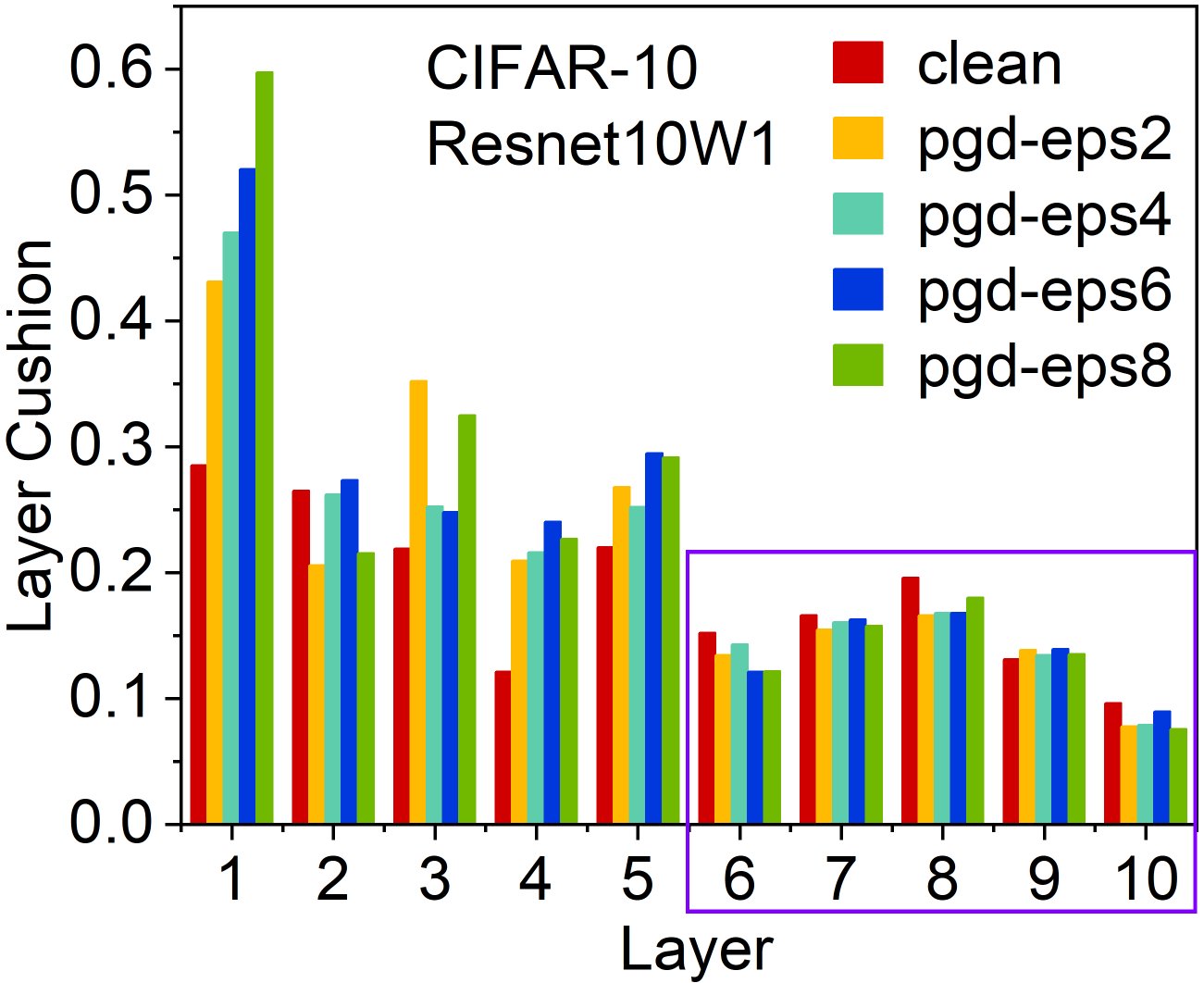}}
\hfil
\sidesubfloat[]{\label{fig:lc_r10w4}\includegraphics[width=0.43\linewidth]{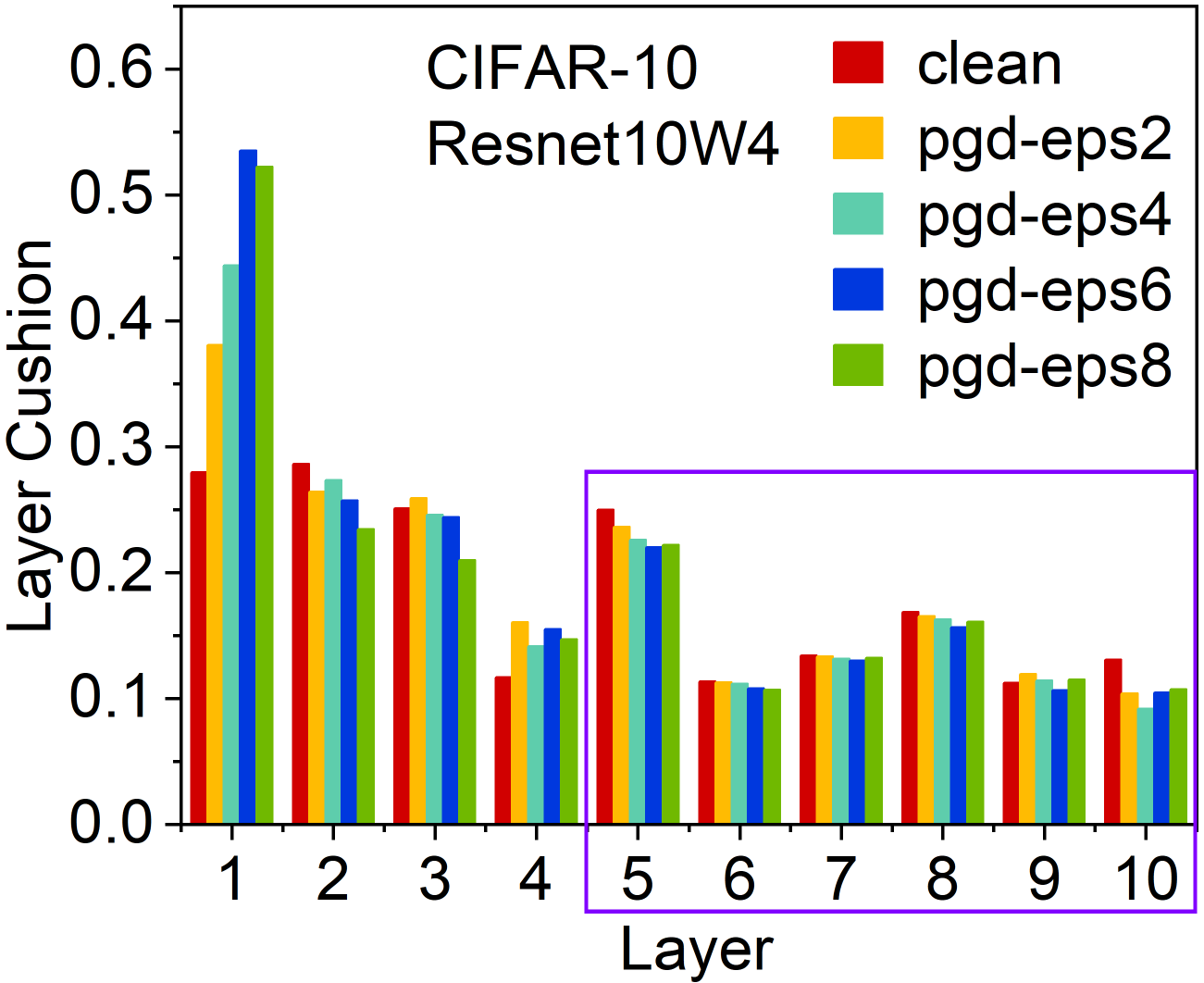}} \\
\sidesubfloat[]{\label{fig:ac_r10w1}\includegraphics[width=0.43\linewidth]{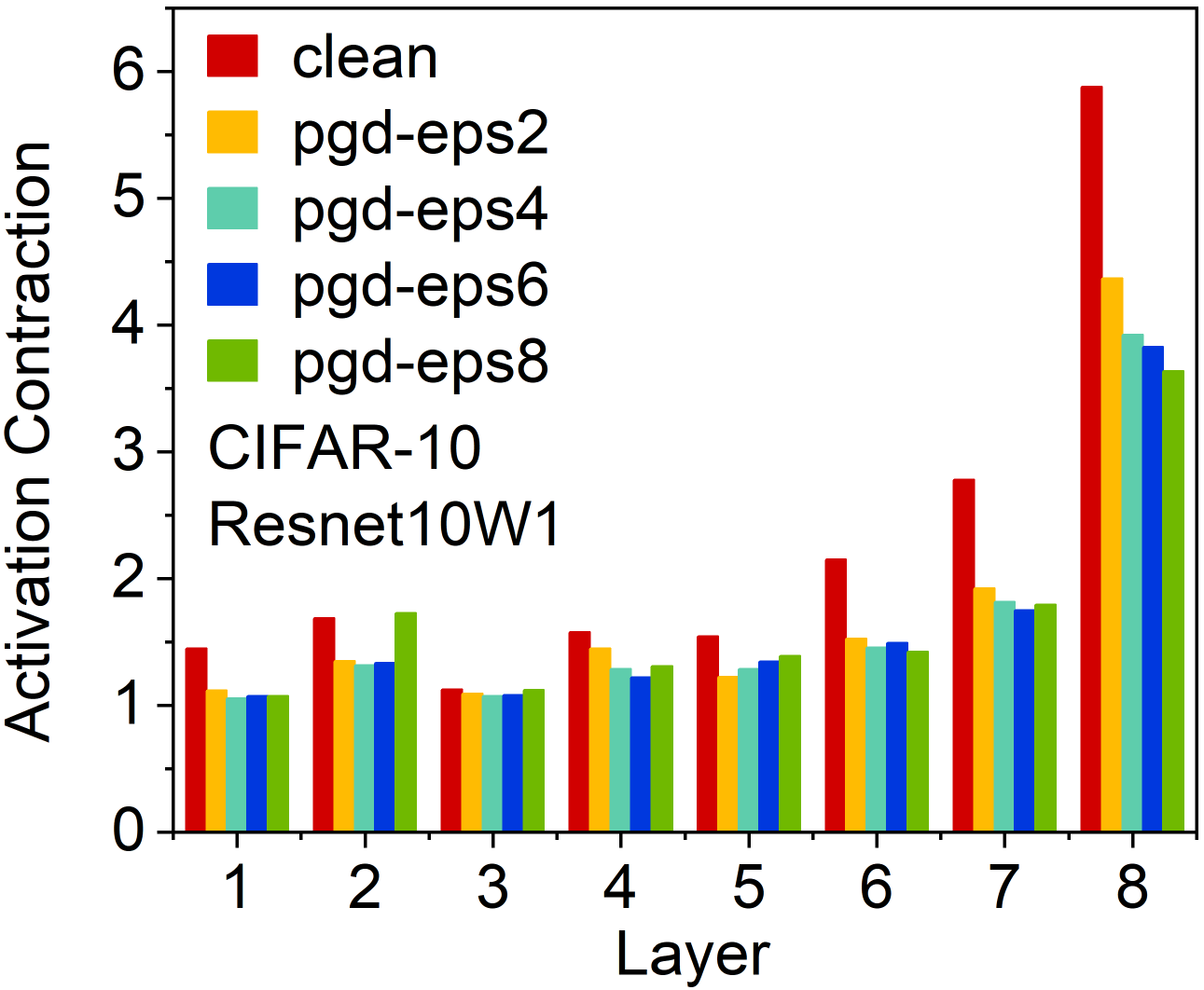}}
\hfil
\sidesubfloat[]{\label{fig:ac_r10w4}\includegraphics[width=0.43\linewidth]{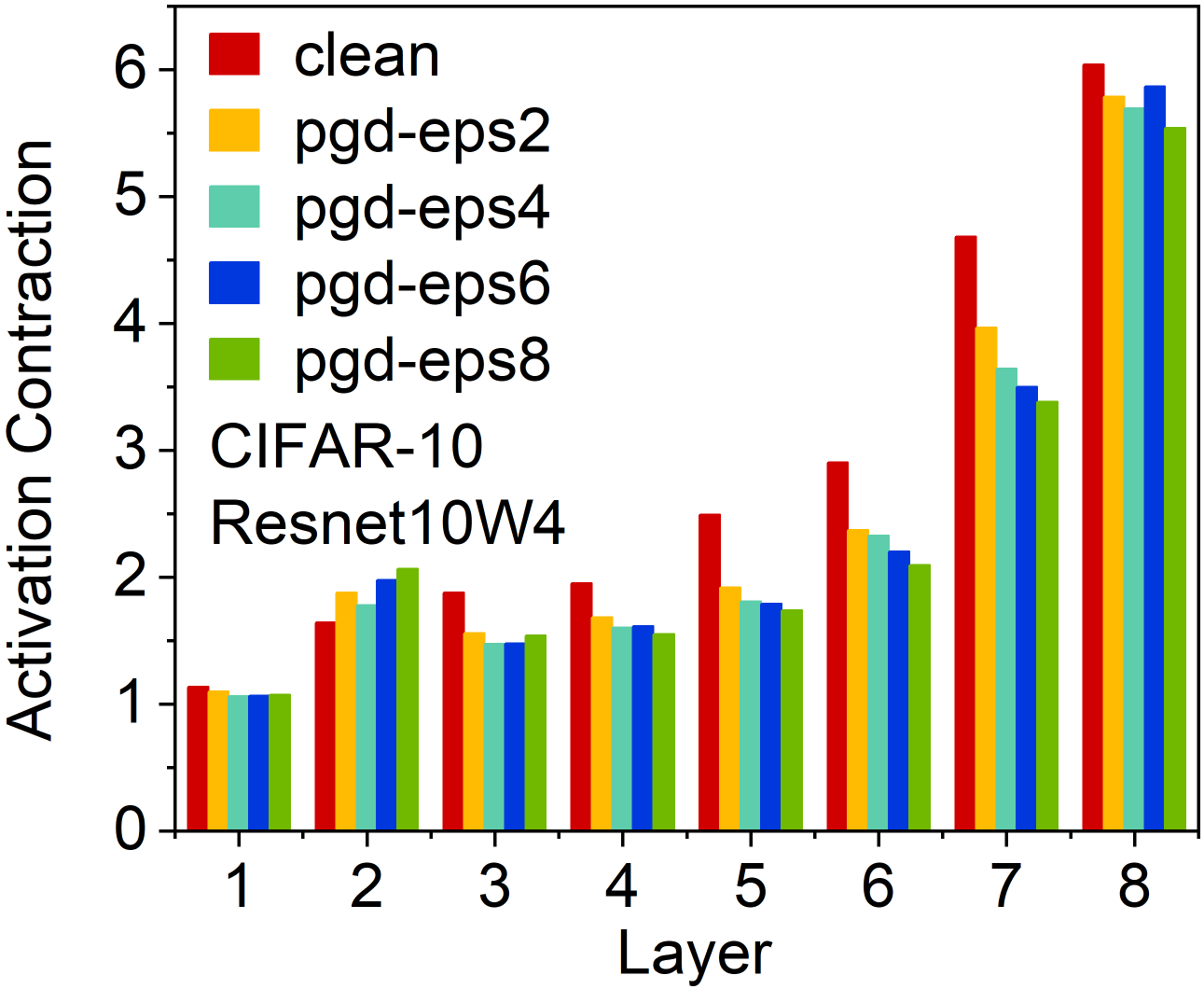}}
\caption{(a)(b) Layer Cushion $\mu_i$ of layer $i$, (c)(d) Activation Contraction $c_i$ of layer $i$, for \textit{vanilla} and \textit{adversarially trained} DNNs. NVM crossbar model: None.}\vspace{-0.3cm}
\label{fig:lcac}
\end{figure}

We randomly sample 1000 images of both CIFAR-10 and CIFAR-100, and evaluated DNNs on accurate digital hardware. For each layer $i$, we computed the minimum layer cushion $\mu_i$ over the collection of images, i.e. largest $\mu_i$ such that $\forall x \in \mathcal{S}$, Eq. \ref{eq:lc} holds. In Fig. \ref{fig:lc_r10w1} and Fig. \ref{fig:lc_r10w4}, we find $\mu$ to be lower for the adversarially trained models than vanilla models in layers $6-10$ of Resnet10w1 and layers $5-10$ of Resnet10w4. From Eq. \ref{eq:lc} this intuitively means that the vector norm of outputs of convolution layers of adversarially trained versions of DNNs have lesser cushion toward hitting an upper bound, $\lVert A^i \rVert _{F} \lVert \phi(x^{i-1}) \rVert$. The adversarially trained models have smaller margins and are more sensitive to noise in MVM operations, especially in the latter convolutional layers that are more crucial for classification.

We again randomly sample 1000 images from CIFAR-10 and CIFAR-100 testset, and evaluate DNNs on digital hardware. For each layer $i$, we computed the maximum activation contraction $c_i$ over 1000 images, i.e. largest $c_i$ such that $\forall x \in \mathcal{S}$, Eq. \ref{eq:ac} holds. We know from Eq. \ref{eq:ac} that $c_i$ measures the ratio of pre-activation to post-activation vector norm for each ReLU layer. In Fig. \ref{fig:ac_r10w1} and Fig. \ref{fig:ac_r10w4} we observe $c_i$ for all ReLU layers 1-8 of Resnet10w1 and Resnet10w4 to be smaller for adversarially trained models than their vanilla counterparts, indicating that the activations in adversarially trained DNNs are not as contractive as those in vanilla DNNs. A large ratio of the noise arising from non-idealites in preceding MVM operations pertains until after $i^{th}$ RELU activation, and thus the deviations in activation values aggregate as the noises propagate through the network. This result is in line with the accuracy degradation observed with adversarially trained DNNs during natural inference on NVM crossbars, as shown in Fig. \ref{fig:dig_vs_an}.

\begin{figure}[!htb]
\centering
\sidesubfloat[]{\label{fig:interlc_r20w1_1819}\includegraphics[width=0.43\linewidth]{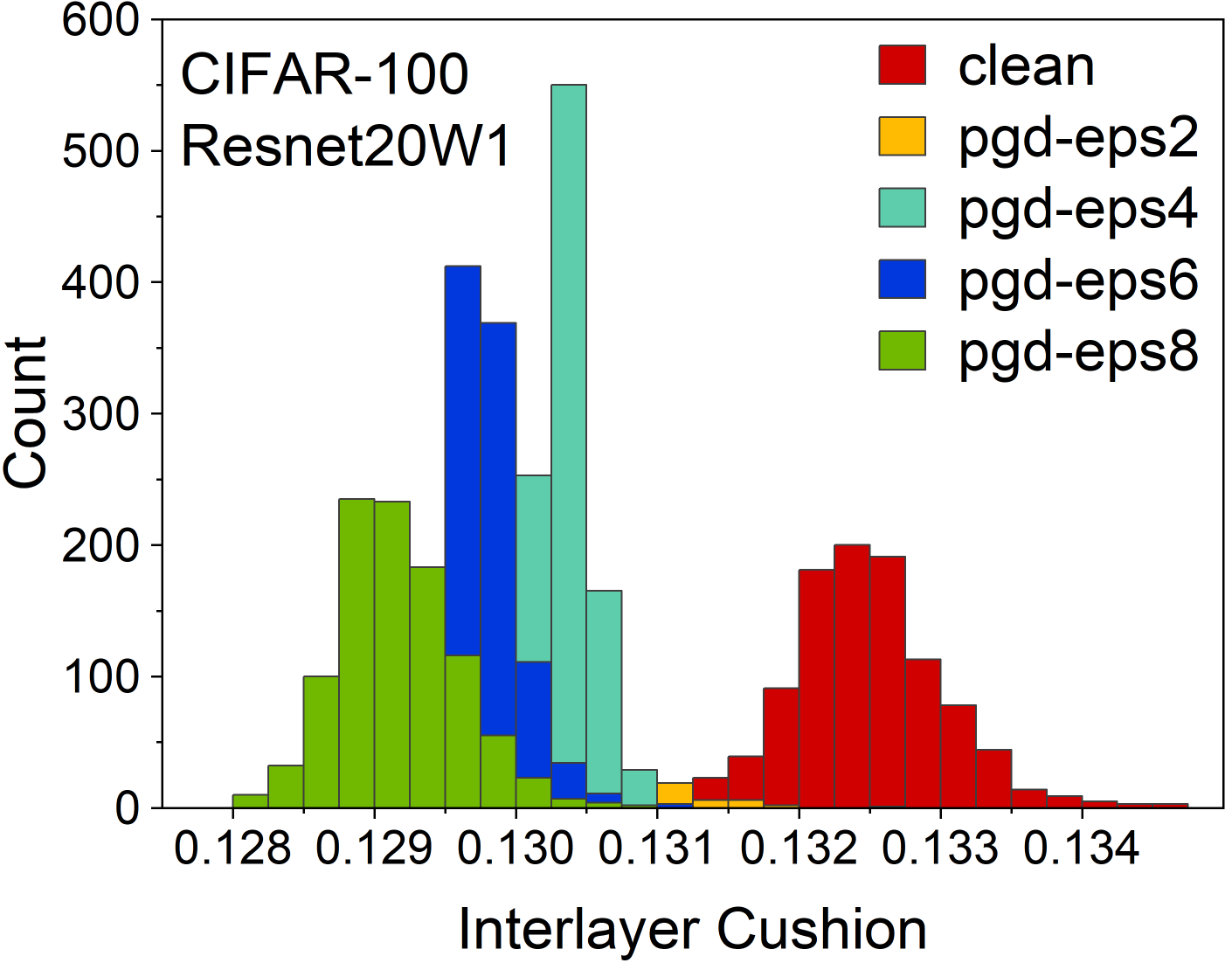}}
\hfil
\sidesubfloat[]{\label{fig:intersmooth_r20w1_1819}\includegraphics[width=0.43\linewidth]{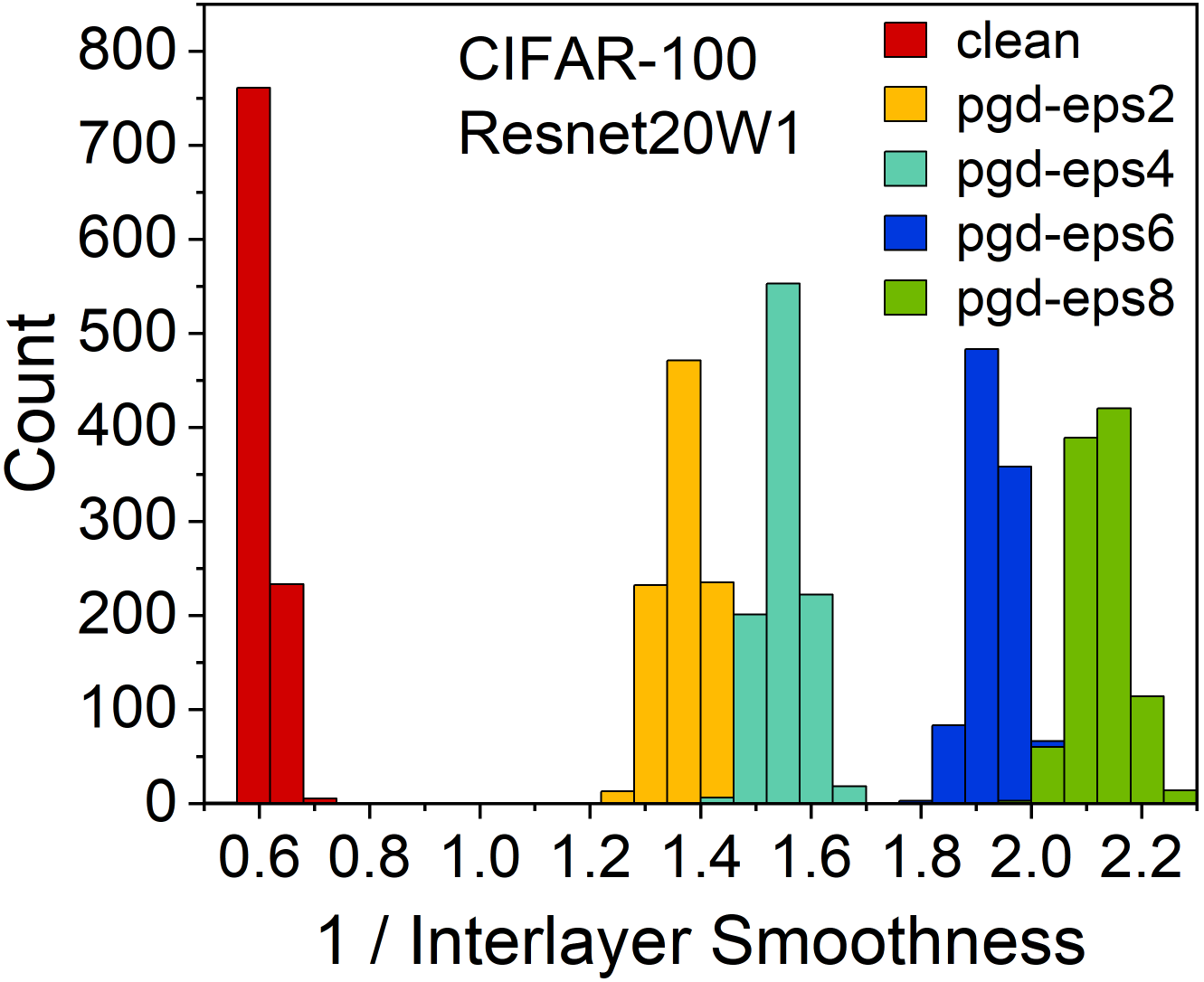}}
\caption{(a) Interlayer Cushion $\mu_{18,19}$ (b) 1/Intersmoothness $1/\sigma$ between layer 18,19, for \textit{vanilla} and \textit{PGD, $\epsilon_{train} = [2,4,6,8]$} DNNs. Distribution over 1000 samples. NVM crossbar model: None.}\vspace{-0.3cm}
\label{fig:interlc&intersmooth}
\end{figure}

We compute the interlayer cushion between convolutional layers 18 and 19, the last 2 conv layers of Resnet20w1. We plot a historgram of the distribution of $\mu_{18,19}$ values over the 1000 randomly drawn samples from CIFAR-100 testset. In Fig. \ref{fig:interlc_r20w1_1819} the distribution of $\mu_{18,19}$ for the vanilla model has a mean of $0.1325$, and a standard deviation of $0.0005$; while that for the model trained with PGD, $\epsilon_{train} = 4$ has a mean of $0.1304$, and a standard deviation of $0.0002$. Overall, the interlayer cushion between latter convolutional layers for adversarially trained DNNs is smaller than that for vanilla models. We conclude that after adversarial training, the final convolutional outputs are less stable because of accumulated noise during MVM operations in the preceding layers.

We then incorporate in Resnet10 and Resnet20 architectures a Gaussian layer with $\mu=0, \sigma_{relative}=0.1$ to each post-convolution vector $x^i$ at $i^{th}$ layer. The norm of the noise vectors is 0.1 times $\lVert x^i \rVert$. We use two forward functions to propagate to layer $j$, one with noise injected at each $x^i$ and another without. With the noisy forward function, we compute the Jacobian $J^{i,j}_{x^i}\left(x^i+\eta\right)$. $M^{i,j}_{x^i}\left(x^i+\eta\right)$ then becomes the post-convolution vector at layer $j$. Distribution of the reciprocal of $\rho$ for Resnet20w1, over 1000 samples of the CIFAR-100 test-set is plotted in Fig. \ref{fig:intersmooth_r20w1_1819}.

Interlayer smoothness  computes the ratio between input/weight and noise/weight alignments. From Fig. \ref{fig:intersmooth_r20w1_1819}, we find that the vanilla models have much greater interlayer smoothness between conv layers 18 and 19. This observation suggests that the adversarially-trained models "carry" the noise vectors more effectively through forward propagation in the last few convolutional layers. The noise vectors are more aligned to the weight matrices, while the input vectors at the convolutional layer are less aligned to the weight matrices \cite{arora2018stronger}. The information contained in the inputs propagated from previous convolutional layers is not propagated as efficiently as arbitrary noise that's injected at inputs. Because conv layers 18 and 19 are the last layers before the final linear classification layer, lower $\mu_{18,19}$ of adversarially trained models incur unstable model outputs. The smoothness value again validates that the adversarially-trained models are sensitive to analog hardware noise.

\subsection{Adversarial Robustness of Analog NVM crossbars}
\label{subsec:advrobust}
\begin{figure}[!htb]
\centering
\sidesubfloat[]{\label{fig:bb-robust-32-eps2}\includegraphics[width=0.43\linewidth]{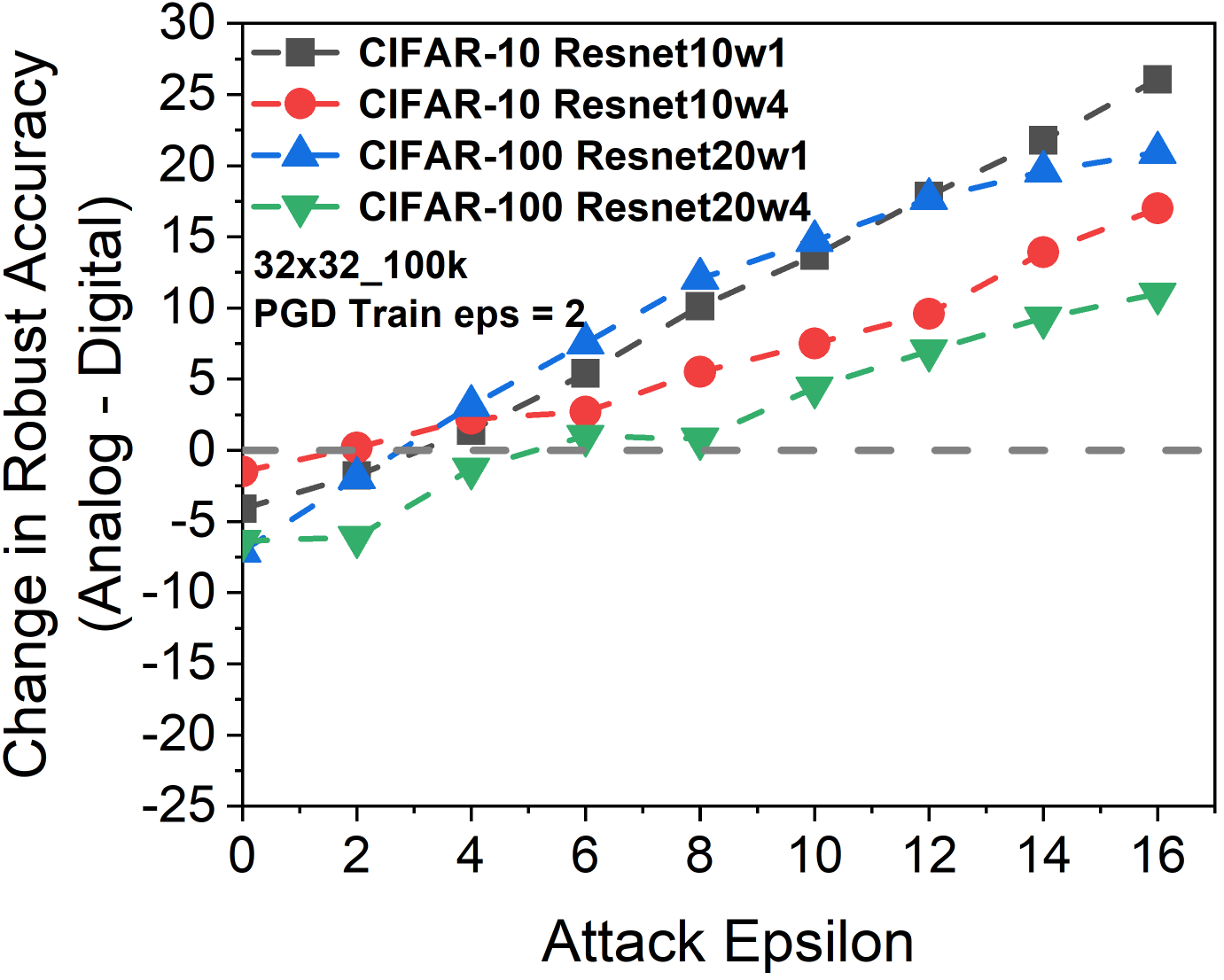}}
\hfil
\sidesubfloat[]{\label{fig:bb-robust-32-eps6}\includegraphics[width=0.43\linewidth]{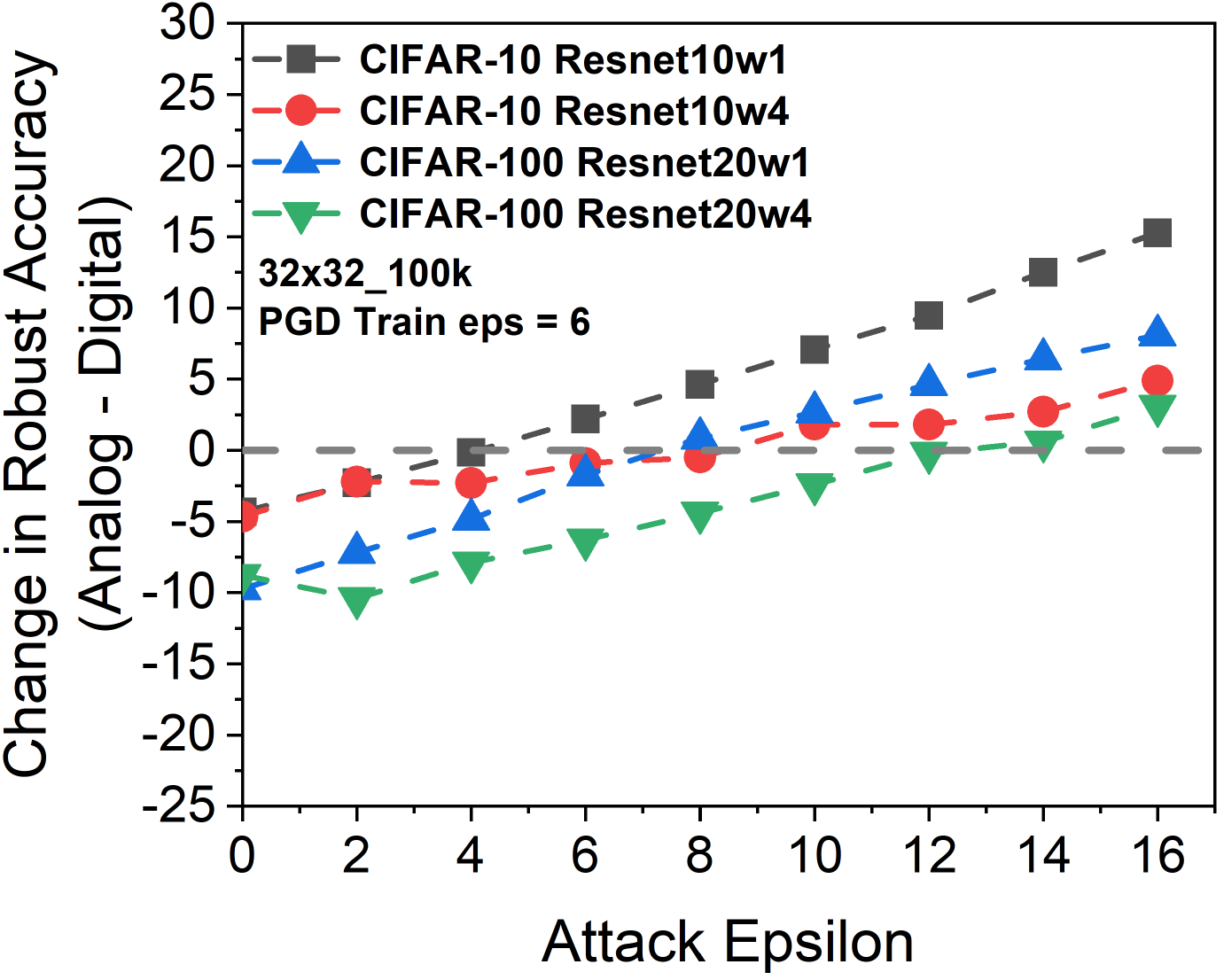}}\\
\sidesubfloat[]{\label{fig:bb-robust-64-eps2}\includegraphics[width=0.43\linewidth]{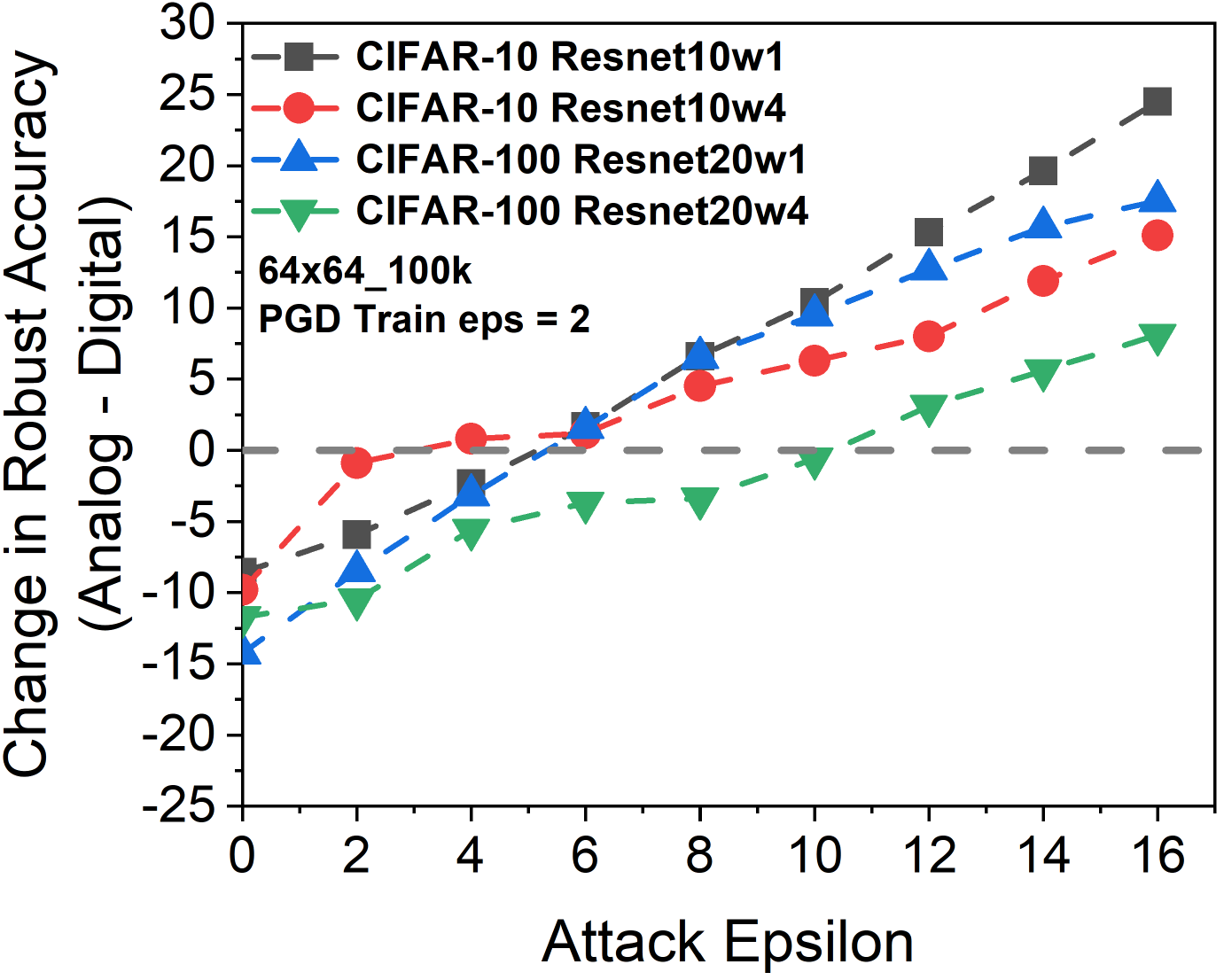}}
\hfil
\sidesubfloat[]{\label{fig:bb-robust-64-eps6}\includegraphics[width=0.43\linewidth]{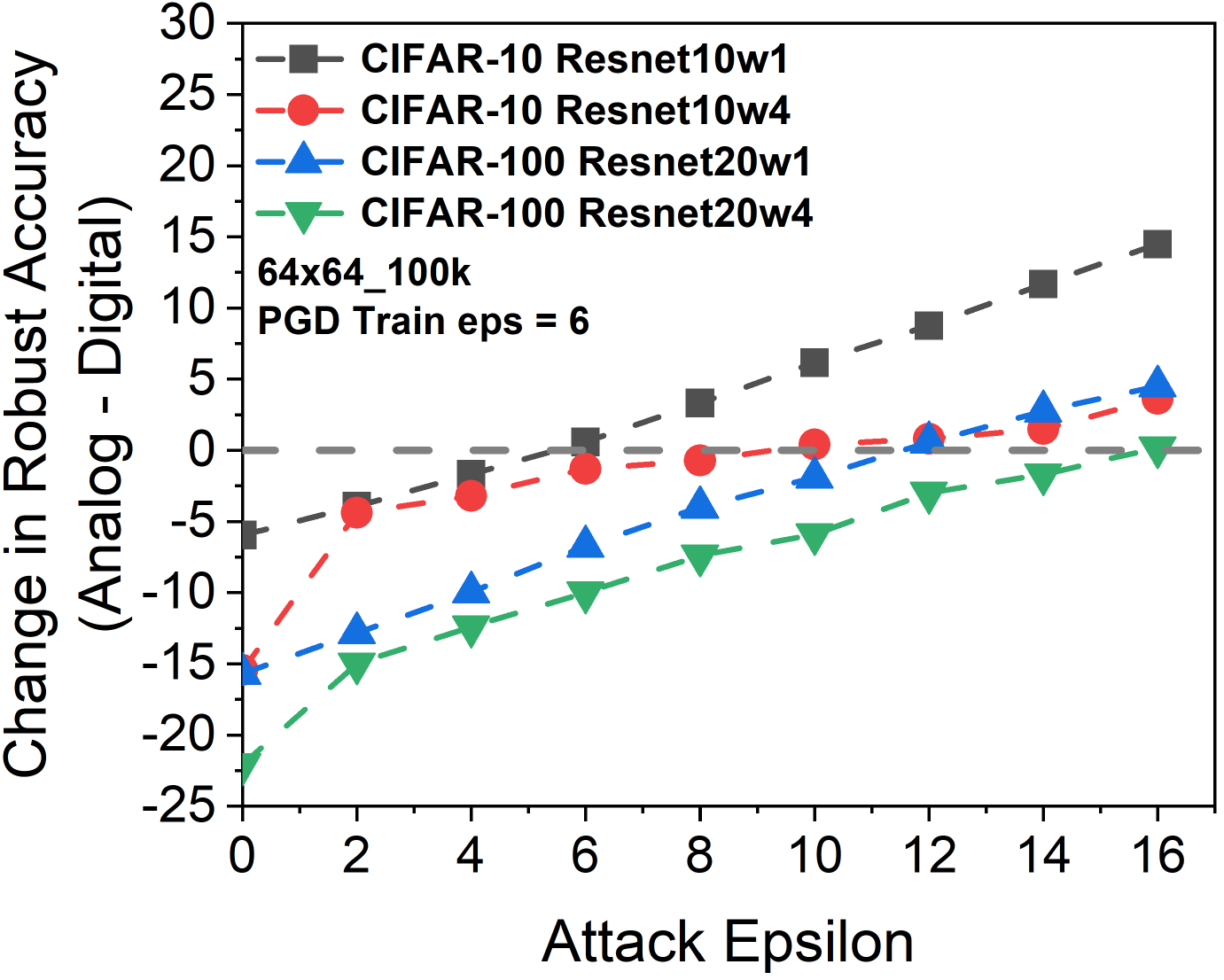}}
\caption{Difference in Adversarial Accuracy (Robustness Gain = Analog - Digital) for varying Square Black Box Attack $\epsilon_{attack}$. NVM crossbar model: (a)(b) 32x32\_100k ($NF = 0.14$), (c)(d) 64x64\_100k ($NF = 0.26$). 4 network architectures (ResNet10w1, ResNet10w4, ResNet20w1, ResNet20w4) on 2 datasets (CIFAR-10, CIFAR-100) are adversarially trained with (1) PGD, $\epsilon_{train} = 2$ and iter$ =50$ (2) PGD, $\epsilon_{train} = 6$ and iter$ =50$}
\label{fig:bb-robust-32-64}
\end{figure}

The query-efficient Black-Box square attack is non-adaptive, just like the PGD White-Box attack, possessing no information about the underlying hardware. Also, because it only queries the activation values, or weights, of the linear layer (logits), it's gradient-free in nature and doesn't rely on local gradient information. The "zero order" attack \cite{madry2018towards} is thereby guaranteed with stronger robustness, when the DNN is PGD adversarially trained. From Table \ref{table:nonadaptivetable}, we note that adversarially trained DNNs always have much better adversarial accuracy than vanilla DNNs, and thus adversarial training guarantees robustness. On top of that, implementation on NVM crossbars can potentially boost that robustness. In Fig. \ref{fig:bb-robust-32-64} the change in inference accuracy between digital and analog implementations is plotted. We observe the hardware non-idealities to demonstrate a varying degree of robustness to DNNs trained with PGD-eps2 and PGD-eps6, depending on the degree of $\epsilon_{attack}$. Adversarially trained DNNs on NVM crossbars suffer robustness loss against attacks with smaller $\epsilon_{attack}$, but benefit from robustness gain with stronger attacks with greater $\epsilon_{attack}$. The change in robustness increases monotonically with the degree of $\epsilon_{attack}$, up to $20-30\%$ at $\epsilon_{attack}=16/255$. In Fig. \ref{fig:bb-robust-32-eps2}, for $\epsilon_{attack} = [0,2]$ most of the DNNs evaluated suffer negative change in robust accuracy. When $\epsilon_{attack} \geq 4$, three of the four DNNs under evaluation show positive robustness. In Fig. \ref{fig:bb-robust-32-eps6}, the majority of DNNs show robustness for $\epsilon_{attack} \geq 10$. We believe that evaluation on NVM crossbars can provide DNNs trained with a smaller $\epsilon_{train}$ higher robustness. From Fig. \ref{fig:bb-robust-64-eps2} and Fig. \ref{fig:bb-robust-64-eps6}, we find the respective regions of robustness gain for the majority of DNNs to be $\epsilon_{attack} \geq 6$ and $\epsilon_{attack} \geq 12$. 32x32\_100k crossbars ensure higher robust accuracy compared to the 64x64\_100k counterpart. Thus, the choice of NVM crossbar model also determines the amount of additional robustness we can gain on top of PGD adversarially trained DNNs. The Black-Box square attack based on random search is a weaker form of adversary, and we have shown that by calibrating both $\epsilon_{train}$ in adversarial training, and the choice of analog hardware model, one can optimize the design of robust DNNs against high $\epsilon_{attack}$ Black-Box adversaries.

\begin{figure*}[h]
\centering
\sidesubfloat[]{\label{fig:r10w1-digital}\includegraphics[width=0.23\linewidth]{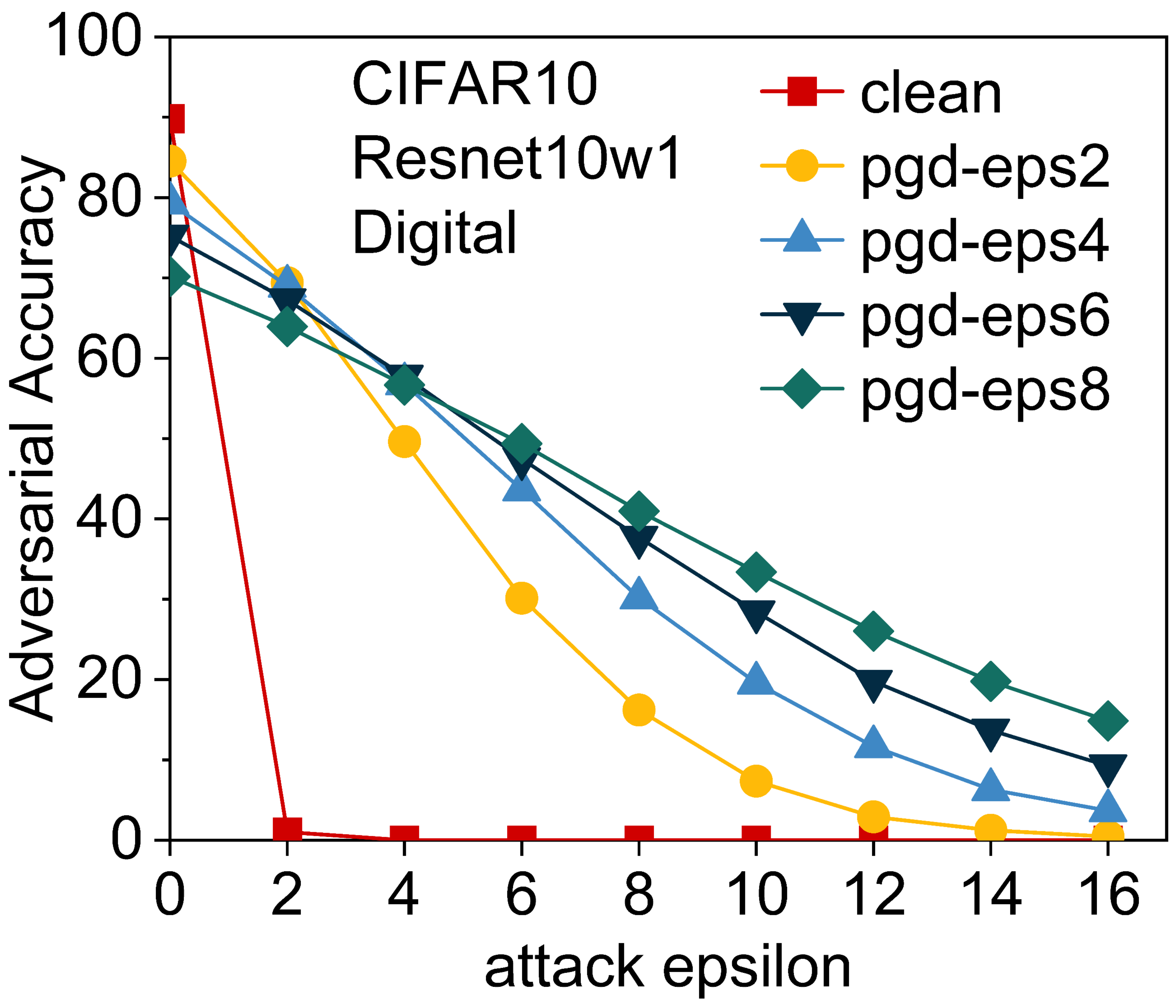}} 
\hfil
\sidesubfloat[]{\label{fig:r10w1-32x32-100k}\includegraphics[width=0.22\linewidth]{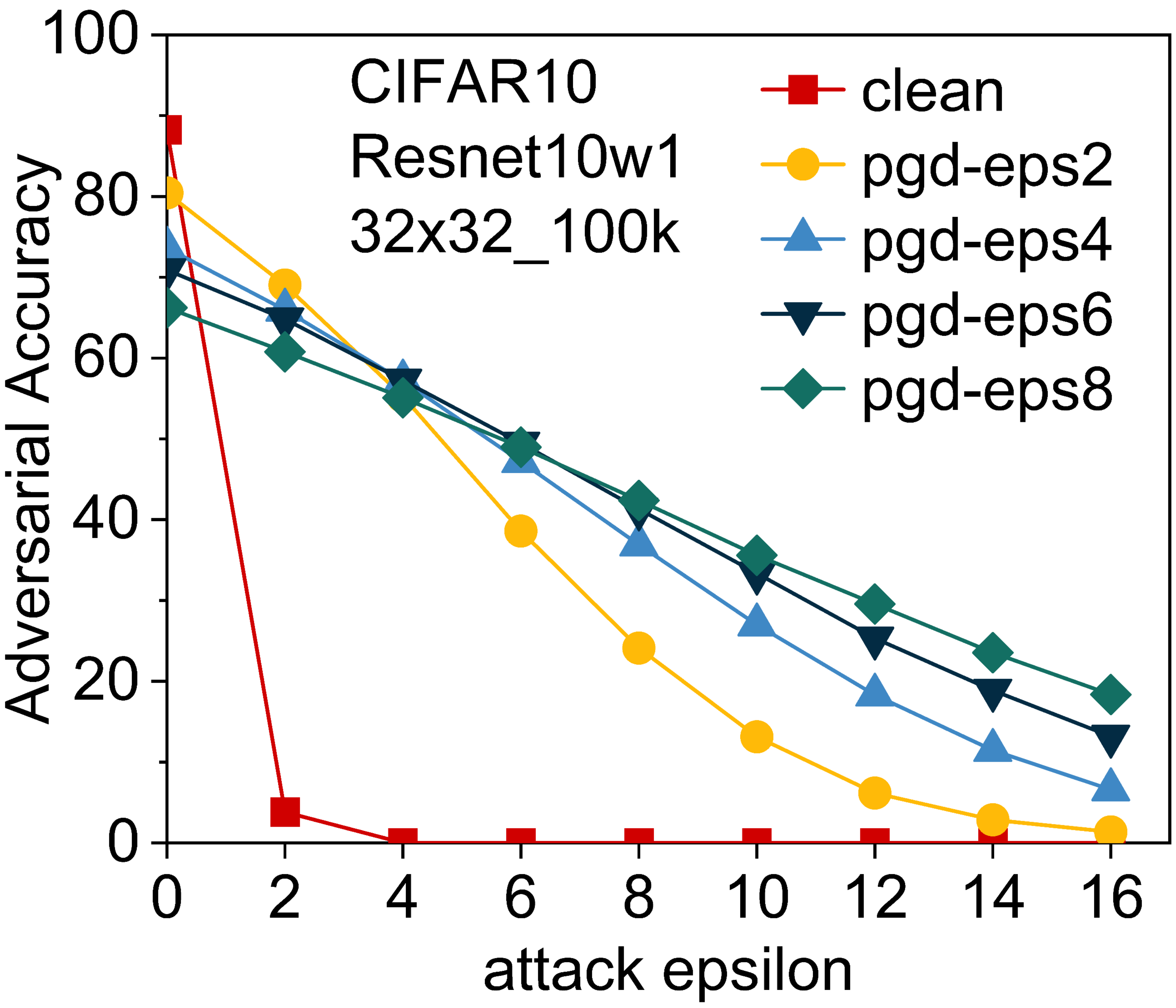}}
\hfil
\sidesubfloat[]{\label{fig:r10w1-64x64-100k}\includegraphics[width=0.22\linewidth]{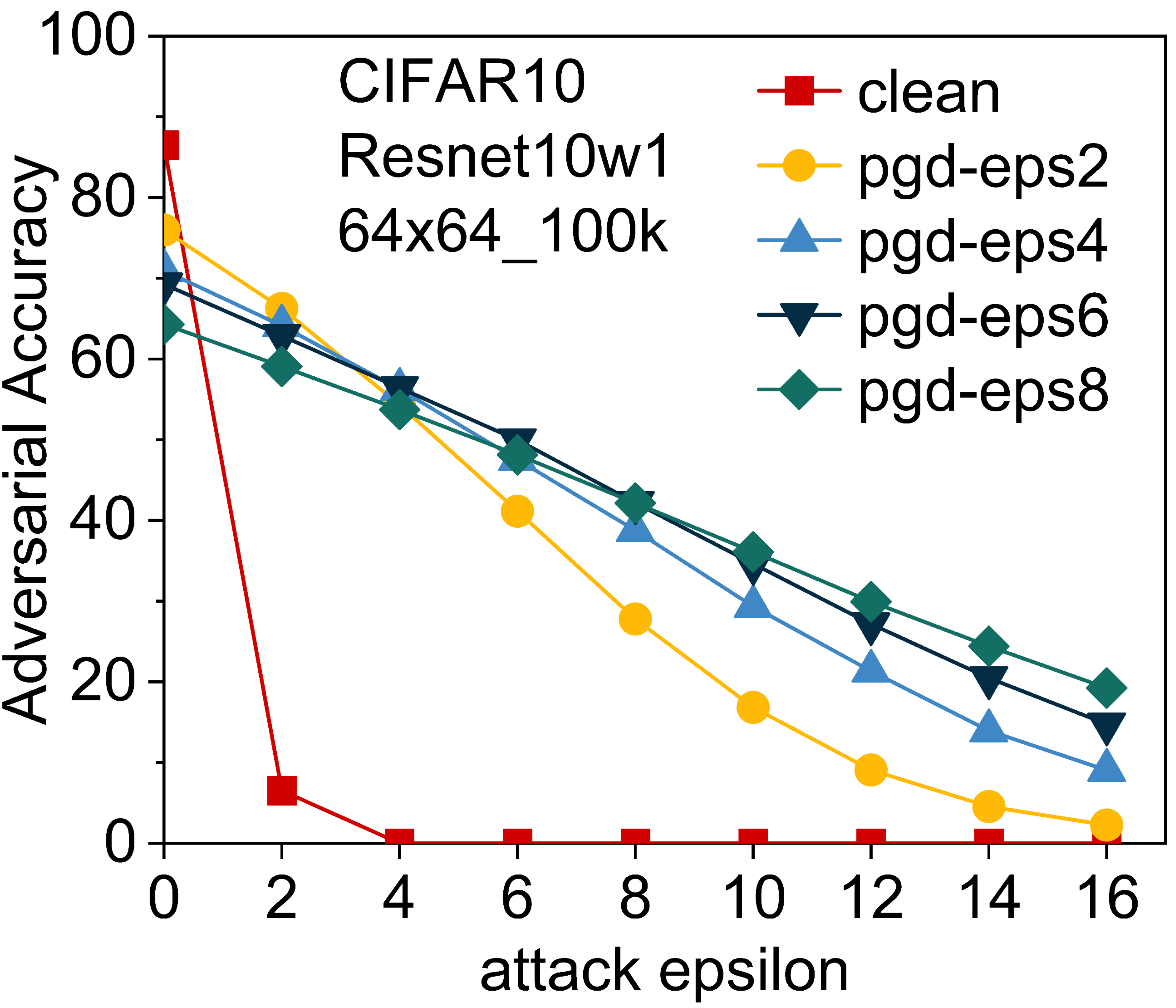}}
\caption{Adversarial Accuracy under PGD White Box Attack (iter $=50$) for ResNet10w1 architecture and CIFAR-10 implemented on 3 different hardware (a) Accurate Digital, (b) NVM crossbar model: 32x32\_100k ($NF=0.14$), and (c) NVM crossbar model: 64x64\_100k ($NF=0.26$). \textit{clean}: vanilla training with unperturbed images. \textit{pgd-epsN}: PGD adversarial training with $\epsilon_{train}= N = [2,4,6,8]$ and iter $=50$.}
\label{fig:adv-r10w1}
\end{figure*}

\begin{figure*}[h]
\centering
\sidesubfloat[]{\label{fig:r10w4-digital}\includegraphics[width=0.22\linewidth]{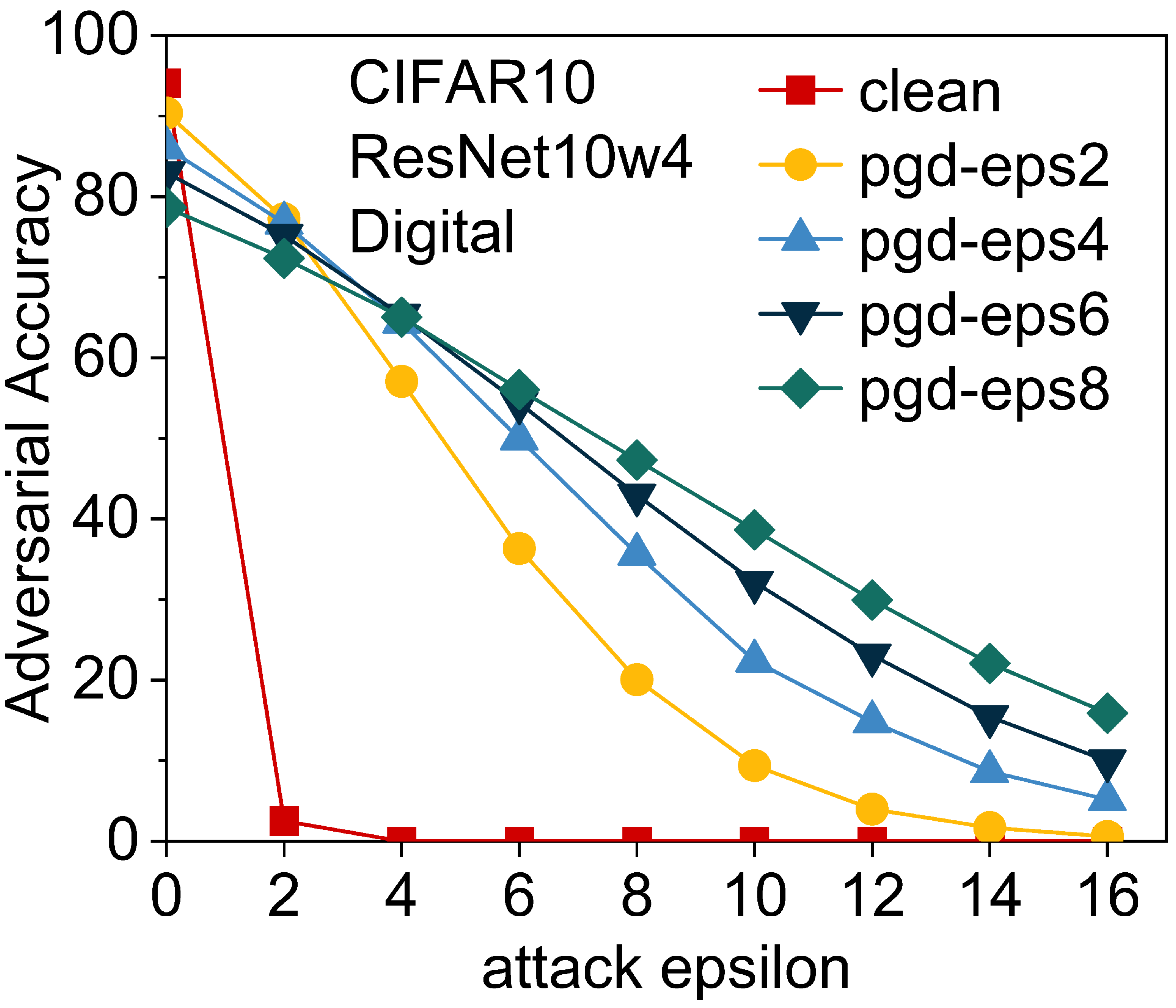}} 
\hfil
\sidesubfloat[]{\label{fig:r10w4-32x32-100k}\includegraphics[width=0.22\linewidth]{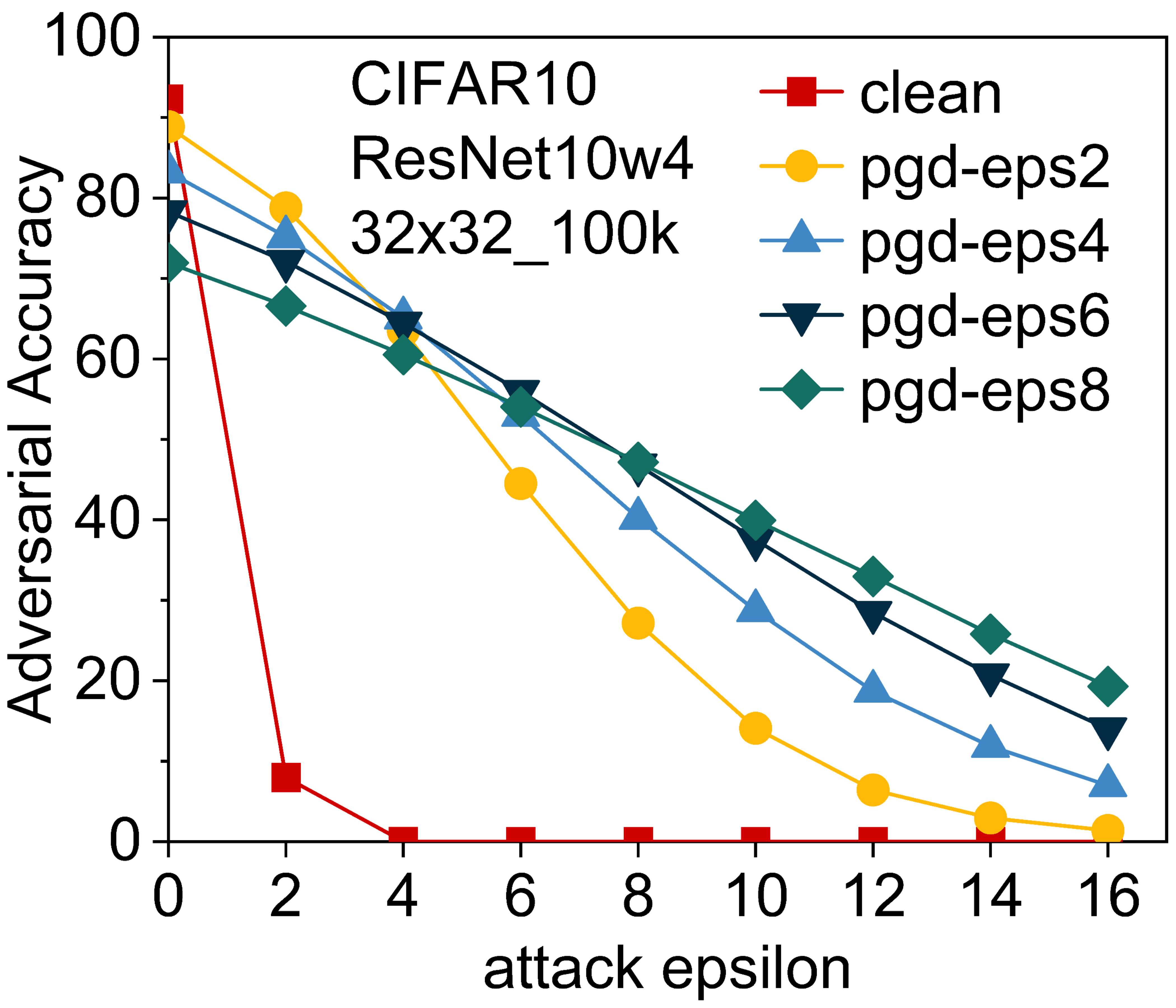}}
\hfil
\sidesubfloat[]{\label{fig:r10w4-64x64-100k}\includegraphics[width=0.22\linewidth]{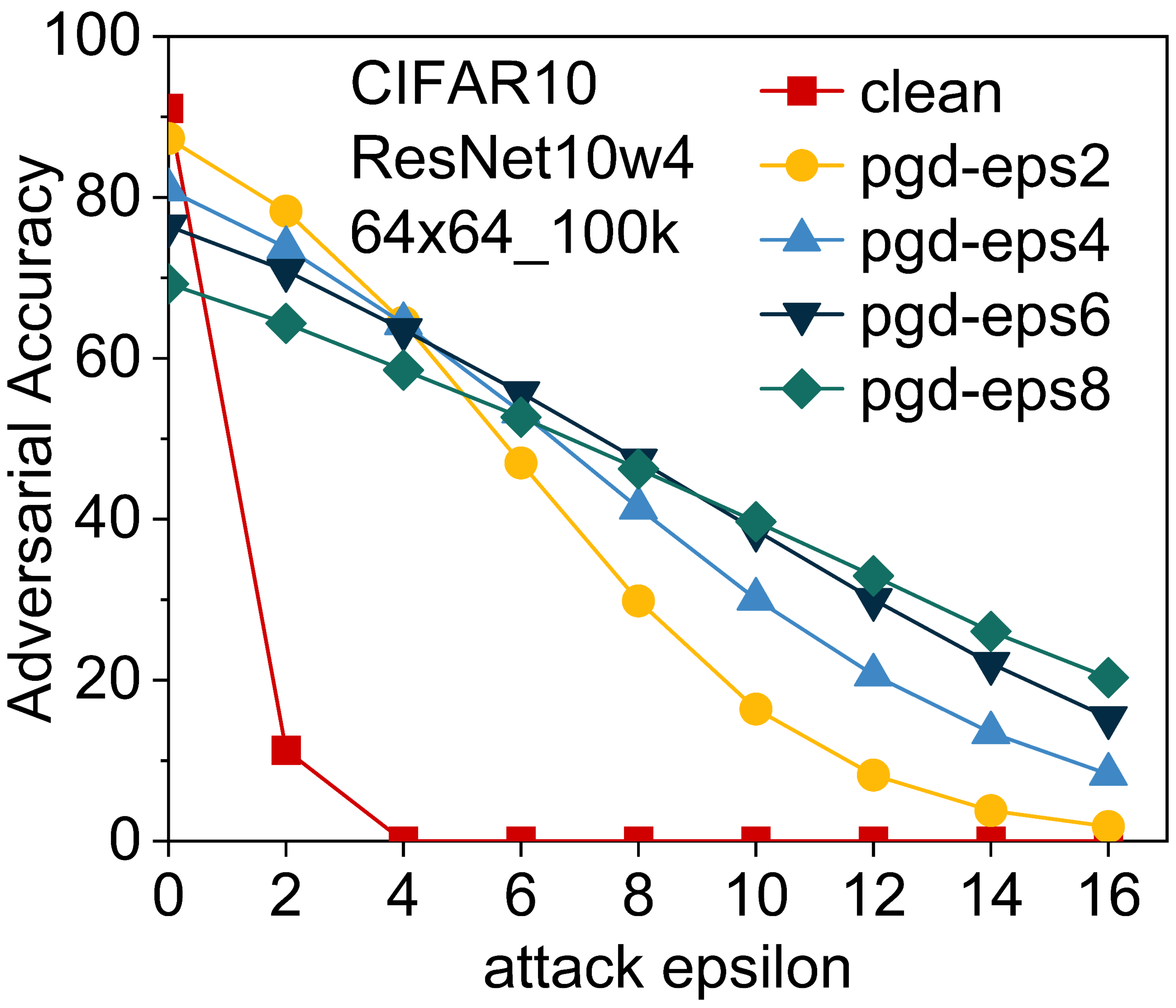}}
\caption{Adversarial Accuracy under PGD White Box Attack (iter $=50$) for ResNet10w4 architecture and CIFAR-10 implemented on 3 different hardware (a) Accurate Digital, (b) NVM crossbar model: 32x32\_100k ($NF=0.14$), and (c) NVM crossbar model: 64x64\_100k ($NF=0.26$). \textit{clean}: vanilla training with unperturbed images. \textit{pgd-epsN}: PGD adversarial training with $\epsilon_{train}= N = [2,4,6,8]$ and iter $=50$.}
\label{fig:adv-r10w4}
\end{figure*}

\begin{figure*}[h]
\centering
\sidesubfloat[]{\label{fig:r20w1-digital}\includegraphics[width=0.22\linewidth]{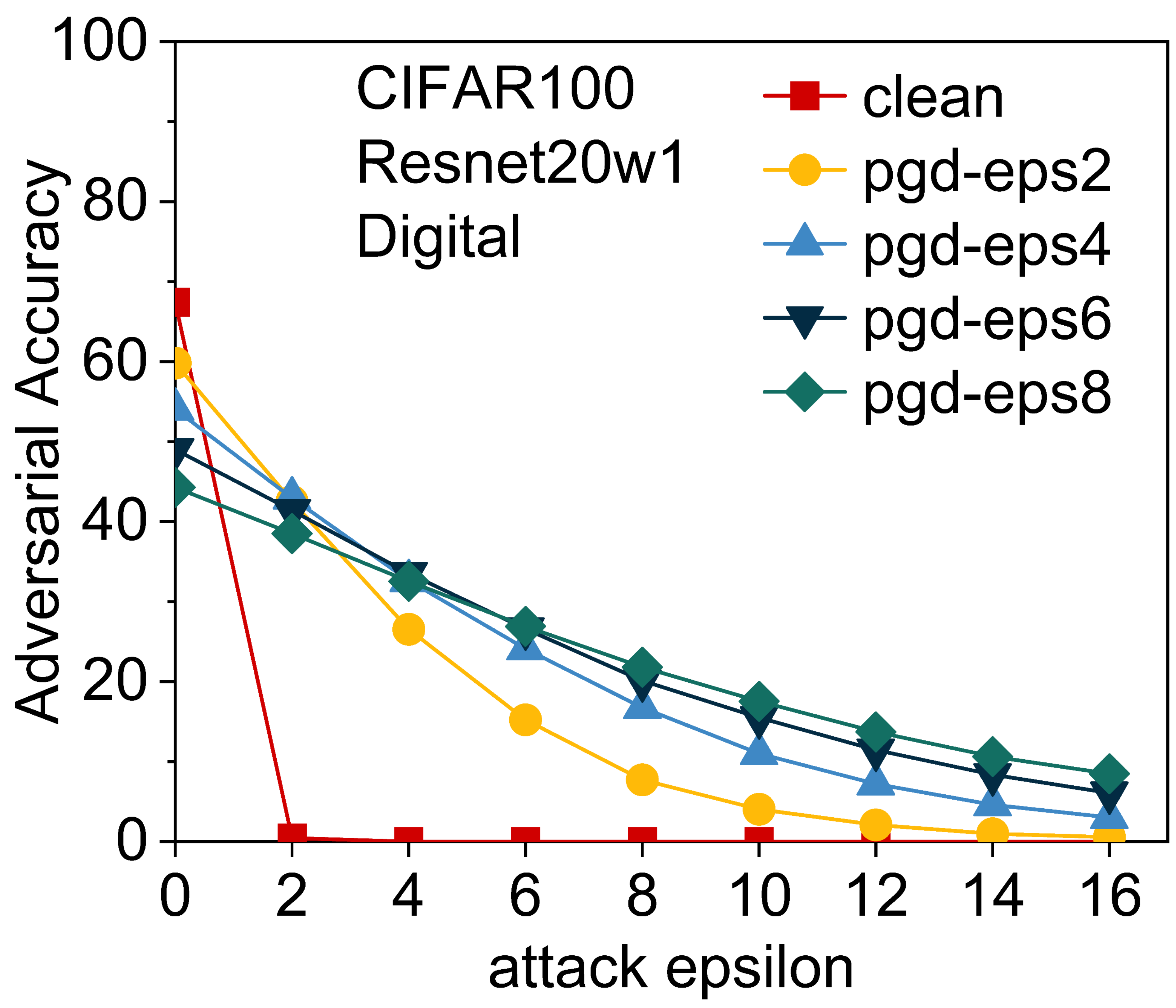}} 
\hfil
\sidesubfloat[]{\label{fig:r20w1-32x32-100k}\includegraphics[width=0.22\linewidth]{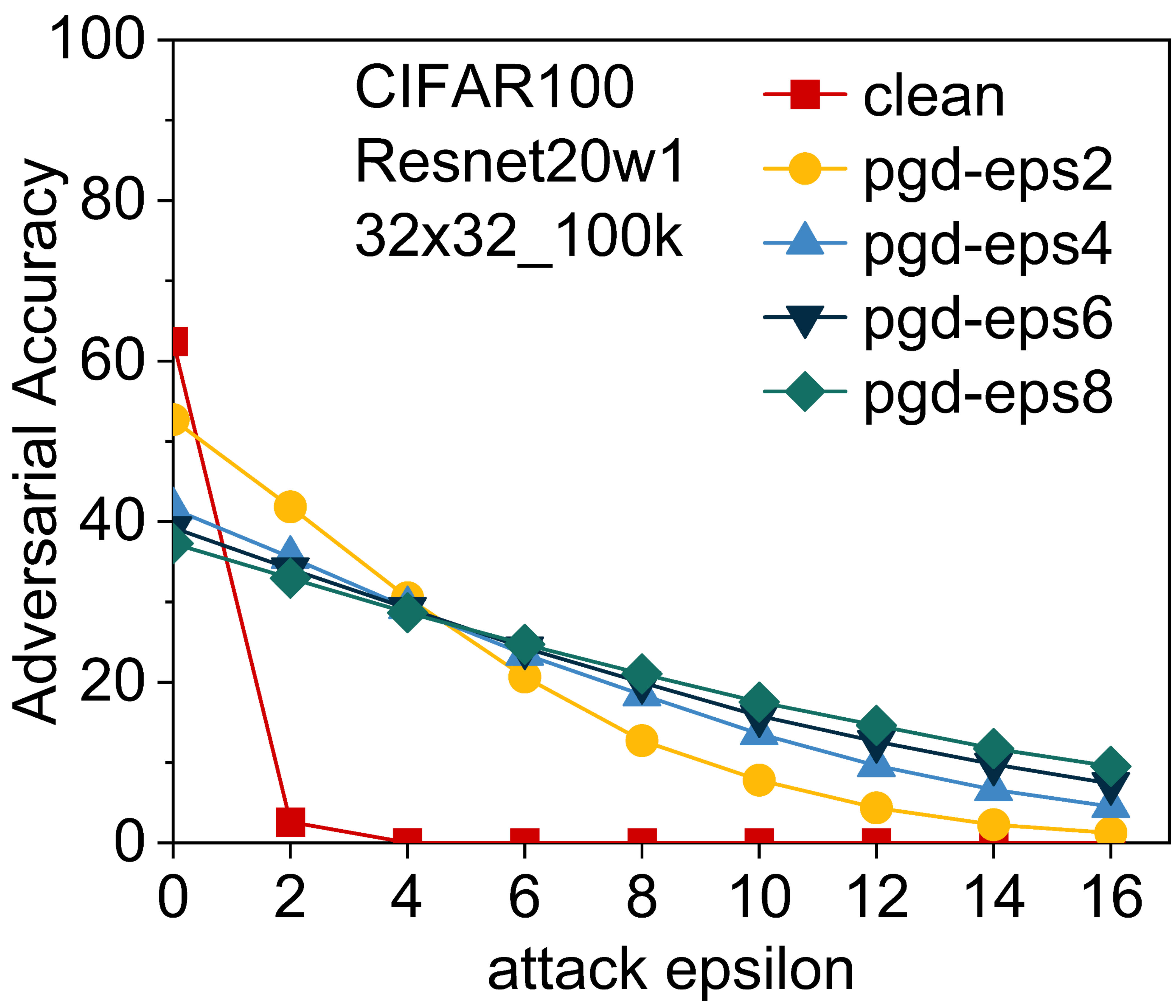}}
\hfil
\sidesubfloat[]{\label{fig:r20w1-64x64-100k}\includegraphics[width=0.22\linewidth]{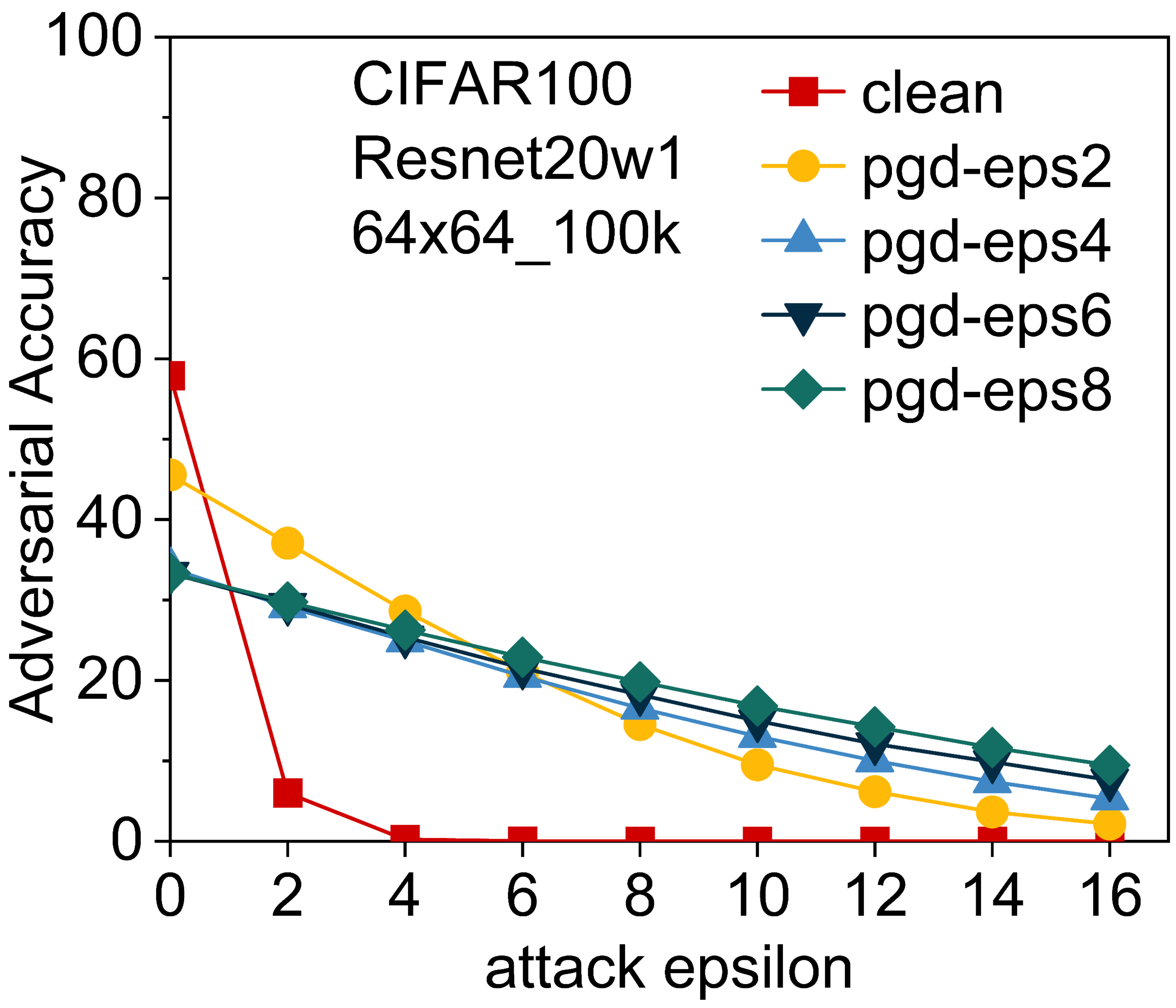}}
\caption{Adversarial Accuracy under PGD White Box Attack (iter $=50$) for ResNet20w1 architecture and CIFAR-100 implemented on 3 different hardware (a) Accurate Digital, (b) NVM crossbar model: 32x32\_100k ($NF=0.14$), and (c) NVM crossbar model: 64x64\_100k ($NF=0.26$). \textit{clean}: vanilla training with unperturbed images. \textit{pgd-epsN}: PGD adversarial training with $\epsilon_{train}= N = [2,4,6,8]$ and iter $=50$.}
\label{fig:adv-r20w1}
\end{figure*}

\begin{figure*}[h]
\centering
\sidesubfloat[]{\label{fig:r20w4-digital}\includegraphics[width=0.22\linewidth]{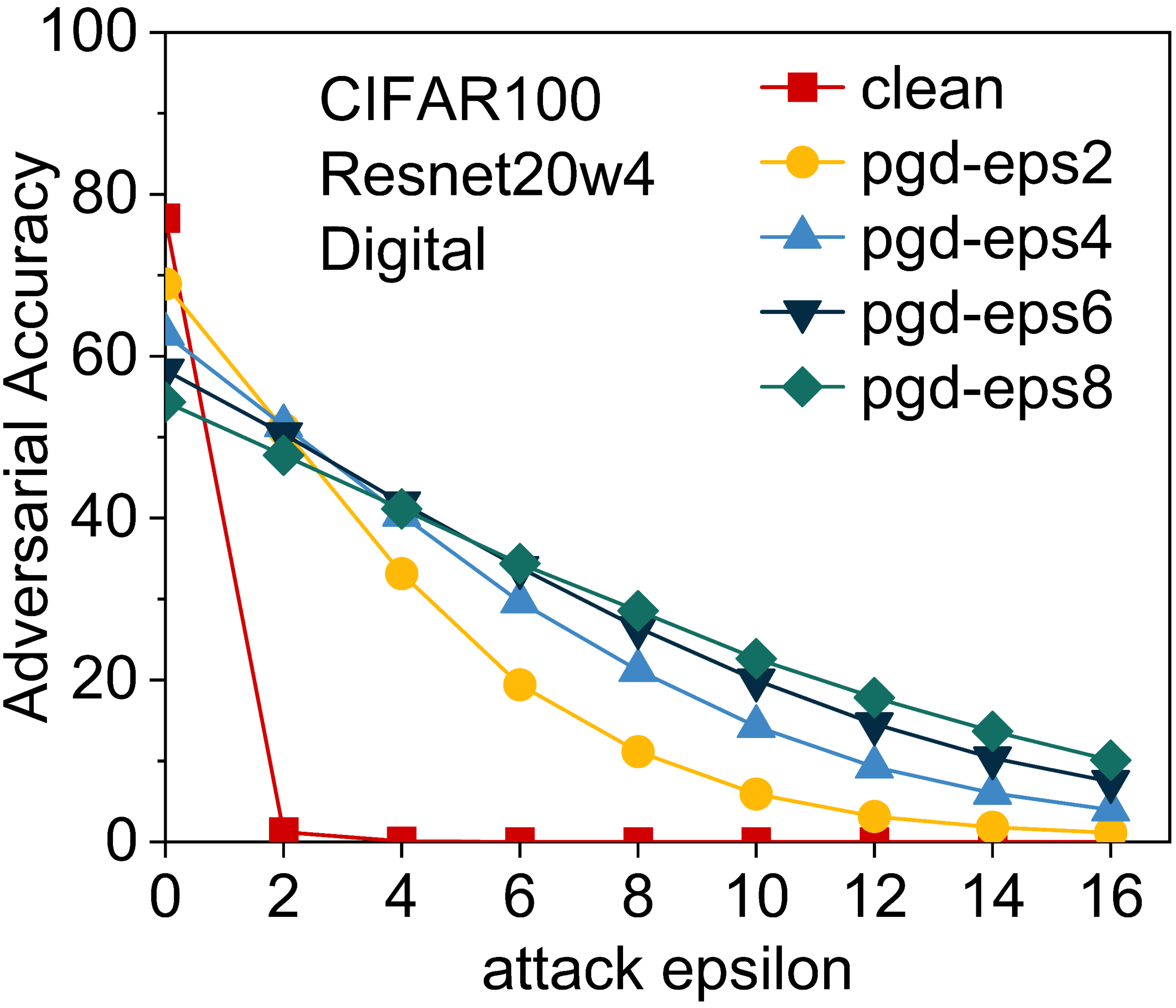}} 
\hfil
\sidesubfloat[]{\label{fig:r20w4-32x32-100k}\includegraphics[width=0.22\linewidth]{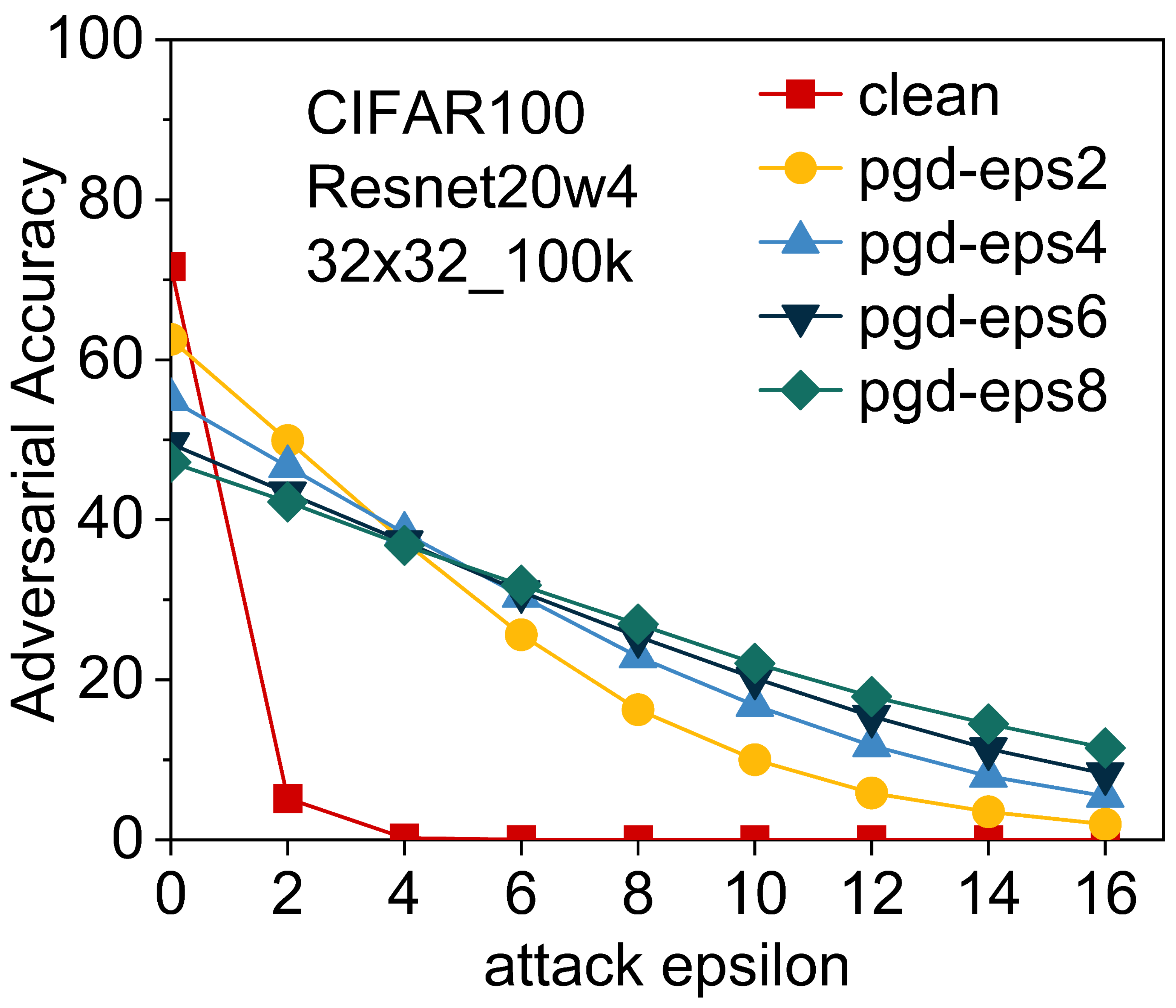}}
\hfil
\sidesubfloat[]{\label{fig:r20w4-64x64-100k}\includegraphics[width=0.22\linewidth]{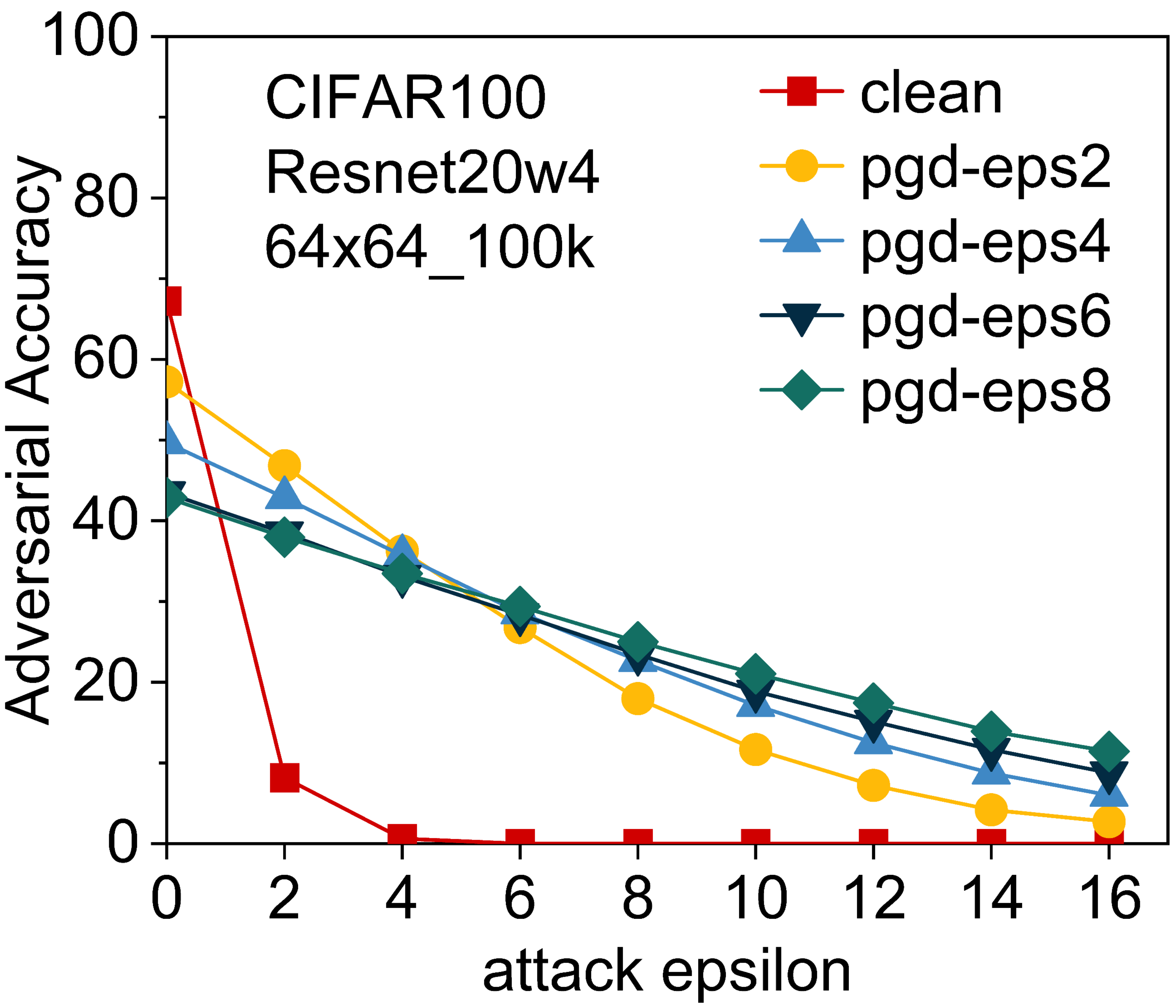}}
\caption{Adversarial Accuracy under PGD White Box Attack (iter $=50$) for ResNet20w4 architecture and CIFAR-100 implemented on 3 different hardware (a) Accurate Digital, (b) NVM crossbar model: 32x32\_100k ($NF=0.14$), and (c) NVM crossbar model: 64x64\_100k ($NF=0.26$). \textit{clean}: vanilla training with unperturbed images. \textit{pgd-epsN}: PGD adversarial training with $\epsilon_{train}= N = [2,4,6,8]$ and iter $=50$.}
\label{fig:adv-r20w4}
\end{figure*}

\begin{figure*}[h]
\centering
\sidesubfloat[]{\label{fig:pgd-eps2-32x32-100k}\includegraphics[width=0.22\linewidth]{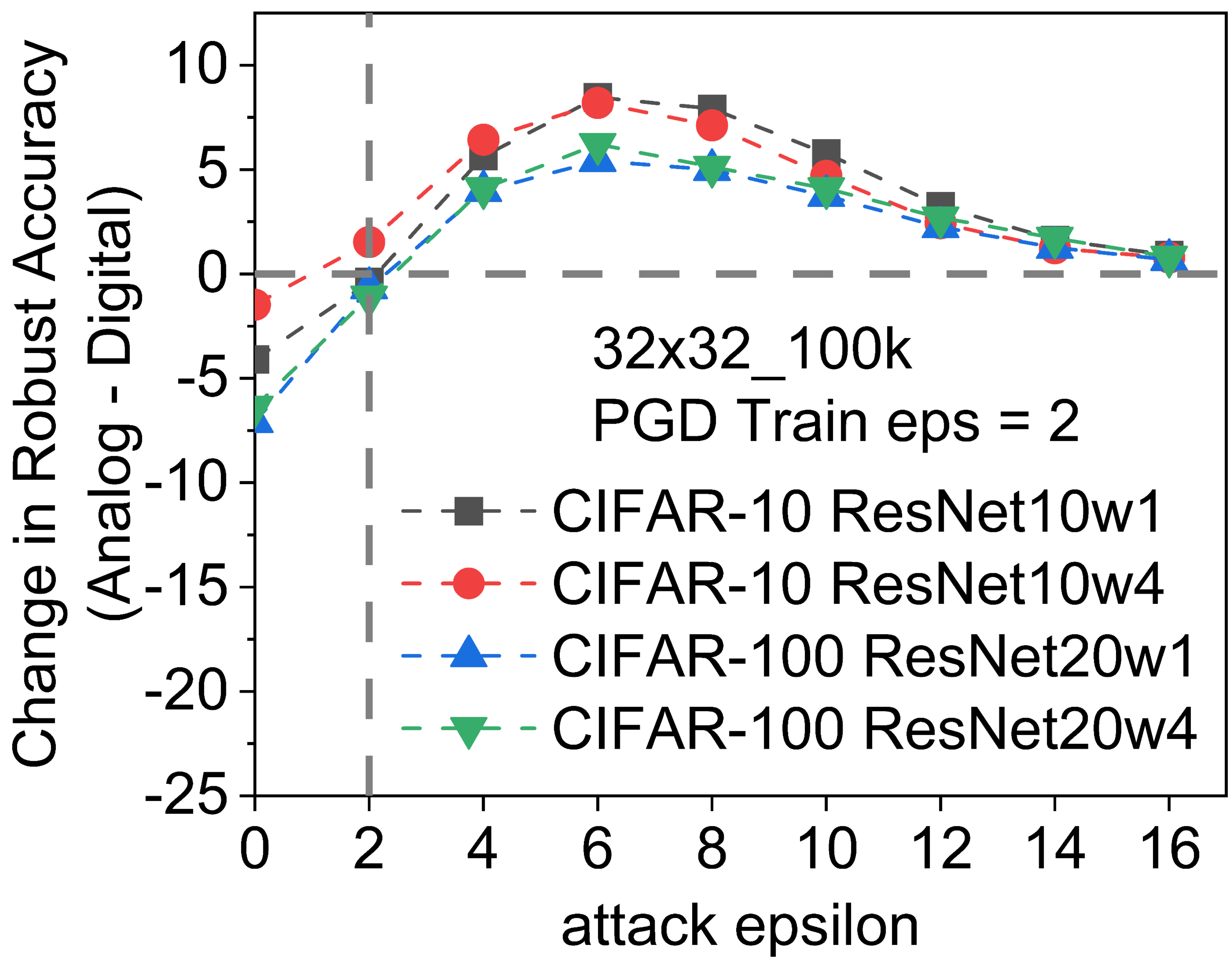}} 
\hfil
\sidesubfloat[]{\label{fig:pgd-eps4-32x32-100k}\includegraphics[width=0.22\linewidth]{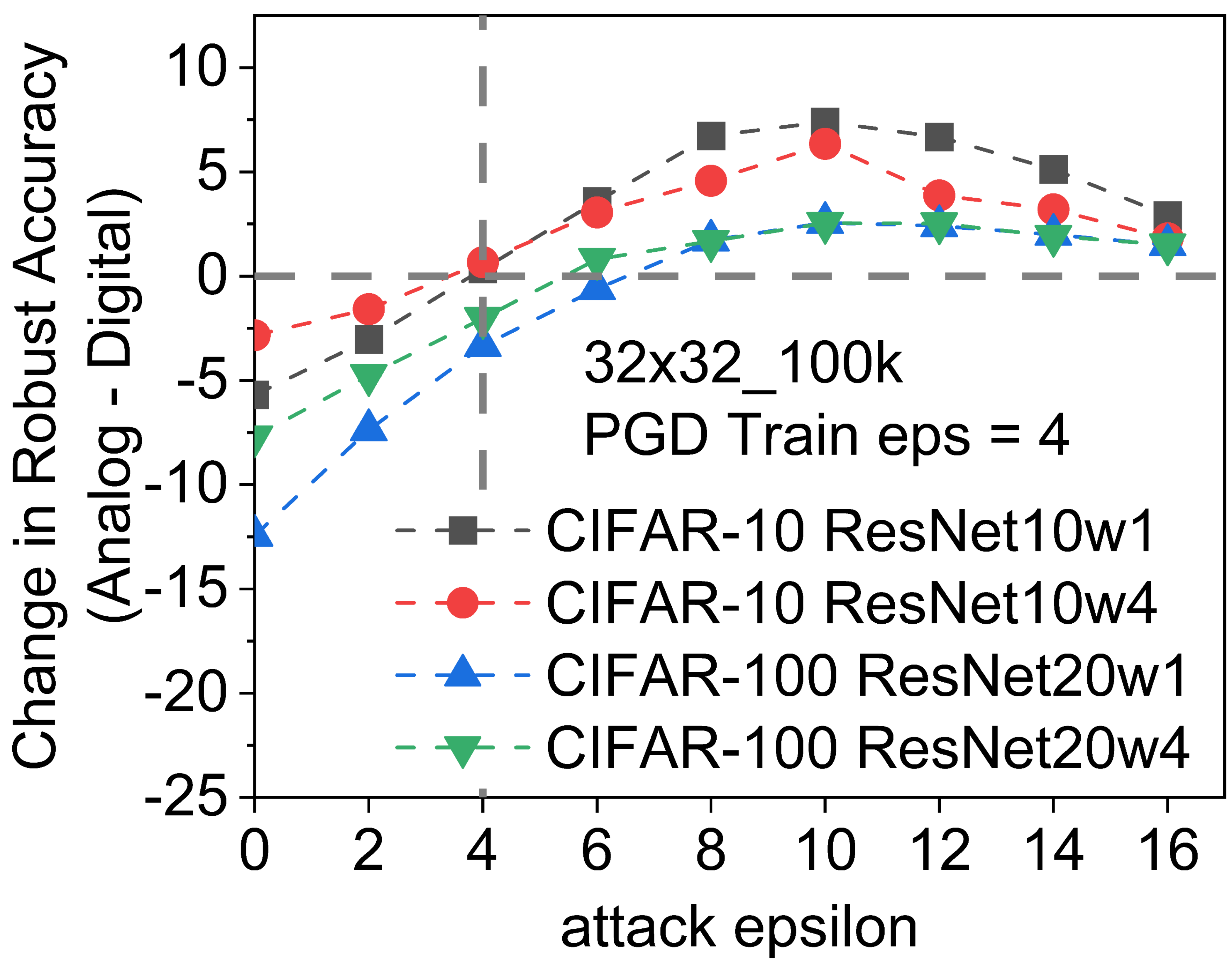}}
\hfil
\sidesubfloat[]{\label{fig:pgd-eps6-32x32-100k}\includegraphics[width=0.22\linewidth]{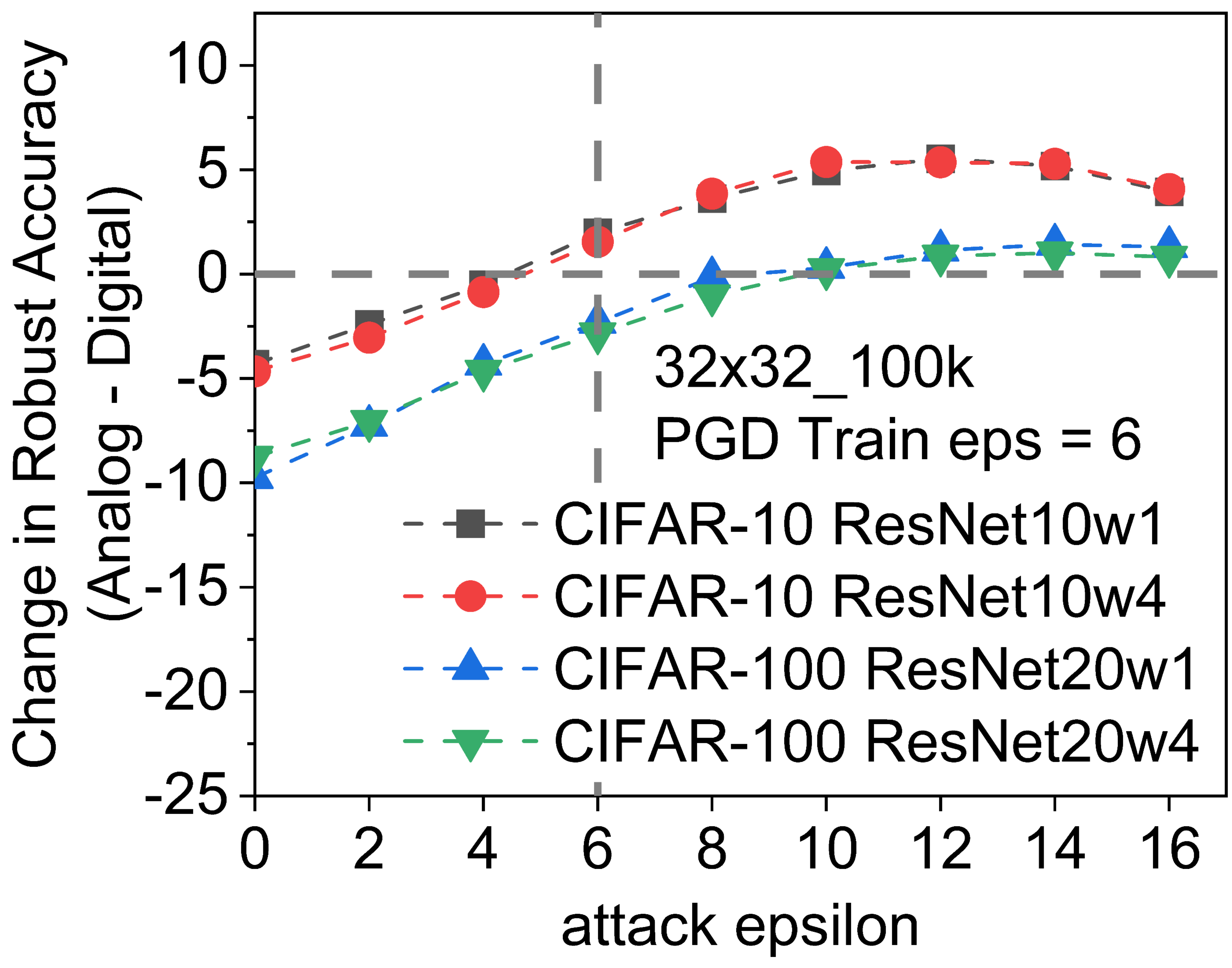}} 
\hfil
\sidesubfloat[]{\label{fig:pgd-eps8-32x32-100k}\includegraphics[width=0.22\linewidth]{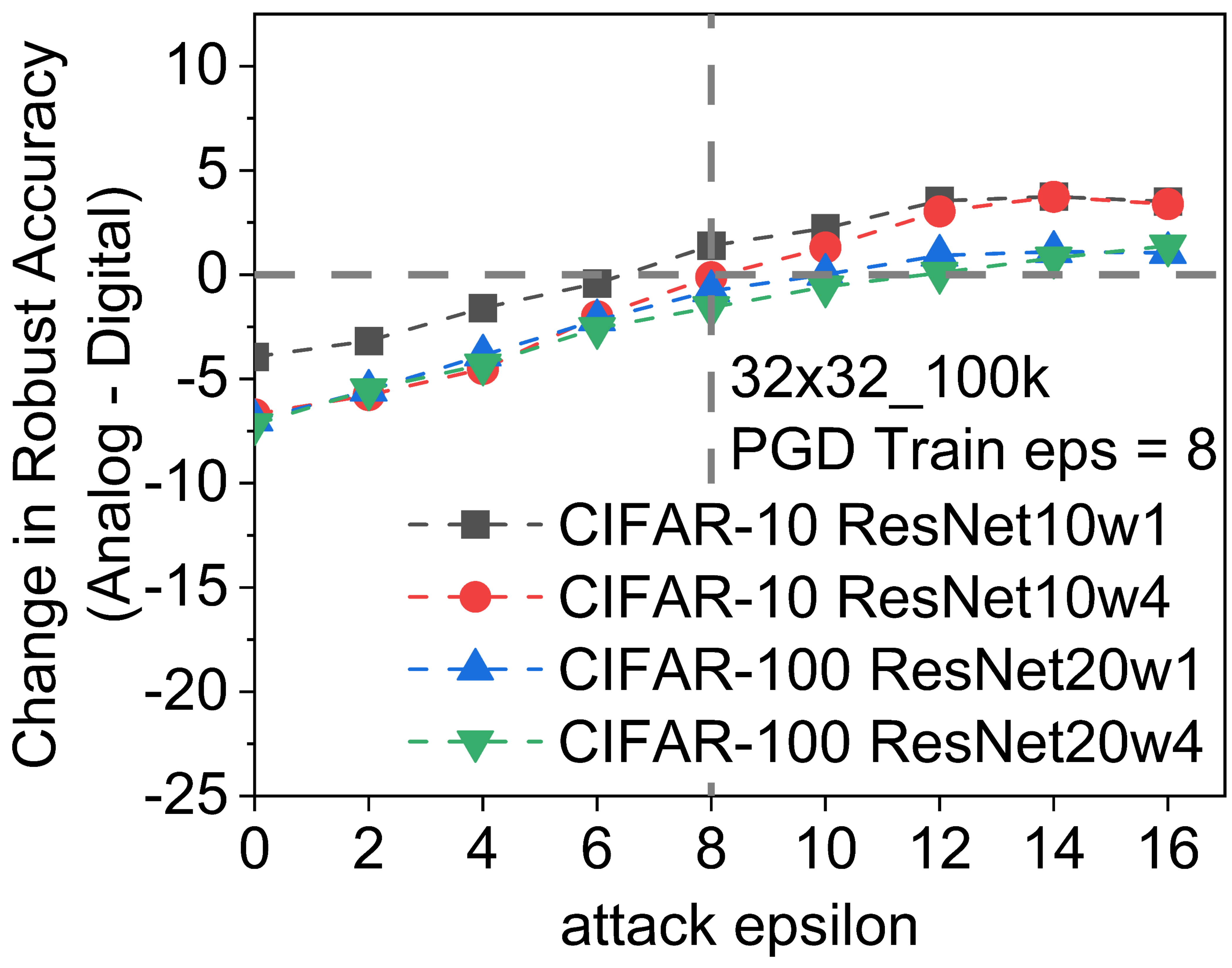}}
\caption{Difference in Adversarial Accuracy (Robustness Gain = Analog - Digital) for varying PGD Attack ($\epsilon_{attack}$). NVM crossbar model: 32x32\_100k ($NF = 0.14$). 4 network architectures (ResNet10w1, ResNet10w4, ResNet20w1, ResNet20w4) on 2 datasets (CIFAR-10, CIFAR-100) are adversarially trained with (a) PGD, $\epsilon_{train} = 2$ and iter$=50$ (b) PGD, $\epsilon_{train} = 4$ and iter$=50$ (c) PGD, $\epsilon_{train} = 6$ and iter$=50$ (d) PGD, $\epsilon_{train} = 8$ and iter$=50$}
\label{fig:pgd-32x32-100k}
\end{figure*}

\begin{figure*}[h]
\centering
\sidesubfloat[]{\label{fig:pgd-eps2-64x64-100k}\includegraphics[width=0.22\linewidth]{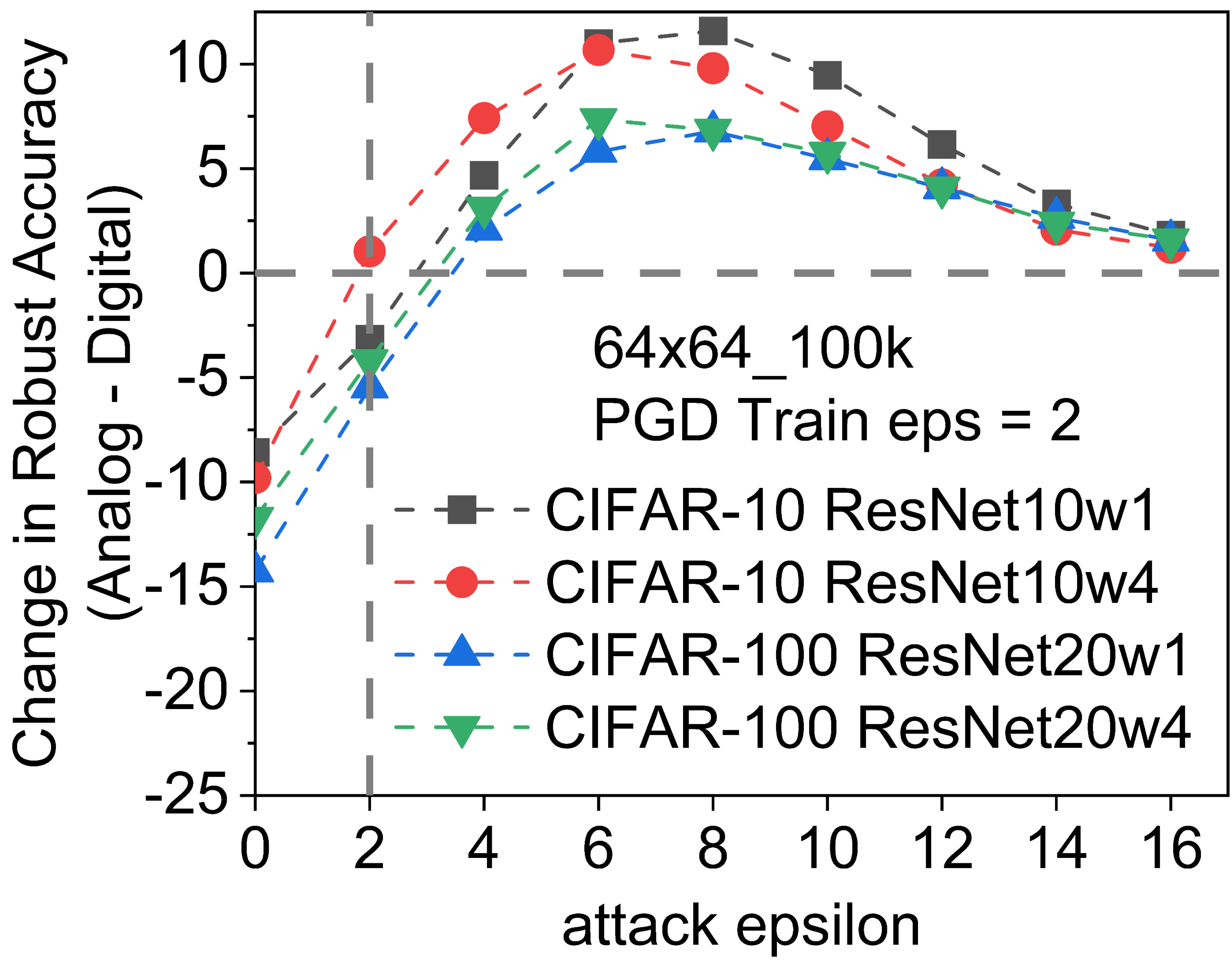}} 
\hfil
\sidesubfloat[]{\label{fig:pgd-eps4-64x64-100k}\includegraphics[width=0.22\linewidth]{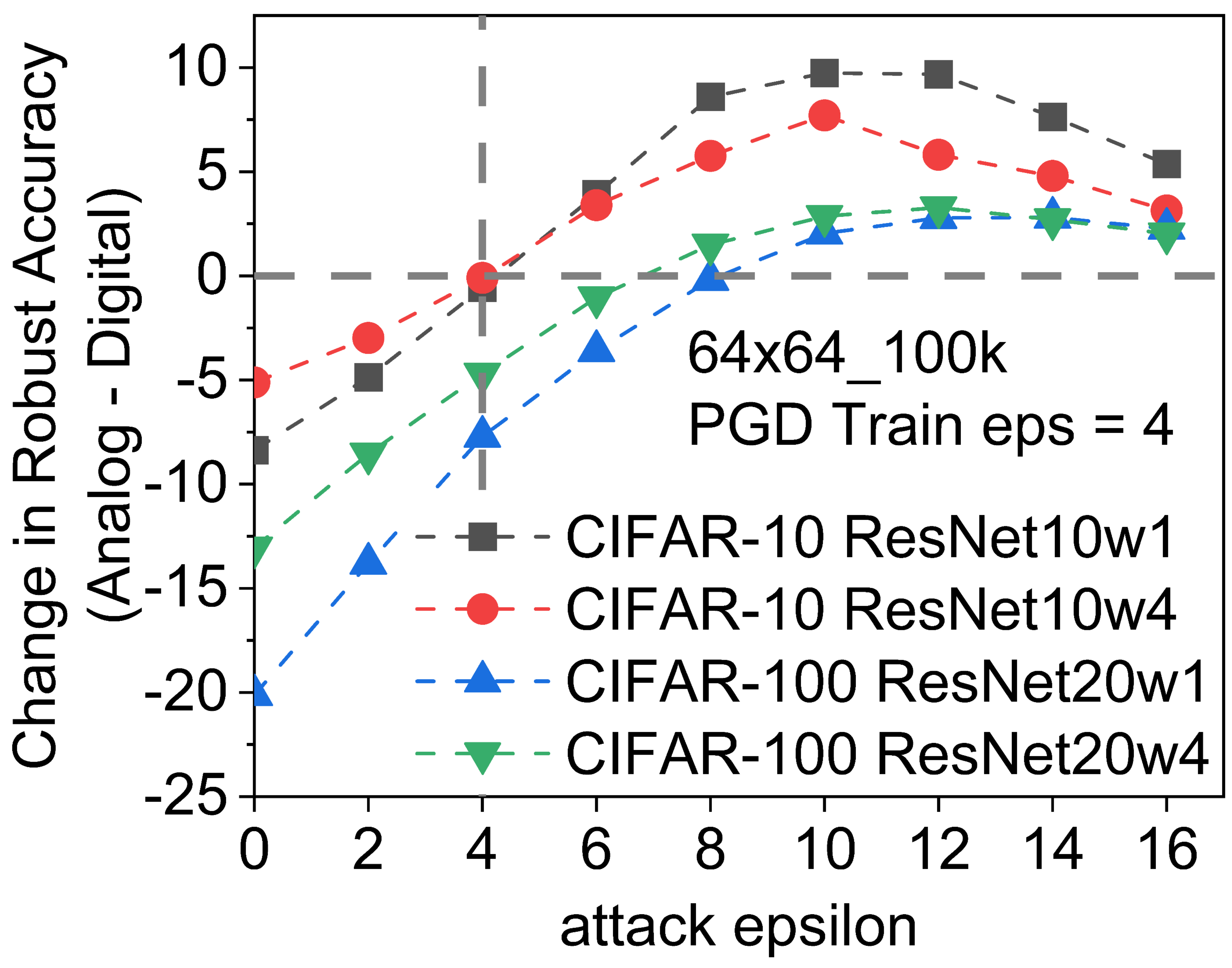}}
\hfil
\sidesubfloat[]{\label{fig:pgd-eps6-64x64-100k}\includegraphics[width=0.22\linewidth]{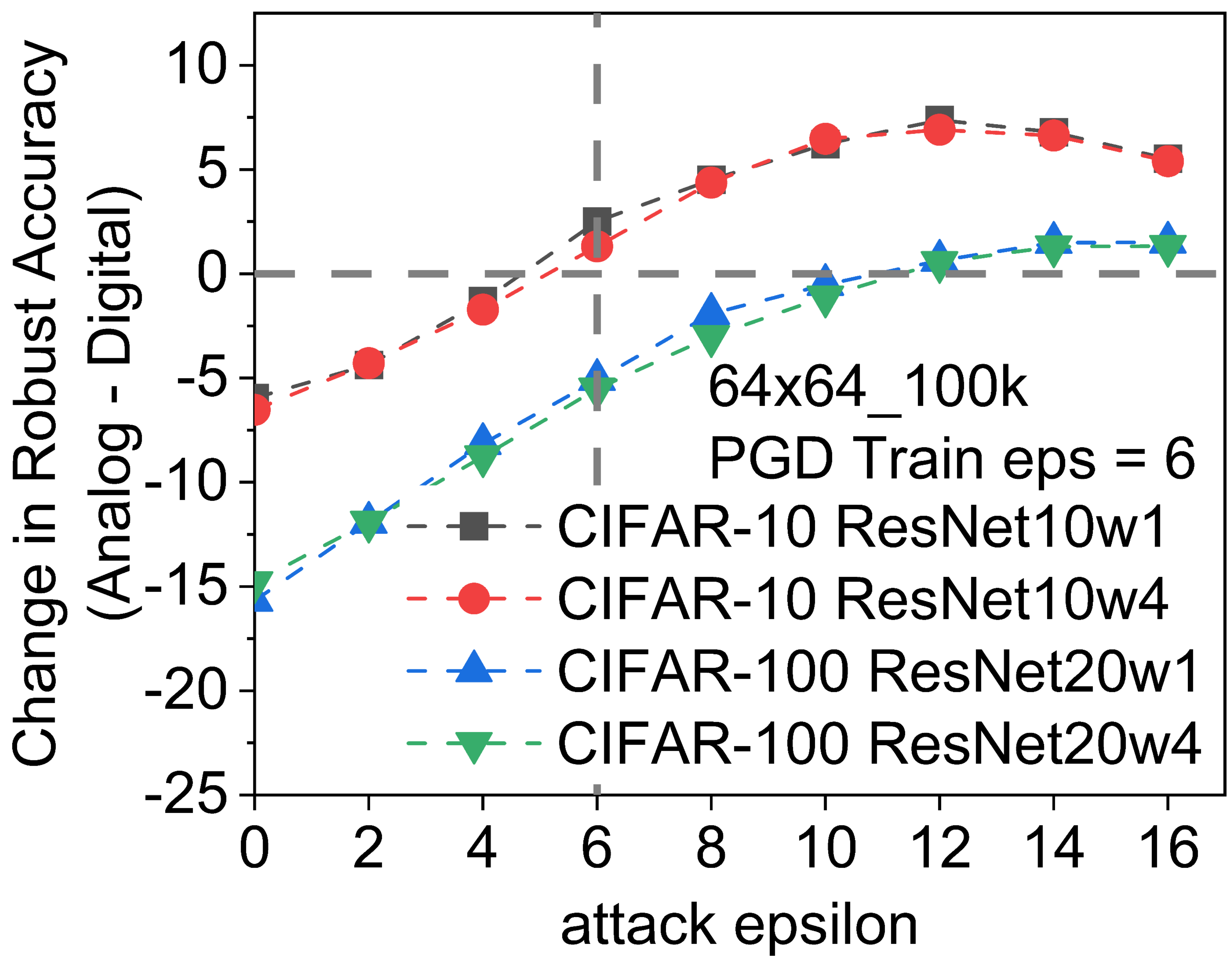}} 
\hfil
\sidesubfloat[]{\label{fig:pgd-eps8-64x64-100k}\includegraphics[width=0.22\linewidth]{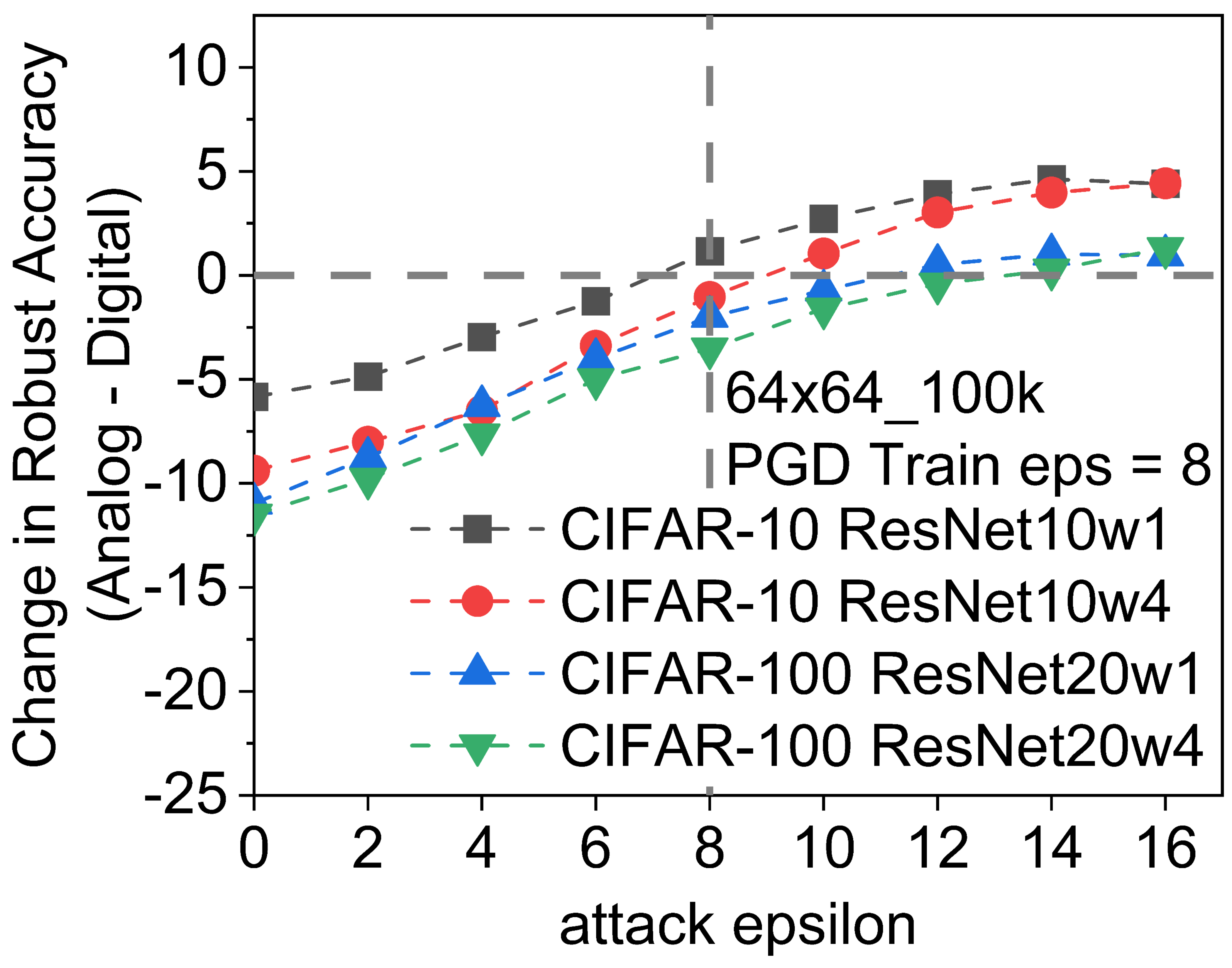}}
\caption{Difference in Adversarial Accuracy (Robustness Gain = Analog - Digital) for varying PGD Attack ($\epsilon_{attack}$). NVM crossbar model: 64x64\_100k ($NF = 0.26$). 4 network architectures (ResNet10w1, ResNet10w4, ResNet20w1, ResNet20w4) on 2 datasets (CIFAR-10, CIFAR-100) are adversarially trained with (a) PGD, $\epsilon_{train} = 2$ and iter$=50$ (b) PGD, $\epsilon_{train} = 4$ and iter$=50$ (c) PGD, $\epsilon_{train} = 6$ and iter$=50$ (d) PGD, $\epsilon_{train} = 8$ and iter$=50$ }
\label{fig:pgd-64x64-100k}
\end{figure*}

Next, we evaluate the DNNs under Non-Adaptive White Box Attack, where the attacker has complete information of the network weights, however they have no knowledge of the NVM crossbar. The attacker assumes the underlying hardware is accurate, and generates adversarial images using iterative PGD (iter =50). From Fig. \ref{fig:adv-r10w1}-\ref{fig:adv-r20w1}, we see that adversarially trained DNNs perform better than vanilla DNNs for all values of $\epsilon_{attack}$, and in every hardware. 

In Fig. \ref{fig:pgd-32x32-100k} and Fig. \ref{fig:pgd-64x64-100k}, we plot the net gain in robustness, that is the difference in adversarial accuracy when the same DNN is implemented on an accurate digital hardware and NVM crossbar under the same adversarial attack condition. We observe that for a network trained with $\epsilon_{train}=N$, the intrinsic robustness of the network starts having a positive effect only in attacks where $\epsilon_{attack} > N$. At $\epsilon_{attack}=0$, i.e in the absence of any attack, the DNNs' accuracy on the analog hardware is significantly lower than digital implementation. As $\epsilon_{attack}$ is increased, the difference in accuracy slowly reduces, and reaches a tipping point when the $\epsilon_{attack}$ is equal to $\epsilon_{train}$. Beyond that, the analog implementation has a higher accuracy than the digital, i.e. we are in the zone of robustness gain due to the non-idealities. We can draw the following conclusion:
\begin{itemize}
    \item \textbf{The Push-Pull Effect}: As noted in \cite{roy2020robustness}, when under adversarial attack, the non-idealities have a dual effect on the accuracy. The first is the loss in accuracy because of inaccurate computations. And the second is the increase in accuracy due to gradient obfuscation. Because the hardware is unknown to the attacker, the adversarial attacks do not transfer fully to our implementation. At lower $\epsilon_{attack}$, the loss due to inaccurate computations is higher, while at higher $\epsilon_{attack}$, the intrinsic robustness becomes more significant.
    \item \textbf{Out of Distribution Samples}: In adversarial training, we train DNNs on images with adversarial perturbations bound by the training $\epsilon_{train}$. During inference, adversarial images with $\epsilon_{attack} < \epsilon_{train}$ would have a distribution similar to the training images, and the DNN behaves similar to a vanilla DNN inferring unperturbed images. We only observe accuracy loss due to non-ideal computations. However, adversarial images with $\epsilon_{attack} > \epsilon_{train}$ are out of distribution with respect to the training data. Now, the DNN is under adversarial attack, and the intrinsic robustness of the NVM crossbar boosts the accuracy. 
\end{itemize}

\textit{Design for Robust DNNs on NVM Crossbars}: When designing robust DNNs on digital hardware, the tradeoff between the adversarial and the natural accuracy of a DNN is fairly straightforward. A higher $\epsilon_{train}$ during training leads to lowering of natural accuracy, and increase in adversarial accuracy, as shown in Fig. \ref{fig:adv-r10w1}(a), Fig. \ref{fig:adv-r10w4}(a), Fig. \ref{fig:adv-r20w1}(a), and Fig. \ref{fig:adv-r20w4}(a). However, the interplay of non-idealities and adversarial loss creates opportunities for design of DNNs where a DNN trained on lower $\epsilon_{train}$ can achieve both higher natural accuracy and higher adversarial accuracy, as shown in Fig. \ref{fig:adv-r20w1}(b) and (c), and Fig. \ref{fig:adv-r20w4}(b) and (c). For $\epsilon_{attack} = [2,4]$ on 32x32\_100k and for $\epsilon_{attack} = [2,4, 6]$ on 64x64\_100k, the ResNet20w1 DNN trained with $\epsilon_{train}=2$ performs better or at par than the DNNs trained with higher $\epsilon_{train}$. Thus, the non-idealities play a significant role in determining the best DNN within a desired robustness range.

\textit{Natural Accuracy, Adversarial Robustness, and Energy Efficiency}: In most of non-adversarial machine learning applications on analog NVM crossbars, we perfer to perform MVM operations on smaller-sized crossbars (e.g. 32x32\_100k) because they have smaller non-ideality factor (NF) compared to larger crossbars (e.g. 64x64\_100k), shown in Table \ref{tab:params}. When DNNs are adversarially trained with iterative PGD \cite{madry2018towards}, we also observe smaller accuracy drop for 32x32\_100k compared to 64x64\_100k, as shown in Fig. \ref{fig:dig_vs_an}. While the larger crossbars like 64x64\_100k exhibit higher energy efficiency for MVM operations \cite{NAX2021}, they generally incur lower classifcation accuracy on unattacked images. However, when DNNs encounter strong adversarial attacks, we leverage the interplay between adversarial training and non-idealities of differnt NVM crossbars to harvest adversarial robustness for DNNs. For example, in Fig. \ref{fig:pgd-eps2-64x64-100k}, we evaluate DNNs trained with low $\epsilon_{train}=2$ on 64x64\_100k crossbar, and harvest robustness gain (up to 11.59\%) against all PGD white-box attacks with $\epsilon_{attack}>2$. We conclude that we can sacrifice natural accuracy of larger crossbars for their adversarial robustness gains, while benefiting from their energy efficiency.

\begin{table}[h]
\centering
\caption{Non-Adaptive Attacks on NVM Crossbar Models}
\vspace{-7pt}
\label{table:nonadaptivetable}
\resizebox{0.9\linewidth}{!}{
\begin{tabular}{llcrr} 
\toprule
& & & \multicolumn{2}{c}{NVM Crossbar Models} \\
\cmidrule(lr){4-5}
Attack Type & Train Type & Baseline (Digital) & \multicolumn{1}{c}{32x32\_100k} & \multicolumn{1}{c}{64x64\_100k}\\
\cmidrule(lr){1-5}
& & \multicolumn{3}{c}{CIFAR-10 (ResNet-10w1) (test samples = 10000)}\\
\cmidrule(lr){3-5}
\multirow{3}{*}{Clean} & Vanilla & 89.85 & 88.26 (-1.59) & 86.51 (-3.34)\\ 
& PGD-eps2 & 84.55 & 80.45 (-4.10) & 75.98 (-8.57)\\
& PGD-eps6 & 75.22 & 70.91 (-4.31) & 69.27 (-5.95)\\
\cmidrule(lr){2-5}
Square Attack (Black Box) & Vanilla & 11.10 & 61.82 (+50.72) & 62.93 (+51.83)\\
$\epsilon=4/255$ & PGD-eps2 & 72.62 & 74.01 (+1.39) & 70.29 (-2.33)\\
queries $=1000$ & PGD-eps6 & 67.49 & 67.35 (-0.14) & 65.82 (-1.67)\\
\cmidrule(lr){2-5}
Square Attack (Black Box) & Vanilla & 0.32 & 46.27 (+45.95) & 49.35 (+49.03)\\ 
 $\epsilon=8/255$ & PGD-eps2 & 57.14 & \textbf{67.26 (+10.12)} & 63.75 (+6.61)\\
queries $=1000$ & PGD-eps6 & 58.77 & 63.40 (+4.63) & 62.09 (+3.32)\\
\cmidrule(lr){2-5}
Whitebox PGD & Vanilla & 1.02 & 3.81 (+2.79) & 6.53 (+5.51)\\ 
$\epsilon=2/255$ & PGD-eps2 & 69.42 & 69.04 (-0.38) & 66.26 (-3.16)\\
iter $=50$ & PGD-eps6 & 67.25 & 64.84 (-2.41) & 62.88 (-4.37)\\
\cmidrule(lr){2-5}
Whitebox PGD & Vanilla & $\approx0$ & $\approx0$ (+0.00) & $\approx0$ (+0.00)\\ 
$\epsilon=8/255$ & PGD-eps2 & 16.2 & 24.12 (+7.92) & \textbf{27.79 (+11.59)}\\
iter $=50$ & PGD-eps6 & 37.69 & 41.31 (+3.62) & 42.2 (+4.51)\\
\midrule
& & \multicolumn{3}{c}{CIFAR-100 (ResNet-20w1) (test samples = 1000)}\\
\cmidrule(lr){3-5}
\multirow{3}{*}{Clean} & Vanilla & 67.44 & 62.47 (-4.97) & 57.92 (-9.52)\\ 
& PGD-eps2 & 59.81 & 52.72 (-7.09) & 45.55 (-14.26)\\
& PGD-eps6 & 49 & 39.25 (-9.75) & 33.33 (-15.67))\\
\cmidrule(lr){2-5}
Square Attack (Black Box) & Vanilla & 3.18 & 44.35 (+41.17) & 41.66 (+38.48)\\ 
$\epsilon=4/255$ & PGD-eps2 & 44.43 & 47.57 (+3.14) & 41.30 (-3.13)\\
queries $=1000$ & PGD-eps6 & 40.90 & 36.05 (-4.85) & 30.94 (-9.96)\\
\cmidrule(lr){2-5}
Square Attack (Black Box) & Vanilla & 0.23 & 30.74 (+30.51) & 27.29 (+27.06)\\ 
$\epsilon=8/255$ & PGD-eps2 & 30.48 & \textbf{42.55 (+12.07)} & 36.99 (+6.51)\\
queries $=1000$ & PGD-eps6 & 32.85 & 33.70 (+0.85) & 28.85 (-4.00)\\
\cmidrule(lr){2-5}
Whitebox PGD & Vanilla & 0.45 & 2.60 (+2.15) & 5.99 (+5.54)\\ 
$\epsilon=2/255$ & PGD-eps2 & 42.52 & 41.85 (-0.67) & 37.06 (-5.46)\\
iter $=50$ & PGD-eps6 & 41.39 & 34.13 (-7.26) & 29.46 (-11.93)\\
\cmidrule(lr){2-5}
Whitebox PGD & Vanilla & $\approx0$ & $\approx0$ (+0.00) & $\approx0$ (+0.00)\\ 
$\epsilon=8/255$ & PGD-eps2 & 7.73 & 12.72 (+4.99) & \textbf{14.52 (+6.79)}\\
iter $=50$ & PGD-eps6 & 20.12 & 19.97 (-0.15) & 18.21 (-1.91)\\
\bottomrule
\end{tabular}
}
\end{table}

\subsection{Gaussian Noise Injection}
\label{subsec:gaussian}
\begin{figure}[!htb]
\centering
\sidesubfloat[]{\label{fig:noise-eps2-std-32x32}\includegraphics[width=0.43\linewidth]{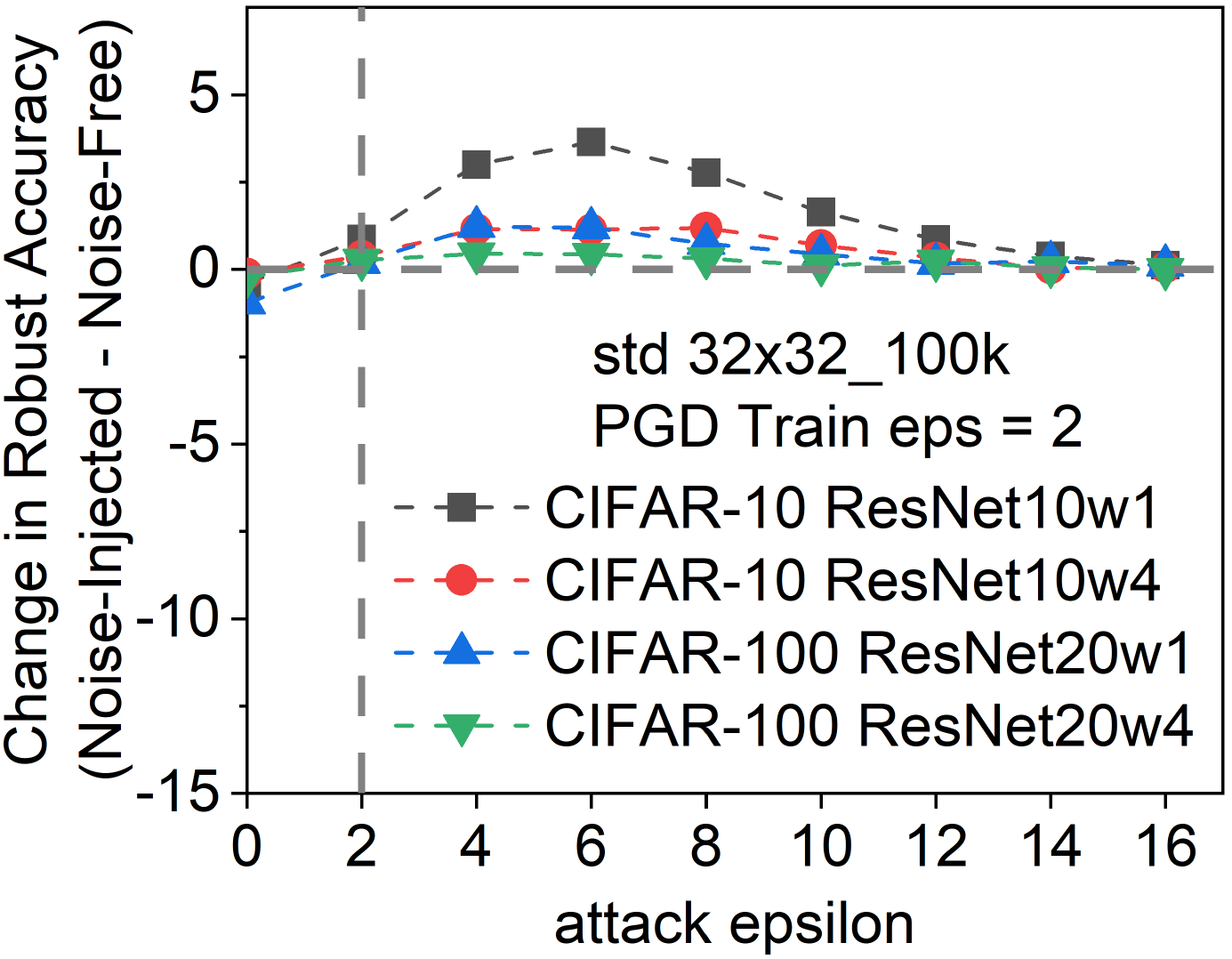}} 
\hfil
\sidesubfloat[]{\label{fig:noise-eps2-std-64x64}\includegraphics[width=0.43\linewidth]{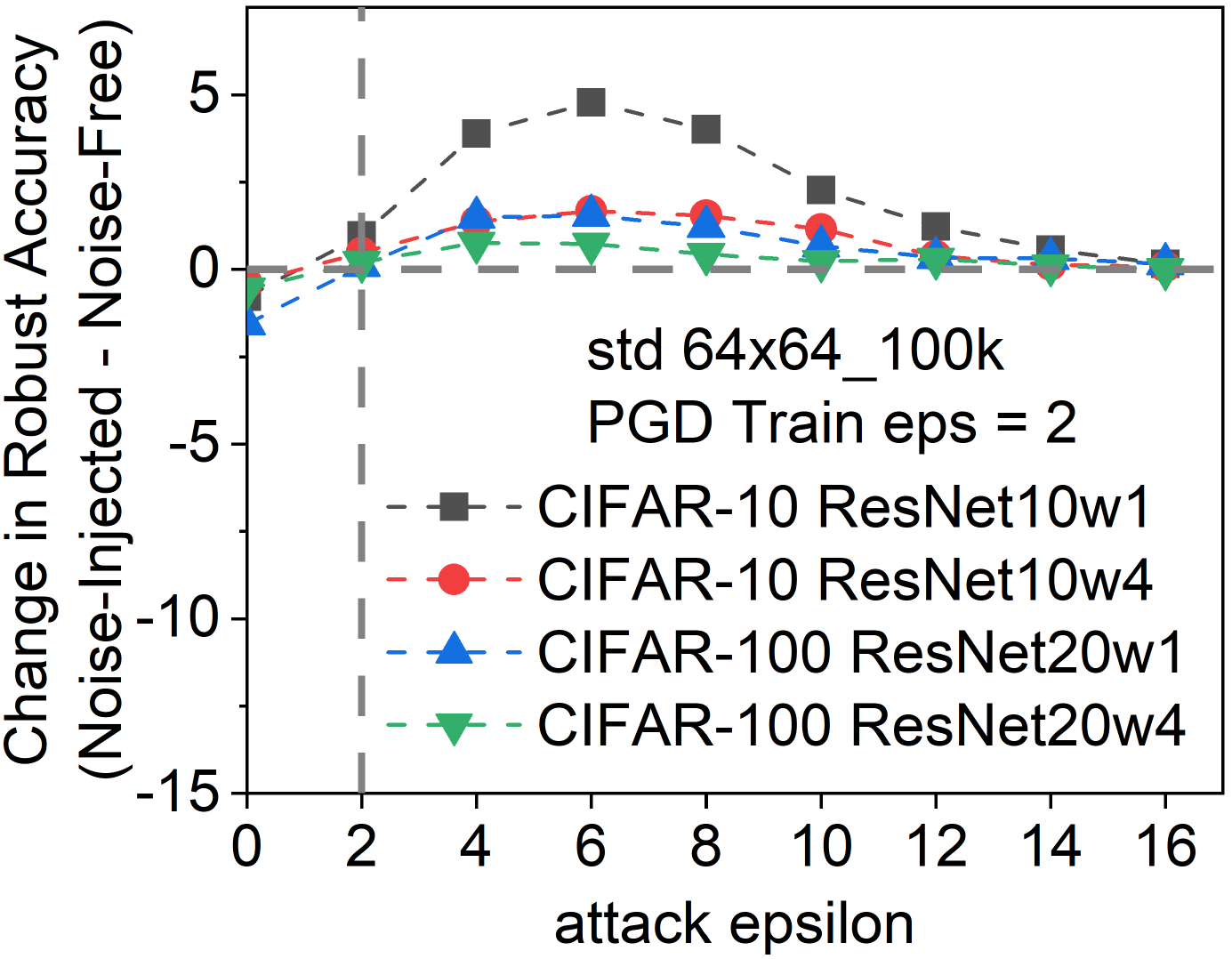}} \\
\sidesubfloat[]{\label{fig:noise-eps2-std+mean-32x32}\includegraphics[width=0.43\linewidth]{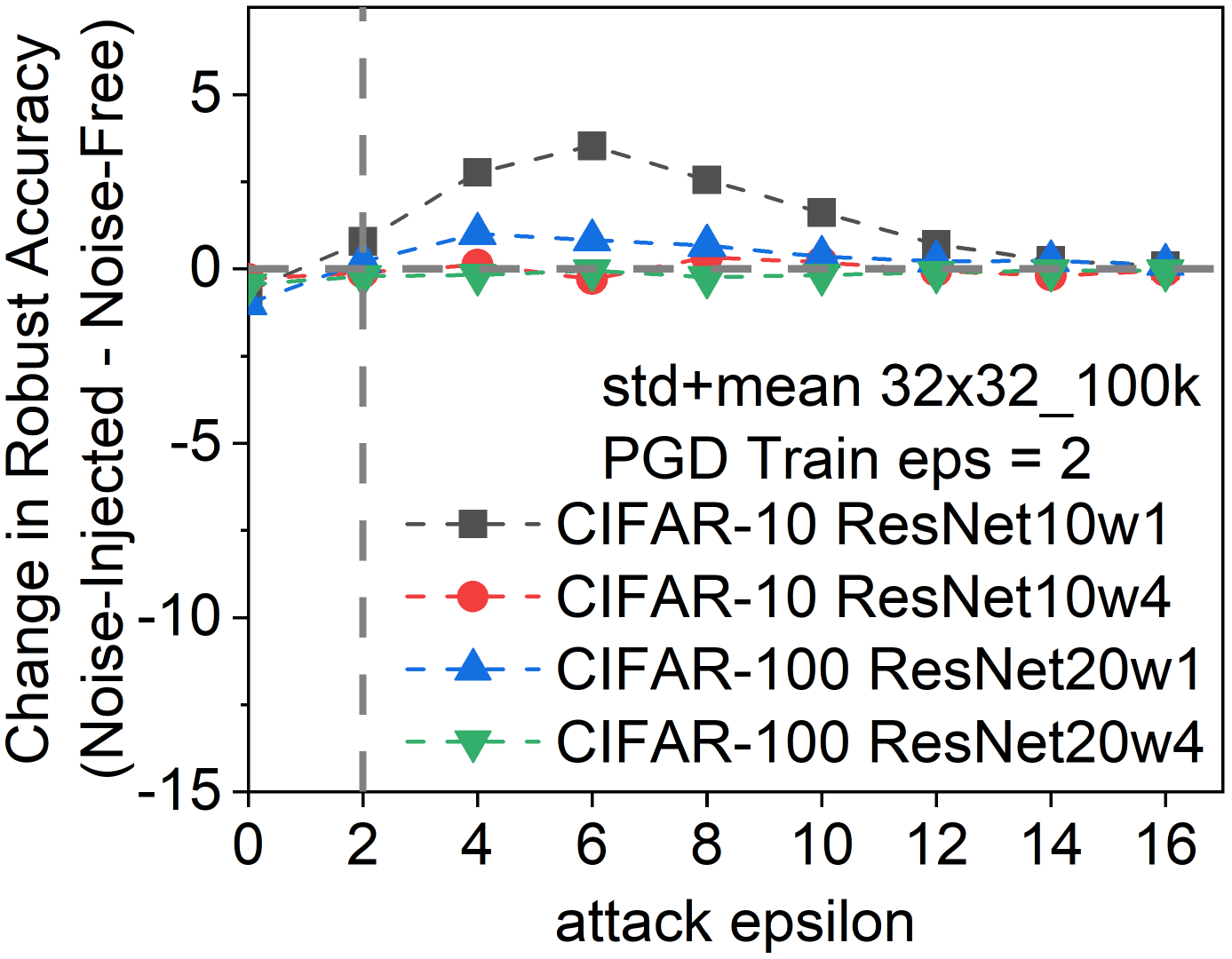}} 
\hfil
\sidesubfloat[]{\label{fig:noise-eps2-std+mean-64x64}\includegraphics[width=0.43\linewidth]{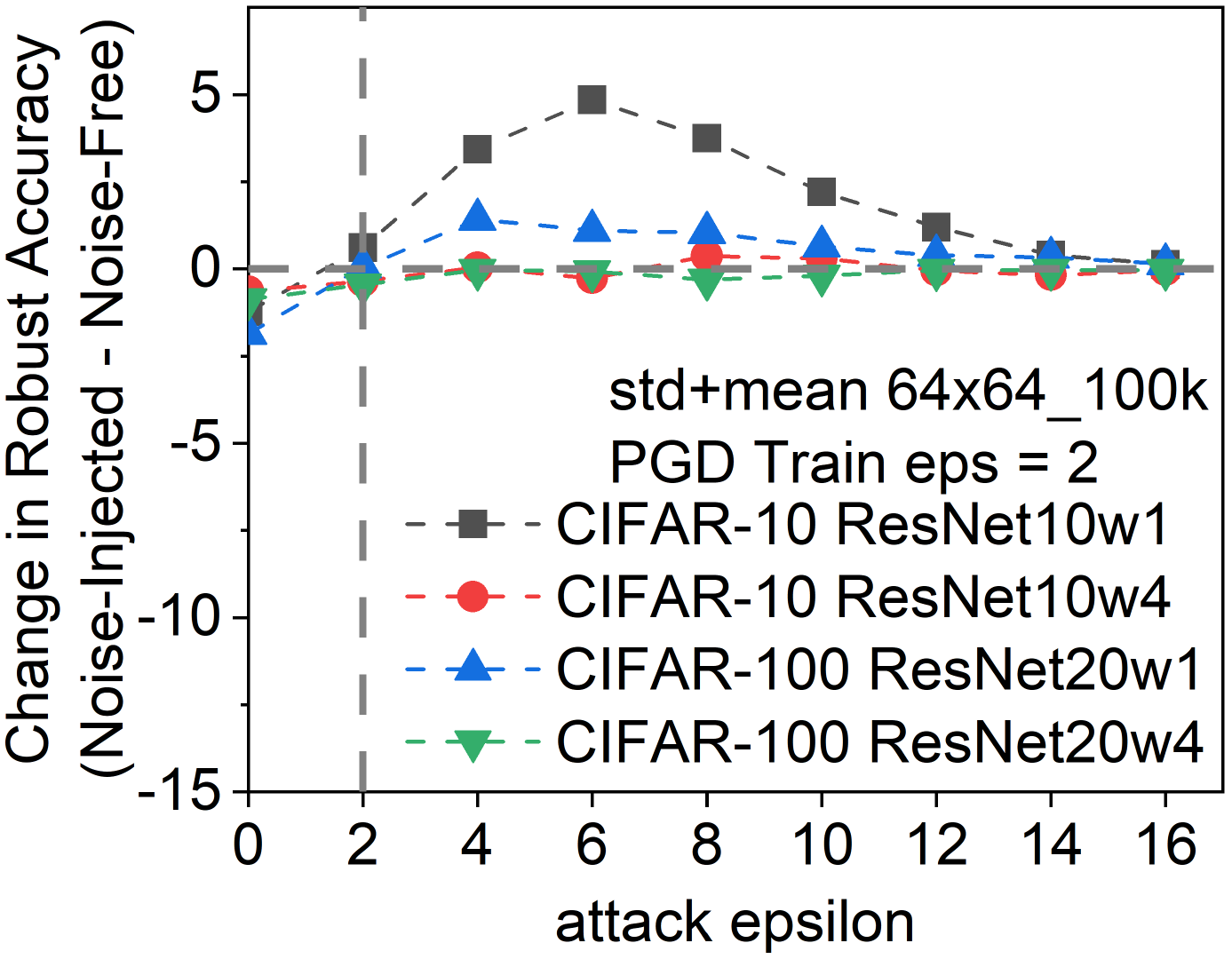}}
\caption{Difference in Adversarial Accuracy (Robustness Gain = Noise-Injected - Noise-Free) for $\epsilon_{attack} \in [0,16]$. 4 network architectures (ResNet10w1, ResNet10w4, ResNet20w1, ResNet20w4) on 2 datasets (CIFAR-10, CIFAR-100) are injected with layer-wise Gaussian noise $\eta \sim \mathcal{N}(\mu,\sigma)$: (a) $\mu=0$, $\sigma$ matching with 32x32\_100k (b) $\mu=0$, $\sigma$ matching with 64x64\_100k (c) $\mu, \sigma$ matching with 32x32\_100k (d) $\mu, \sigma$ matching with 64x64\_100k}
\label{fig:noise}
\end{figure}

In order to further understand the robustness stemming from non-ideal MVM computation due to NVM crossbar non-idealities, we perform an experiment where we compute on digital hardware, but insert Gaussian noise layers after each convolutional and linear layers. The amount of perturbation to the convolutional and linear outputs of Resnet-10 and Resnet-20 is equivalent to the corresponding noise due to NVM hardware i.e. the layer-wise mean or standard deviation are matched with NVM hardware. The evaluations are performed under PGD White-Box attack with $\epsilon_{attack} \in [0,16]$. The attacker assumes noise-free model on digital hardware, and has no knowledge of the noise injection. Because adversarial training is imperative for robustness, the models are PGD adversarially trained \cite{madry2018towards} with $\epsilon_{train} = 2$. In Fig. \ref{fig:noise} the change in robustness between noise-injected and noise-free evaluations is plotted. At $\epsilon_{attack} = 0$, the accuracy is lower for the noise-injected DNN, but we found a robustness gain for $\epsilon_{attack} > 2$, conforming to the trend in experiments on analog hardware robustness. Nevertheless, An immediate observation from Fig. \ref{fig:noise} and Fig. \ref{fig:pgd-eps2-32x32-100k}, Fig. \ref{fig:pgd-eps2-64x64-100k} is that the degree of robustness that one can expect to gain by inserting Gaussian noise layers is far less than the implementation of DNNs on NVM hardware. The models with noise injection whose parameters match with the deviations caused by NVM crossbar 64x64\_100k are more robust than 32x32\_100k counterparts, as the former is more non-ideal \cite{chakraborty2020geniex} and causes greater deviations in layer outputs. In Fig. \ref{fig:noise-eps2-std-32x32} and Fig. \ref{fig:noise-eps2-std-64x64}, the white noise injected has the form $\mathcal{N} \sim \eta (0, \sigma(Z_{analog}-Z_{digital}))$. The robustness gain tops at $3.65\%$ and $4.78\%$ when $\epsilon_{attack}=6$, when the pretrained DNN has architecture ResNet10w1. On the other hand, from Fig. \ref{fig:noise-eps2-std+mean-32x32} and Fig. \ref{fig:noise-eps2-std+mean-64x64}, the robustness gain peaks at $3.53\%$ and $4.85\%$, with the same DNN. Even when both mean and standard deviation are matched, the noise injected DNNs demonstrate only half the robustness as when they are evaluated on analog hardware.

\textit{Adversarially Trained DNNs on Analog Hardware}: The results in Fig. \ref{fig:noise} indicates that there are more inherent characteristics of NVM crossbar non-idealities that make them sufficiently adversarially robust. When we adversarially train the DNNs, the solutions, i.e. model weights, lie in sharp minima in the loss landscape. Then as we perform inference on NVM crossbars, after tiling and bit-slicing, each non-ideal MVM operation results in smaller output values \cite{chakraborty2020geniex}, which effectively deviates the activation values of the convolutional and linear layers from their ideal values, which causes the weights in latter layers to deviate. The weight deviations present at each layer of the DNN alter classifying function, and the locations of sharp minima in the loss landscape. In turn, adversarial images with $\epsilon_{attack} > \epsilon_{train}$ (under advesarial attack) are classified with a DNN model that's first adversarially trained, and then whose activation values and weight matrices are further changed due to underlying hardware. The combined effects of adversarial training and emulating hardware non-idealities ultimately results in the DNN model under strong adversaries classifying some of these out-of-distribution images correctly, i.e. pushing the solution to new local minima, henceforth elevating robustness gain. This robustness can not be simply reproduced with arbitrary perturbations to post-convolution/linear outputs, i.e. Gaussian noise injection. Therefore, both adversarial training and implementation on analog hardware, with careful calibration, are essential to taking advantage of the intrinsic robustness of NVM crossbars.

\section{Conclusion}
\label{sec:conclusion}
NVM crossbars offer intrinsic robustness for defense against Adversarial attacks. While, prior works have quantified these benefits for vanilla DNNs, in this work, we propose design principles of robust DNNs for implementation on NVM crossbar based analog hardware by combining algorithmic defenses such as adversarial training with the intrinsic robustness of the analog hardware. First, we extensively analyze adversarially trained networks on NVM crossbars. We demonstrate that adversarially trained networks are less stable to the non-idealities of analog computing compared to vanilla networks, which impacts their classification accuracy on clean images. Next, we show that under adversarial attack, the gain in accuracy from the non-idealities is conditional on the epsilon of the attack ($\epsilon_{attack}$) being sufficiently large, or greater than the epsilon used during adversarial training ($\epsilon_{train}$). By implementing on NVM crossbar based analog hardware, a DNN trained with lower $\epsilon_{train}$ will have the same or even higher robustness than a DNN trained with a higher $\epsilon_{train}$ while maintaining a higher natural test accuracy. This work paves the way for a new paradigm of hardware-algorithm co-design which incorporates both energy-efficiency and adversarial robustness.

\bibliographystyle{IEEEtran}
\bibliography{main}

\vfill
\end{document}